%% file: main.tex
\documentclass[10pt,twocolumn,letterpaper]{article}

\usepackage{authblk}
\usepackage[pagenumbers]{cvpr} 
\usepackage{graphicx}
\usepackage{amsmath}
\usepackage{amssymb}
\usepackage{booktabs}
\usepackage{mathtools}  
\usepackage{amsmath}
\usepackage{amssymb}
\usepackage{tabulary}
\usepackage{booktabs}
\usepackage[toc,page]{appendix}
\usepackage{xcolor,mdframed}
\usepackage{float}
\usepackage[accsupp]{axessibility}
\usepackage[pagebackref,breaklinks,colorlinks]{hyperref}
\usepackage[capitalize]{cleveref}

\crefname{section}{Sec.}{Secs.}
\Crefname{section}{Section}{Sections}
\Crefname{table}{Table}{Tables}
\crefname{table}{Tab.}{Tabs.}

\def\cvprPaperID{****} %
\def\confName{CVPR}
\def\confYear{2022}

\input{sections/notation}

\begin{document}

\title{OSSO: Obtaining Skeletal Shape from Outside}

\author[1]{Marilyn Keller}
\author[2]{Silvia Zuffi}
\author[1]{Michael J. Black}
\author[3]{Sergi Pujades}

\affil[1]{\normalsize Max Planck Institute for Intelligent Systems, T\"ubingen, Germany} 
\affil[2]{\normalsize IMATI-CNR, Milan, Italy}
\affil[3]{\normalsize Universit\'e Grenoble Alpes, Inria, CNRS, Grenoble INP, LJK, France \authorcr 
  {\tt \small \{mkeller,~black\}@tue.mpg.de \authorcr 
  {\tt \small silvia@mi.imati.cnr.it \quad sergi.pujades-rocamora@inria.fr}}
}

\twocolumn[{
\renewcommand\twocolumn[1][]{#1}
\maketitle
\begin{center}
    \vspace{-0.4in}
    \centering
    \includegraphics[trim=0pt 0pt 40pt 0pt, clip,height=4.6cm]{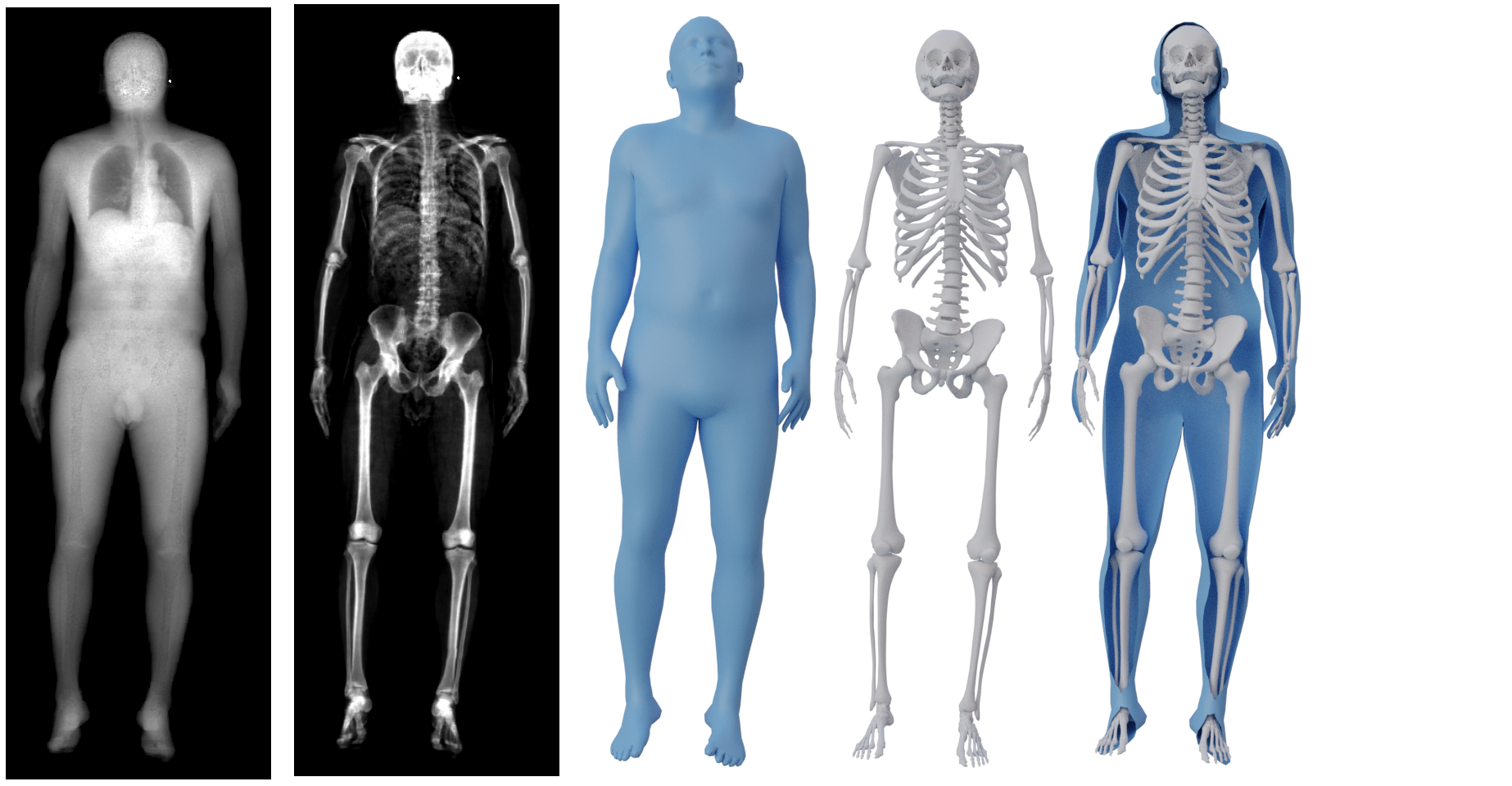}
    \includegraphics[trim=0pt 0pt 80pt 0pt, clip,height=4.7cm]{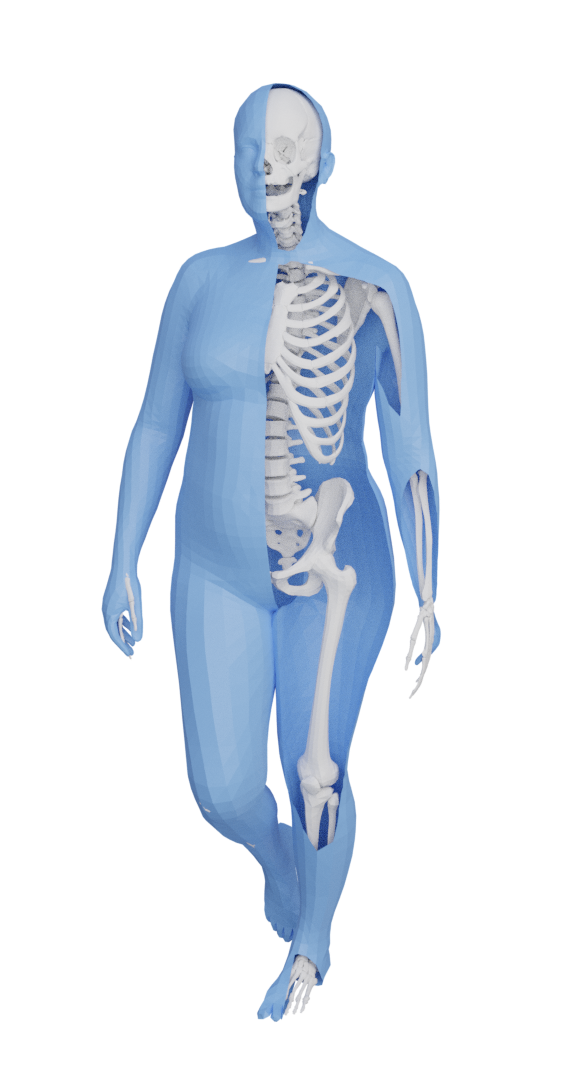}
    \includegraphics[trim=80pt 0pt 0pt 0pt, clip,height=4.7cm]{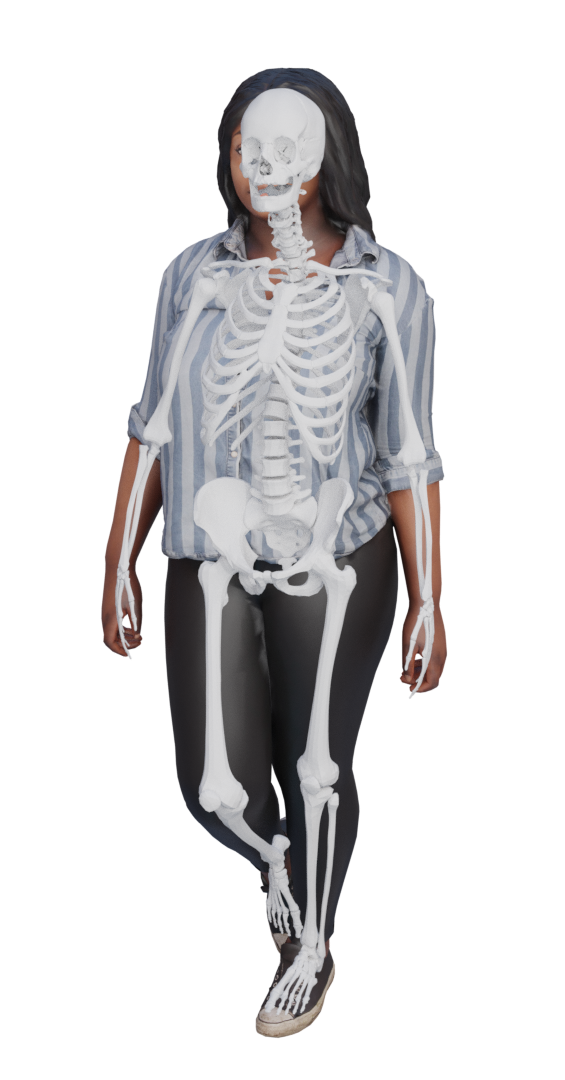}
    \includegraphics[trim=0pt 0pt 50pt 0pt, clip,height=4.7cm]{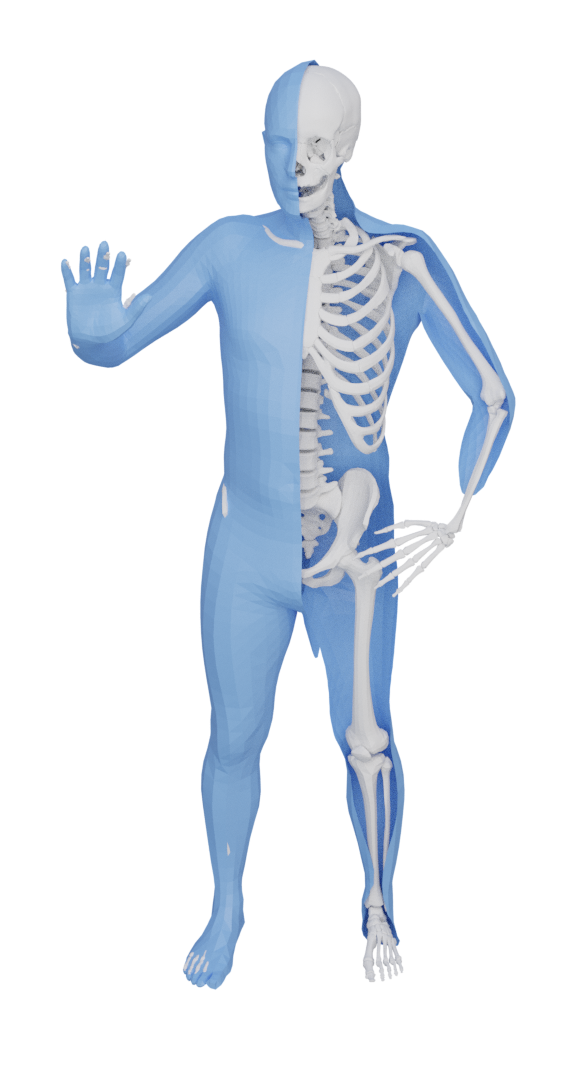}
    \includegraphics[trim=50pt 0pt 0pt 0pt, clip,height=4.7cm]{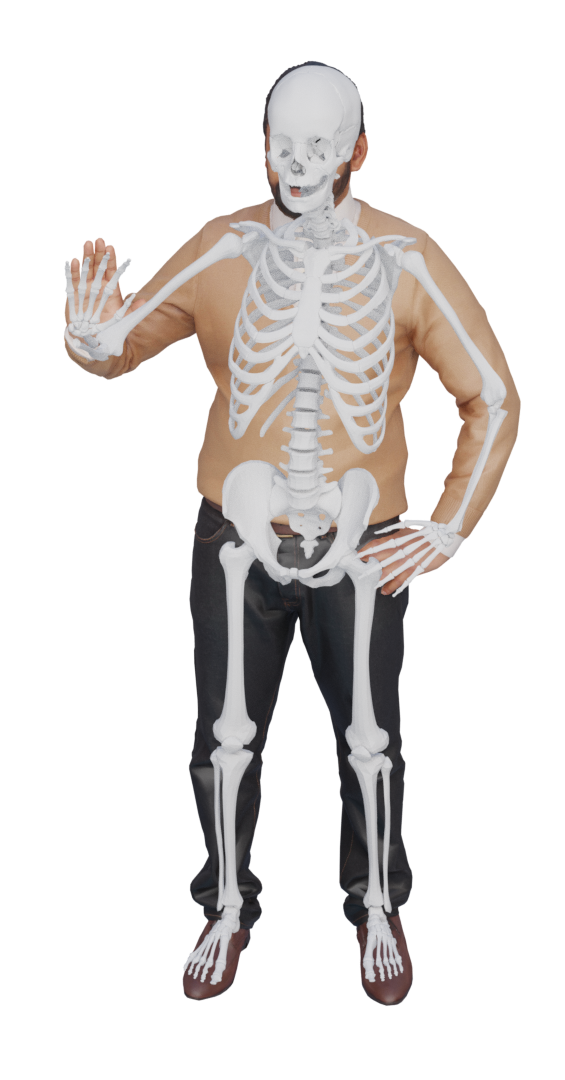}
    \captionsetup{type=figure}
    \captionof{figure}{From DXA scans we learn to predict the skeleton from the body surface. Left: input DXA soft tissue and bone images; body and skeleton shapes fit to the images; bones predicted from the skin; overlay. Right: predicted OSSO skeletons from RenderPeople~\cite{renderpeople} scans.}
    \label{fig:teaser}
\end{center}
}]

\maketitle

\begin{abstract}
\input{sections/abstract}
\end{abstract}

\section{Introduction}
\input{sections/introduction}
\input{sections/related_work}

\input{sections/data}
\input{sections/method}
\input{sections/experiments}
\input{sections/conclusion}
\input{sections/appendix}

\clearpage

{\small
\bibliographystyle{ieee_fullname}
\bibliography{egbib}
}

\end{document}

%% file: sections/notation.tex
\newcommand{\vect}[1]{\mathbf{#1}}
\newcommand{\mat}[1]{\vect{#1}}

\newcommand{\eqnref}[1]{Eq.~(\ref{#1})}
\newcommand{\sectref}[1]{Sec.~\ref{#1}}
\newcommand{\figref}[1]{Fig.~\ref{#1}}
\newcommand{\tabref}[1]{Tab.~\ref{#1}}

\newcommand{\pose}{\boldsymbol{\theta}}                 %
\newcommand{\shape}{\boldsymbol{\beta}}                 %

\newcommand{\vt}{\mat{T}}                        %

\newcommand{\joints}{\mat{J}}

\newcommand{\bones}{K}

\newcommand{\landmarks}{\mathcal{L}}
\newcommand{\regressor}{\mathcal{R}}
\newcommand{\stitched}{{SP}}
\newcommand{\unposed}{{\vt}}
\newcommand{\shapespace}{\mathcal{B}}
\newcommand{\STAR}{{ST}}
\newcommand{\reg}{{\mat{R}}}

\newcommand{\skelinit}{K}

\newcommand{\nbtrainmale}{1000 }
\newcommand{\nbtrainfemale}{1000 }

\newcommand{\nballmale}{1200 }
\newcommand{\nballfemale}{1200 }

%% file: sections/abstract.tex
We address the problem of inferring the anatomic skeleton of a person, in an arbitrary pose,
from the 3D surface of the body; i.e.~we predict the inside (bones) from the outside (skin).
This has many applications in medicine and biomechanics.
Existing  state-of-the-art biomechanical skeletons are detailed but do not easily generalize to new subjects.
Additionally, computer vision and graphics methods that predict skeletons are typically heuristic, not learned from data, do not leverage the full 3D body surface, and are not validated against ground truth.
To our knowledge, our system, called OSSO (Obtaining Skeletal Shape from Outside), is the first to learn the mapping from the 3D body surface to the internal skeleton from real data.
We do so using \nbtrainmale male and \nbtrainfemale female dual-energy X-ray absorptiometry (DXA) scans.
To these, we fit a parametric 3D body shape model (STAR) to capture the body surface and a novel part-based 3D skeleton model to capture the bones.
This provides inside/outside training pairs.
We model the statistical variation of full skeletons using PCA in a pose-normalized space and train a regressor from body shape parameters to skeleton shape parameters.
Given an arbitrary 3D body shape and pose, OSSO predicts a realistic skeleton inside.
In contrast to previous work, we evaluate the accuracy of the skeleton shape quantitatively on held out DXA scans, outperforming the state-of-the art.
We also show 3D skeleton prediction from varied and challenging 3D bodies.
The code to infer a skeleton from a body shape is available at \url{https://osso.is.tue.mpg.de},
and the dataset of paired outer surface (skin) and skeleton (bone) meshes is available as a Biobank Returned Dataset.
This research has been conducted using the UK Biobank Resource.

%% file: sections/introduction.tex
The estimation of 3D human pose and shape (HPS) from images, video, and other data sources is widely studied and has
many applications.
Current methods for HPS exploit detailed models of the  {\em visible surface} of the human body 
learned from 3D scans.
While the surface shape is accurate, these models are all based on a ``skeleton" that only approximately models the kinematic structure of the body using a small number of linear segments with ball joints. 
While these simplified skeletons are useful for the animation of virtual characters and action recognition, they are not appropriate for applications in medicine and biomechanics.
Moreover, each 3D body model defines its own kinematic skeleton;
transferring 3D pose information between them often requires optimization and introduces errors.
To be more widely relevant, HPS methods need to output a skeleton that corresponds to the true, anatomic, human skeleton.
No statistical body model exists that captures both the detailed outer surface of the body and the anatomic skeletal structure inside.
The key problem is the lack of paired data capturing the inside and outside of the body.

In this work, we address the problem of inferring the human anatomic skeleton, i.e.~the bone shapes and locations, solely from surface observations.
That is, we {\em infer the bones from the skin.}
To that end, we learn a statistical model of the skeleton shape and its correlation with the skin surface (Fig.~\ref{fig:teaser} left).
Given a posed body, our method predicts the skeleton from the body shape, and poses it inside subject to anatomic constraints (Fig.~\ref{fig:teaser} right).

Anatomic body models with skin and bones are important in computer graphics, medicine and biomechanics, enabling realistic animation of the body anatomy and physical simulation of body motion.  
Existing  state-of-the-art anatomic models~\cite{delp2007opensim, seth2018opensim, damsgaard2006analysis, mcguan2001human}  represent different body parts: skin, muscles, organs, skeleton. They are mainly developed for sports, health applications or educational purposes.
While very detailed, they don't generalize easily to new subjects.
Graphics-oriented anatomic skeletons~\cite{ali2013anatomy, Zhu2015, bauer2016modelisation, Kadlecek2016} can deform the individual bones with simple geometric transformations (e.g.~scale or affine) and can be fit to new subjects. 
However, these deformations lack anatomic realism relative to actual skeletons.
We argue that this realism can be improved using a data-driven strategy.

In computer vision, 3D statistical shape models of the human body are widely used~\cite{anguelov2005scape, Loper2015smpl, xu2020ghum, osman2020star}.
These have two elements in common:
they model the human external shape, i.e.~the skin surface, and they are learned from data.
Using thousands of 3D scans, these models capture the statistical variability of the human body shape. 
Here we use STAR \cite{osman2020star}, because it has a richer shape representation than SMPL~\cite{Loper2015smpl}.
Such models, however, employ a simplified kinematic skeleton and joints.
While they can be readily inferred from data, the idealized skeletal structure means that they cannot be used for applications in biomechanics.

To address these issues, we take a data-driven approach and learn a statistical skeleton shape model, as well as the mapping from body shape to this skeleton model.
Our method, {\bf OSSO} (Obtaining Skeletal Shape from Outside), takes a STAR model instance of any shape and pose, and estimates its corresponding skeleton. The skeleton can then be animated by reposing the STAR model.

\begin{figure}
     \begin{mdframed}[backgroundcolor=black]
     \includegraphics[width=0.159\columnwidth]{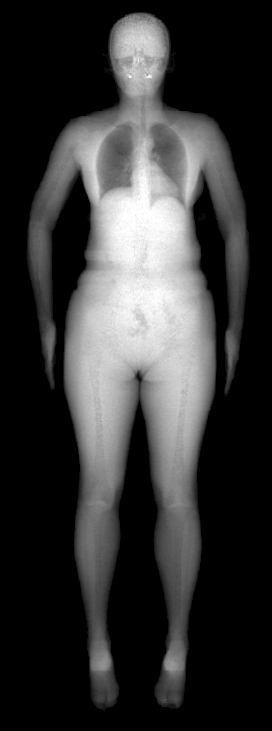}
     \includegraphics[width=0.159\columnwidth]{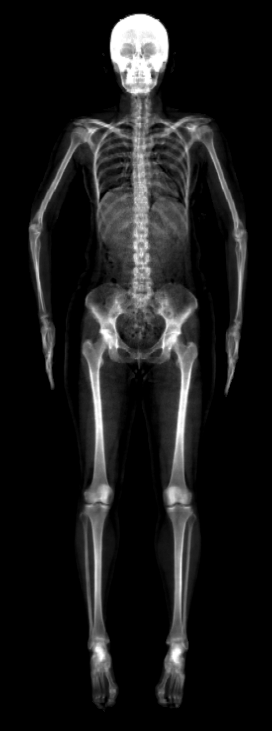}
     \includegraphics[width=0.159\columnwidth]{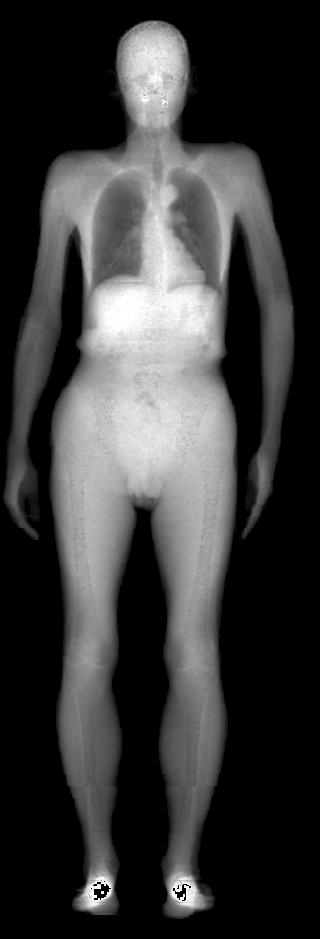}
     \includegraphics[width=0.159\columnwidth]{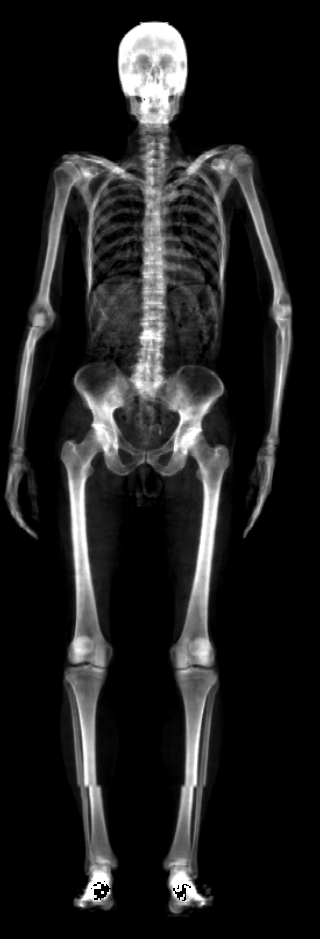}
     \includegraphics[width=0.159\columnwidth]{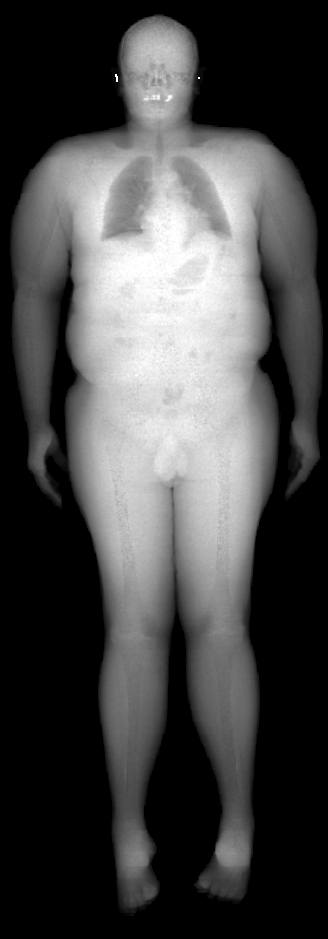}
     \includegraphics[width=0.159\columnwidth]{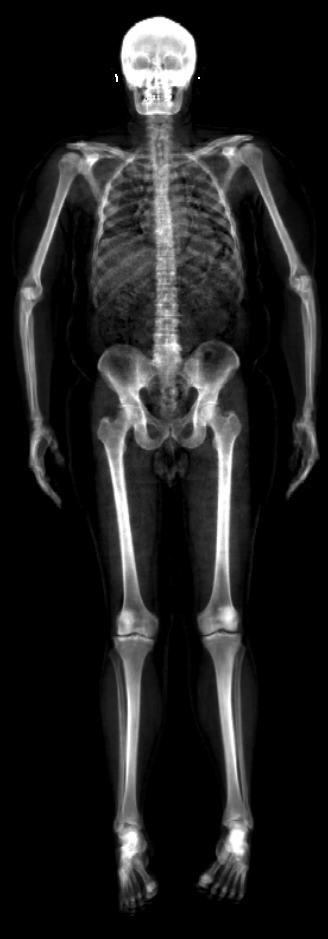}
     \\
     \includegraphics[width=0.159\columnwidth]{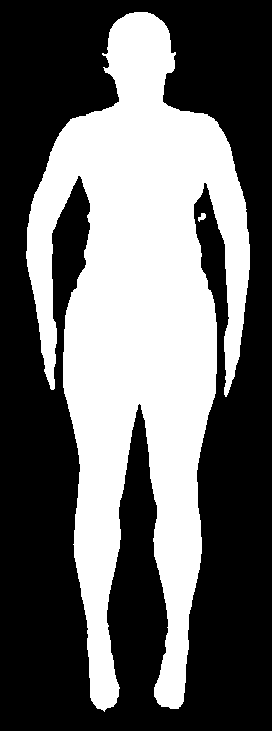}
     \includegraphics[width=0.159\columnwidth]{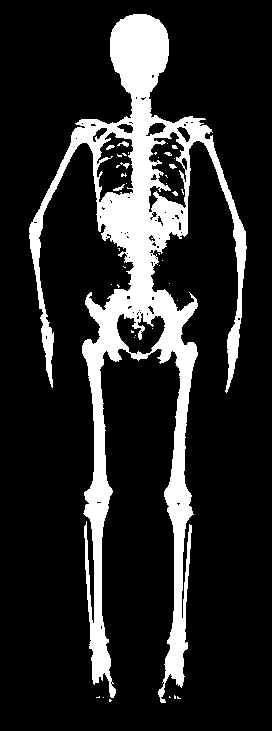}
     \includegraphics[width=0.159\columnwidth]{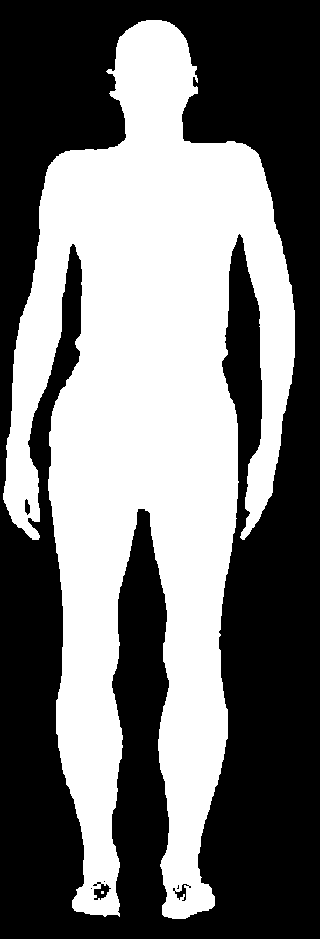}
     \includegraphics[width=0.159\columnwidth]{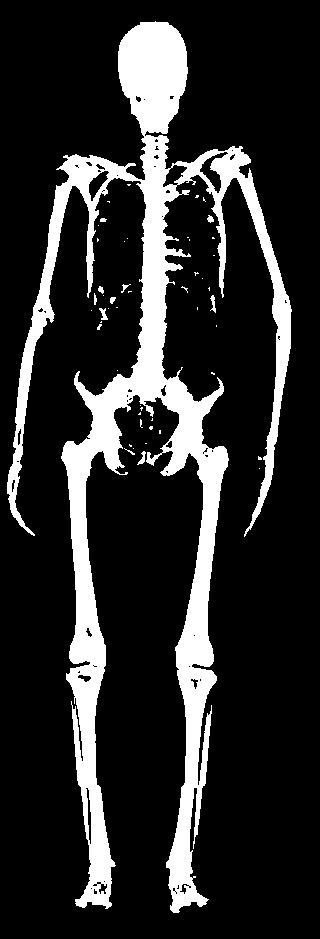}
     \includegraphics[width=0.159\columnwidth]{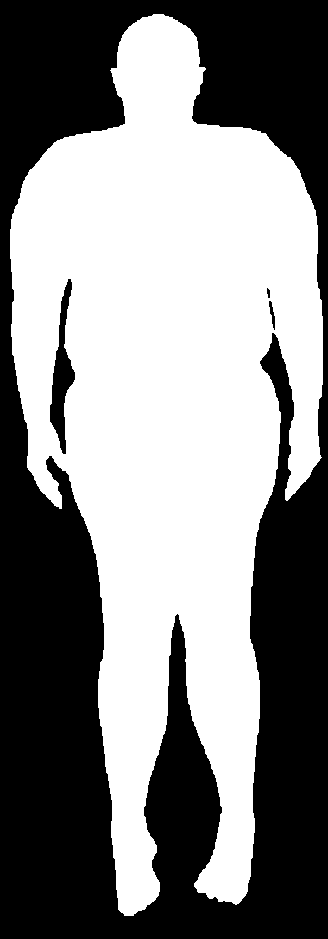}
     \includegraphics[width=0.159\columnwidth]{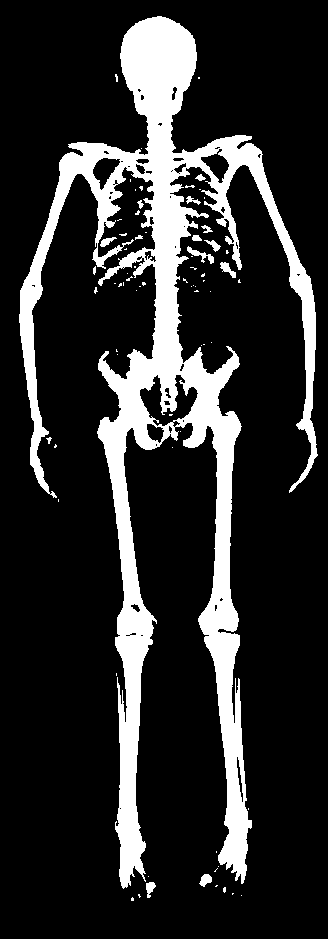}
     \end{mdframed}
     \vspace{-0.15in}
     \caption{Top: Pairs of soft tissue ($I_{S}$) and bone ($I_{B}$) DXA images.
     Bottom: Computed skin ($M_S$) and bone ($M_B$) masks.}
 \label{fig:skin_skel_segmentations}
 \end{figure}

The key problem, however, is obtaining training data that simultaneously gives the inside and outside of the body in 3D.
Most imaging technologies that simultaneously capture the inside and outside of the body use ionizing radiation, which is harmful to humans; e.g.~Computed Tomography (CT) and X-rays.
This means that such data is extremely limited, preventing learning-based methods.
Our insight is to use dual-energy X-ray absorptiometry (DXA) data. 
DXA scans use low-dose X-rays to measure bone mineral density and body fat composition. 
The radiation level is so low that it is certified to be used on healthy patients for clinical studies, such as the UK Biobank~\cite{sudlow2015biobank}. 
In a DXA scan, two images are computed by combining two different energy levels: a soft-tissue image $I_{S}$ and a bone image $I_{B}$ (\figref{fig:skin_skel_segmentations}).
In $I_{S}$ the silhouette of the body can clearly be seen, whereas $I_{B}$ reveals the structure and shape of the bones.

Unfortunately, DXA does not produce 3D data.
Consequently, we fit the STAR body model to the soft-tissue image to obtain an estimate of the outer 3D body surface.
We also employ a constrained part-based fitting method to fit bones to the DXA bone image.
These then provide pairs of inside and outside data for training.
We use \nballmale male and \nballfemale female DXA images from UK Biobank~\cite{sudlow2015biobank}, which we split into training and evaluation sets.
From the training set we learn skeletal shape variation and the mapping from outside to inside.
Given a new body shape and pose, OSSO reposes the body to a canonical pose and predicts the skeleton inside.
It then reposes the skeleton to the input pose, subject to various anatomic constraints.
With ground truth DXA scans, we validate the reposing in lying down poses
and show that OSSO outperforms Anatomy Transfer \cite{ali2013anatomy}.
We also demonstrate the use of OSSO by estimating skeletons for a variety of body shapes and poses.

OSSO makes the following contributions:
(1) We fit a 3D body model to DXA images to obtain 3D body shape.
(2) We fit a collection of bones to DXA images using a variety of constraints.
(3) We learn a statistical (PCA) model of skeleton shape variation, capturing correlations between bones.
(4) We learn a mapping from external body shape to internal skeleton shape.
(5) Given a 3D body in any pose, we repose the body, predict the skeleton, and repose under physical constraints to obtain a plausible posed skeleton.
(6) We demonstrate superior performance to other approaches, validated on DXA imagery.
(7) We show varied reposing results for 3D bodies estimated from 3D scans.
(8) Inference code is available for research purposes.
(9) The paired outer surfaces (skin) and skeleton (bone) meshes are made available  as a Biobank Returned Dataset.

In summary, OSSO provides a data-driven approach to enrich 3D human pose and shape estimation with skeletal information.
This is a step towards biomechanics in the wild. 
Methods that estimate models like SMPL or STAR from images and video can immediately use OSSO to estimate the skeletal structure.
While many methods provide some sort of visually appealing skeleton, our work is the first to learn and validate such a skeleton based on data of the inside and the outside of the body.

%% file: sections/related_work.tex
\section{Related Work}

We review work on data-driven skin and bones models,
and methods that create personalized anatomic models.

{\bf Data-driven skin models.}
Learned statistical body models ~\cite{allen2003space, anguelov2005scape, Loper2015smpl, xu2020ghum, osman2020star} leverage large datasets of 3D scans.
In our work, we use STAR~\cite{osman2020star} to represent the body surface with two parameter vectors $(\shape_S, \pose_S)$ controlling the shape and pose of the body, respectively.

{\bf Data-driven bone models.}
In medicine, patient-specific 3D bone models are very valuable. 
Since many scanning modalities are 2D (X-ray), numerous methods address fitting 3D models to 2D images. 
However, as pointed out by a review of existing methods~\cite{reyneke2018review}, most models are restricted to individual bones (or groups) and are learned from 3D information. 
Our method learns a 3D skeleton model from 2D DXA images.

{\bf Fitting models to images.}
The literature of methods fitting 3D body models to 2D images is  wide~\cite{guan2009estimating, sigal2007combined,bogo2016keep,lassner2017unite,varol2018bodynet, Pavlakos2019smplx, pavlakos2018learning, kanazawa2018end, chen2020towards, guler2018densepose, sun2019deep} and was recently surveyed~\cite{zheng2020deep}.
However, less work fits such models to X-ray images.
Pansiot and Boyer \cite{pansiot2018cbct} leverage video-based surface motion capture to recover a volumetric representation of the hand from planar X-ray images.
In our work, we leverage a silhouette term \cite{zuffi2018lions} and regressed landmarks to fit models of the body surface and the skeleton to the segmented DXA images.

{\bf Personalized anatomic models.}
Several works have addressed the problem of creating a personalized anatomic model of a subject from external or internal observations.

Gilles et al.~\cite{Gilles2010} propose a morphing algorithm to register a template skeleton to a target skeleton mesh or 3D image. The registration is done by alternating elastic and plastic deformations, and joint position corrections constrained by prior kinematic information. 
At each step, the deformed individual bones are projected onto a statistical model of the bone to ensure plausible bone shapes. Since the bone shape space is built from synthetically deformed bone shapes and not from actual bone scans, it is unclear how representative the shape space is of the population. 

Saito et al.~\cite{saito2015computational} simulate the growth of fat, muscle, and bones to generate new body shapes. Kadle\v{c}ek et al.~\cite{Kadlecek2016} propose a physics-based anatomic model that can be adapted to input 3D scans, where, similar to Zhu et al.~\cite{Zhu2015}, the bones are deformed using linear blend skinning with bounded bi-harmonic weights.
While the obtained bones are visually plausible, these models are neither learned from data nor validated against it.

In Phace~\cite{Ichim2017phace}, two independent face and skull shape models are combined to infer a probabilistic distribution of the face given a skull. This goal is similar to ours, as we want to infer the skeleton shape from the body shape. 
In contrast, however, we do not have a statistical shape space for the whole skeleton, and thus we learn it.

Wang et al.~\cite{wang2019hand} propose a method to scan a hand with MRI (Magnetic Resonance Imaging) and create an accurate, personalized anatomic model. The model can plausibly extrapolate to new unseen poses with high visual realism. 
The created model is person-specific and cannot be inferred from skin observations.

Zoss et al.~\cite{zoss2019accurate} propose a method to track the invisible jaw from the visible skin surface. In their method, they propose a calibration phase to adapt the jaw size to a new subject.
OSSO goes further, as the shape of the bones is estimated from the outside in addition to their location.

Bauer et al.~\cite{bauer2016modelisation} infer the skeleton of a subject from RGBD images of the skin. Their skeleton inference method is based on Anatomy Transfer~\cite{ali2013anatomy} with extra constraints positioning the bones inside the body and avoiding bone intersections. Bones are parametrized with affine transformations, and results are not validated on medical data.

The recent BASH model~\cite{schleicher2021bash} couples a musculoskeletal biomechanical model to the SCAPE model~\cite{anguelov2005scape}. 
BASH generates a skeleton from sparse measurements, but the obtained anatomy is not validated on medical images.
Unlike STAR, the SCAPE model does not guarantee constant bone lengths when reposed.
The main difference with BASH is that we use a data-driven approach to learn to infer the shape of the skeleton inside the human body, and we validate on medical images.

The most related work to ours is Anatomy Transfer (AT)~\cite{ali2013anatomy}.
In AT, a skeleton is generated from only the external shape of an avatar, without requiring a particular initialization. 
Given a target body shape, an anatomical template is morphed to match it. The surface of the anatomical model is deformed using Laplacian deformations, and the underlying anatomy is interpolated, except for the bones, which are deformed with affine transformations.
The skeletal structure is enforced by defining springs between the bones that keep them coherent.
In our work, we use a similar approach by leveraging the Stitched Puppet graphical model~\cite{zuffi2015stitched}. 
While AT generates a plausible anatomy for any kind of humanoid avatar, the generated anatomy is not validated on real data. 
Our work goes beyond AT by using data to learn the skeletal deformation space and by providing a quantitative evaluation on real DXA images.
We consider Anatomy Transfer to be the baseline and compare our predictions to theirs.

Lastly, recent work by Wong et al.~\cite{wong2021pose}, shows that the human internal body composition can be predicted from solely body surface measurements.
Our work is complementary to theirs, as OSSO predicts the geometry and location of the skeleton inside the body surface.

%% file: sections/data.tex
\section{Data}
\label{sec:dxa_alignment}

A key contribution of our work is to create a unique dataset for training and evaluation that contains paired outer surface (skin) and skeleton (bone) meshes $( \reg_S, \reg_B)$ from DXA images.
The dataset is made available to the community as a Biobank Returned Dataset.

\begin{figure*}[t]
  \centerline{
  
  \includegraphics[trim=0pt 95pt 0pt 125pt, clip,width=1\textwidth]{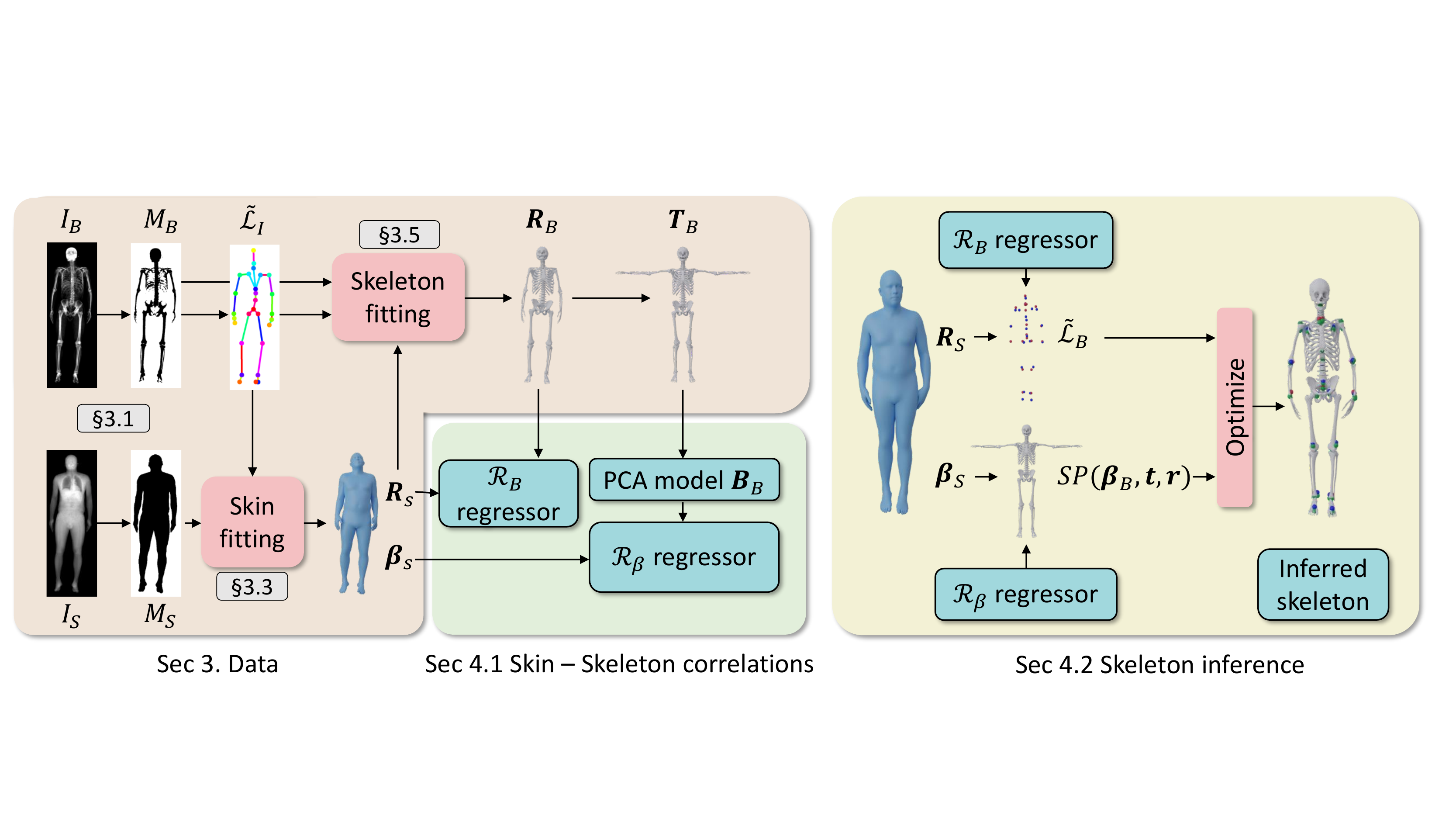}
  }
  \caption{Overview of OSSO: learning and inference. 
  From the input DXA images $(I_B, I_S)$ we obtain the skeleton and skin masks $(M_B, M_S)$. From the skeleton mask $M_B$, we 
  predict 2D landmarks $\landmarks_i$ and use them to register STAR to $M_S$ and obtain $\reg_S$ and $\shape_S$.
  We then register our skeleton graphical model to $M_B$, $\landmarks_i$ and $\reg_S$ and obtain $\reg_B$ and its unposed version $\unposed_B$.
  From the unposed skeletons $\unposed_B$ we learn a skeleton shape space $B_{B}$;
  with paired $(\reg_S, \landmarks_B(\reg_B))$ we learn the regressor $\regressor_B$ 
  and with paired $(\shape_S, \shape_B)$ we learn the regressor $\regressor_{\beta}$.
  At test time, from the body surface ($\reg_S, \shape_S$) we
  regress the skeleton shape $\regressor_{\beta}(\shape_S) = \shape_B$
  and optimize its pose to match the regressed locations $\regressor_B(\reg_S) = \tilde{\landmarks}_B$.
  }
  \label{fig:pipeline}
\end{figure*}

Creating the dataset has several steps: 
(1) we segment DXA images to get the silhouettes of the body and bones (\sectref{sec:dxa_silhouettes}),
(2) we create synthetic skeleton silhouettes and use them to learn to predict landmarks (\sectref{sec:landmarks}),
(3) we register STAR~\cite{osman2020star} to the skin silhouette images (\sectref{sec:skin_fit}),
(4) we create a custom skeleton model (\sectref{sec:skel_gloss}) 
and register it to real skeleton binary masks (\sectref{sec:skel_gloss_fit}).
\figref{fig:pipeline} shows an overview of the dataset creation procedure.

\subsection{Skin and skeleton masks from DXA images}
\label{sec:dxa_silhouettes}

From the input images ($I_{S}$, $I_{B}$), we compute the corresponding skin and skeleton segmentation masks ($M_S$, $M_B$).
For the skin mask $M_S$, we threshold $I_{S}$.
As some small artifacts remain, mainly due to pixels in the lungs with low intensity values,
we detect the closed contours on the image and fill in small areas.
In \figref{fig:skin_skel_segmentations} we show pairs of input $I_{S}$ and the obtained mask $M_S$.

The automatic segmentation of bones in DXA images is still an open problem. Often, 
DXA image regions are obtained by manually annotating keypoints~\cite{shepherd2017modeling}, and accurate segmentations are performed manually~\cite{burkhart2009manual}. Moreover, these methods only focus on a small set of bones.
Jamalud et al.~\cite{jamaludin2018scoliosis} use a U-Net to segment body parts from DXA scans that are relevant for scoliosis classification. Unfortunately the code is not available

In our work, we use a simple heuristic to automatically segment the bone tissue in the bone images:
 we assume that the $X\%$ brightest pixels in each $I_{B}$ image belong to bone tissue. We empirically set $X=20\%$ for the male DXAs and $X=17\%$ for the female.
As small artifacts remain (earrings, clothing, etc.), we remove small connected components with an area less than 50 pixels.
Note that we do not claim to segment all bone tissues in the DXA images.
While our segmentations are coarse, they capture the structure and location of the bones inside the body (as shown in \figref{fig:skin_skel_segmentations}); this is what we need to fit a 3D skeleton to them.

\subsection{Computing landmarks from the silhouettes}
\label{sec:landmarks}
Many model-based human pose estimation methods rely on fitting projected 3D joints to 2D landmarks.
Landmark detection must be automated as we fit thousands of DXA images.
Existing landmark detectors, of course, do not work with DXA imagery. Consequently, we train a landmark detector for skeleton binary masks $M_B$.
To do so, we generate synthetic training data using an initial skeleton mesh similar to the one from Anatomy Transfer (AT) \cite{ali2013anatomy}.
We rig the skeleton so that we can control it with the STAR shape and pose parameters (see Sup.~Mat.~for details),
and define 29 landmarks on the skeleton mesh, $\landmarks_I$,
that are in correspondence with the 3D joints of the STAR model.
This allows us to generate skeletons of different shapes in lying down poses, which we render as binary images with the projected landmarks, giving us paired training data.

To bridge the domain gap between the synthetic silhouettes and the DXA segmentations, we augment the data by eroding and partially masking the rendered skeleton silhouettes, while keeping the landmarks fixed. 
From the synthetic silhouettes of the skeleton, we train the landmark detector using a stacked hourglass network~\cite{newell2016stacked}.
In Sup.~Mat., we provide the details and evaluation of this network.

\subsection{Skin surface from DXA}
\label{sec:skin_fit}

A key step is to estimate the 3D body shape of a subject from their 2D DXA segmentation $M_S$.
There is prior work on fitting a body surface model to DXA images using a silhouette \cite{lee2009measurement, rossi2012novel}. 
These methods, however, assume that a 3D scan of the subject is available. 
Since this is not the case for us, we fit the 3D parametric model STAR~\cite{osman2020star} to the silhouette and our predicted landmarks from above.

Since the registration is only conditioned by a silhouette and 2D landmarks, we need a good pose prior. 
Thus, we collected twelve 3D scans of people lying down, computed their STAR poses,
learned a distribution of poses, and use this as a pose prior
 $E_{\theta}$ as in \cite{bogo2016keep}. 
Moreover, we enforce the hands to stay in the coronal plane with the cost $E_{h}$ penalizing the distance between the hand and the middle of the thighs.

To fit STAR to the silhouettes, we use the same optimization strategy as in~\cite{zuffi2018lions}
and effectively solve for the STAR shape and pose parameters $(\hat{\shape}_S, \hat{\pose}_S)$ that minimize:
\begin{equation}
\begin{split}
    E_S & (\shape_S, \pose_S ; M_S, \tilde{\landmarks}_I) = E_{sil} (\STAR(\shape_S, \pose_S), M_S) \\
                           &  + \lambda_I || P(\joints(\shape_S, \pose_S)) - \tilde{\landmarks}_I || \\
                           &  + \lambda_{\beta} ||\shape_S|| + \lambda_{\theta} E_{\theta}(\pose_S) + \lambda_{h} E_{h}(\STAR(\shape_S, \pose_S)), \\
\end{split}
\end{equation}
where 
$\STAR(\shape_S, \pose_S)$ is the STAR mesh and
$\joints(\shape_S, \pose_S)$ are the STAR joint locations.
$E_{sil}$ enforces the projection of the STAR mesh to match the silhouette (as in Eq. 6 of~\cite{zuffi2018lions})\footnote{We use the open available implementation at \url{https://github.com/silviazuffi/smalr_online}},
$\tilde{\landmarks}_I$ are the predicted landmarks, and
$P$ is the orthographic camera projection function.
We denote the obtained mesh $\reg_S=\STAR(\hat{\shape}_S, \hat{\pose}_S)$ (see $\reg_S$ and $\shape_S$ in \figref{fig:pipeline}).

This approach works well, but can fail for cases like severe scoliosis
or limb atrophy.
These cases have high silhouette fitting errors, and we use these errors to detect and remove failure cases automatically from the final dataset.

\subsection{Skeleton graphical model}
\label{sec:skel_gloss}

Now that we have the skin surface, we need the skeleton inside.
To register a 3D skeleton model to the DXA bone mask $M_B$, we need a model where the individual bones can freely move and deform but can be controlled with connectivity. Our initial skeleton model (\sectref{sec:landmarks}) is not well suited for this task.
Thus, we create a new skeleton model, capitalizing on the {\it stitched puppet} ~\cite{zuffi2015stitched} 
and the synthetic shape deformation space from {\it GLoSS}~\cite{zuffi20173d}.
The {\it stitched puppet} provides an ideal graphical model formulation, allowing each group of bones to move independently, 
while the stitching potentials enforce the coherence of the full skeleton.

Starting with the same AT skeleton template as before,
we manually define groups of bones that belong to the same anatomic part, and define the interfaces between these parts.  
Unlike the original formulation \cite{zuffi2015stitched}, the stitching potentials are not defined as springs between corresponding surface points. 
Instead, we define corresponding points by selecting vertices of connected parts that have a distance below a certain threshold (see Sup.~Mat.).
Also, unlike \cite{zuffi2015stitched}, we do not use a graphical model inference method to register the model to data.
We refer to the skeleton mesh as $\stitched(\shape_B, \vect{t}, \vect{r})$, where $(\shape_B, \vect{t}, \vect{r})$ are respectively the shape, translation and rotation of all the skeleton parts. 
As we use the same AT skeleton template, the landmarks $\landmarks_I$ are properly defined.

\subsection{Skeleton from DXA}
\label{sec:skel_gloss_fit}

Next, we use the binary skeleton mask $M_B$, the estimated landmarks $\tilde{\landmarks}_I$ and the skin registration $\reg_S$ and optimize for the skeleton model parameters $(\hat{\shape}_B, \hat{\vect{t}}, \hat{\vect{r}})$ that minimize $E_B(\shape_B, \vect{t}, \vect{r}) = $
\begin{equation}
E_{data}(\shape_B, \vect{t}, \vect{r}; M_B, \tilde{\landmarks}_I, \reg_S) + E_{prior}(\shape_B, \vect{t}, \vect{r}),
\end{equation}
where $E_{data}(\shape_B, \vect{t}, \vect{r}; M_B, \tilde{\landmarks}_I, \reg_S) = $
\begin{equation}
\begin{split}
& E_{sil} (P(\stitched(\shape_B, \vect{t}, \vect{r})), M_B) \\
                           & + \lambda_I || P(\landmarks_I(\stitched(\shape_B, \vect{t}, \vect{r}))) - \tilde{\landmarks}_I || \\
                           & + \lambda_{i} E_{i}(\stitched(\shape_B, \vect{t}, \vect{r}, \reg_S )) 
\end{split}
\end{equation}
and $E_{prior}(\shape_B, \vect{t}, \vect{r}) = $
\begin{equation}
\begin{split}
& \lambda_{shape} ||\shape_B|| + \lambda_{pose} ||\vect{r}-\vect{r_T}|| + \\
      & + \lambda_{sti} E_{sti} (\stitched(\shape_B, \vect{t}, \vect{r})) + \lambda_{sy} E_{sy} (\stitched(\shape_B, \vect{t}, \vect{r})) \\
\end{split}
\end{equation}
where $\vect{r_T}$ are the rotations of the bones in a manually defined lying down pose, $E_{sti}$ is the $-\log$ of the stitching potentials described in~\cite{zuffi2015stitched}, and $E_{sy}$ forces the symmetric body parts on the right and left to have a similar shape.
Note how landmarks $\landmarks_I$ here are now written as a function of the skeleton mesh.

The cost $E_{i}$ enforces the skeleton to be inside the body $\reg_S$ and in contact with the skin in some manually defined regions (knee, tibia, elbow). The implementation details of this cost are found in Sup.~Mat.
After the optimization, we obtain the skeleton's pose and shape parameters
$(\hat{\shape}_B, \hat{\vect{t}}, \hat{\vect{r}})$ and the registered skeleton mesh $\reg_B=\stitched(\hat{\shape}_B, \hat{\vect{t}}, \hat{\vect{r}})$ (see $\reg_B$ in ~\figref{fig:pipeline}).

\textbf{Unposing the skeleton registration.}
Before we can learn a skeleton shape space, we need to pose-normalize the optimized skeletons $\reg_B$. 
While obtaining the unposed mesh $\unposed_S$ of a STAR fit $\reg_S$ is straightforward - one just zeros the pose parameters, unposing $\reg_B$ is ill-posed as one can zero the rotations $\vect{r}$, but the translations $\vect{t}$ need to be adjusted. 
To constrain the problem, we make the hypothesis that the 3D offsets between the skin and skeleton do not vary much from one pose to another.
From the registrations $(\reg_S, \reg_B)$ of a low BMI subject, we define 3113 pairs of skin and skeleton indices $\{(sn_p, sk_p)\}$ and define ${\bf d_p^0} = (\reg_S[sn_p] - \reg_B[sk_p])$ as their initial 3D offset.

This allows us to define a signed distance cost between the unposed meshes $E_d(\unposed_S, \unposed_B) = \sum_p w_p \cdot (\unposed_S[sn_p] - \unposed_B[sk_p]) - {\bf d_p^0}$, with $w_p = sign(\unposed_S[sn_p] - \unposed_B[sk_p]) \cdot N(\unposed_B[sk_p])$ where $N(\unposed_B[sk_p])$ is the normal on the skeleton mesh at vertex $sk_p$. 
We fix $\shape_B = \hat{\shape}_B$ and find $(\hat{\vect{t_u}}, \hat{\vect{r_u}})$ that minimize $E_u(\vect{t},\vect{r}) =$ 
\begin{equation}
   \lambda_{sti} E_{sti} (\stitched(\hat{\shape}_B, \vect{t}, \vect{r})) + \lambda_d E_{d} (\unposed_S, \stitched(\hat{\shape}_B, \vect{t}, \vect{r}))
\label{eq:skel_unposing}
\end{equation}
to obtain $\unposed_B = \stitched(\hat{\shape}_B, \hat{\vect{t_u}}, \hat{\vect{r_u}})$ (see $\unposed_B$ in ~\figref{fig:pipeline}).

%% file: sections/method.tex
\section{Method -- OSSO}

Now that we have paired meshes $( \reg_S, \reg_B)$ and unposed $\unposed_B$ meshes,
we learn their correlations and how to predict skeleton landmarks $\tilde{\landmarks}_B$ 
from the skin vertices $\reg_S$ (\sectref{sec:pca_learning}).
Then, at test time, given an arbitrary STAR body shape in an arbitrary pose, we predict a skeleton mesh (\sectref{sec:inference})
and then repose it to match the input skin pose (\sectref{sec:reposing}). 
See \figref{fig:pipeline}.

\subsection{Skeleton statistics and correlation to skin shape}
\label{sec:pca_learning}

We next learn the correlation between the skin and the bones.
With the unposed skeletons $\unposed_B$, we first compute a low-dimensional linear subspace $\shapespace_{B}$, representing the skeleton shape variations using Principal Component Analysis (PCA).
We then learn a linear regressor $\regressor_{\beta}$ that predicts 
the skeleton shape space coefficients $\shape_B \in \shapespace_{B}$ from the STAR shape space coefficients $\shape_S$ computed in \sectref{sec:skin_fit}.

To properly constrain the 3D location of the skeleton inside the body, 
we define a new set of landmarks $\landmarks_B$,
composed of three landmarks per bone group.
We learn to infer them from the skin
with one linear regressor per landmark.
The regressor $\regressor_B$ takes as input the vertices of $\reg_S$
and predicts the 3D landmarks on $\reg_B$, i.e. $\landmarks_B(\reg_B)$.
We formulate the problem as a non-negative least squares problem and solve it with an active set method~\cite{lawson1995solving}.

\subsection{Inferring the skeleton from the skin}
\label{sec:inference}

We now have all elements to predict the skeleton shape from an input body surface in STAR format: $(\reg_S, \shape_S)$.
Using the learned regressor $\regressor_{\beta}$ (\sectref{sec:pca_learning}) we predict
the subject's skeleton shape $\shape_B = \regressor_{\beta}(\shape_S)$ from the body surface shape $\shape_S$.
Then, to properly position the skeleton inside the body,
we pose the body surface in a normalized lying pose $\pose_S^L$, obtaining $\reg_S(\pose_S^L)$
and predict 3D bone landmarks $\regressor_B(\reg_S(\pose_S^L)) = \tilde{\landmarks}_B$.
Let us write $\stitched(\shape_{B}, \vect{t}, \vect{r})$
to refer to the skeleton with shape $\shape_B$
posed with the {\it stitched puppet} pose parameters ($\vect{t}, \vect{r}$).
To obtain the bone poses $(\vect{t_0}, \vect{r_0})$ that match
the predicted landmarks, we minimize:
\begin{equation}
\begin{split}
    E(\vect{t}, \vect{r}) & = \lambda_L || \landmarks_B(\stitched(\shape_{B}, \vect{t}, \vect{r})) - \tilde{\landmarks}_B || \\
    & + \lambda_{ct} E_{ct}(\stitched(\shape_{B}, \vect{t}, \vect{r}), \reg_S(\pose_S^L)),
\end{split}
\label{eq:skel_inference}
\end{equation}
where $E_{ct}$ forces the contact between the skeleton and the skin, see Sup.~Mat. for the detailed definition.
The obtained mesh is our skeleton prediction.

\subsection{Reposing the inferred skeleton}
\label{sec:reposing}

For arbitrary poses, the simple stitching cost between bones can not properly model articulations like the knees and shoulders. 
We need a more precise anatomical model of the joints.
This could be very detailed (e.g.~OpenSim \cite{seth2018opensim}) but would complicate optimization.
Hence, we strike a balance between simplicity and realism and model two key articulations in more detail: ball joints and ligaments.

Ball joints like shoulders, elbows or hips should stay in their sockets.
For such joints, we identify sets of vertices on the skeleton that define the joint socket and the insertion bone head.
We fit spheres to them, and define an energy $E_j$ that forces the spheres to stay at a similar distance.

To replicate the ligaments of the human knee we define pairs of vertices at attachment points and an energy $E_l$ to constrain their distance. 
In Sup.~Mat.~we provide implementation details of $E_j$ and $E_l$.

Note that our articulation models, like all models, are an approximation to the truth and could be further refined for specific needs such as extreme bending poses.

Technically, given an inferred skeleton $\stitched_0 = \stitched( \shape_B, \vect{t_0}, \vect{r_0})$ inside the corresponding lying down skin mesh $\STAR_0 = \STAR(\pose_S^L, \shape_S)$, and a skin mesh in a specific pose $\STAR_{\theta} = \STAR(\pose_S^P, \shape_S)$, we pose the skeleton inside $\STAR_{\theta}$. 
We first compute the set of $\vect{d_p^0}$ offsets between $\stitched_0$ and $\STAR_0$ in the lying down pose 
and then minimize \eqnref{eq:skel_unposing} to position the skeleton inside the posed body $\STAR_{\theta}$.
Then, to enforce more realistic anatomic joints we add
\begin{equation}
\begin{split}
 E(\vect{t}, \vect{r})   =    \lambda_t E_j(\vect{t}, \vect{r}; \stitched_0) 
                             + \lambda_j E_l(\vect{t}, \vect{r}; \stitched_0) 
\end{split}
\label{eq:reposing_anat}
\end{equation}
to \eqnref{eq:skel_unposing} and optimize again.

%% file: sections/experiments.tex
\section{Experiments}

To evaluate our approach, we first quantify how accurately the skin registrations $\reg_S$ match the skin masks $M_S$ (\sectref{sec:eval_star_fits}).
We then evaluate how accurately our learned regressors predict the 3D bone landmarks from the skin (\sectref{sec:eval_3d_landmarks}).
Finally, we quantitatively and qualitatively evaluate how the projections of the computed bones overlap with the DXA bones masks (\sectref{sec:eval_bones_masks}).

In our experimental setting, for each gender, we use a training set of 1000 subjects and a test set of 200 subjects from the UK Biobank dataset~\cite{sudlow2015biobank}.
We made sure both sets have the same Body Mass Index distribution.
We compare OSSO with Anatomy Transfer (AT) \cite{ali2013anatomy} on the 
200 male and 200 female test subjects held out from any learning.

\subsection{STAR fits to DXA skin masks}
\label{sec:eval_star_fits}
We first evaluate how well our skin registrations $\reg_S$ overlap with the skin mask $M_S$
(\sectref{sec:skin_fit}) on the whole 2400 subjects dataset.
We compute the intersection over union metric, as ideally, all segmented skin pixels in $M_S$ should be covered by the projection of the skin registrations $\reg_S$.
We obtain a mean of 0.94 for the female subjects, 0.95 for the males,
with standard deviations below 0.01.

The small failure regions are due to soft tissue compression deformations of a lying down person that the STAR model does not capture.
Overall, the skin registrations faithfully capture the skin masks (see examples in Sup.~Mat.).

\subsection{Skin to 3D landmark regressors}
\label{sec:eval_3d_landmarks}

We next evaluate the accuracy of the regressors $\landmarks_B$ (\sectref{sec:pca_learning}) by predicting skeleton landmark locations from the body surface on the test set.
We train the regressors on the train set and we evaluate the 3D distance between the landmarks on the aligned skeleton $\landmarks_B(\reg_B)$ and the predicted landmarks $\regressor_B(\reg_S) = \tilde{\landmarks_B}$.
In Sup.~Mat., we present the table with the distances for all landmarks (male and female).
Our predictions have a mean distance (MD) below $1$ cm: 8.0 $\pm$ 6.1 mm for males and 8.4 $\pm$ 6.7 mm for females
and all individual landmarks results are consistent among male and female ($\pm 1 $mm).
The more accurate landmarks correspond to the upper skull (MD $< 2$ mm) and feet (MD $< 4$ mm),
whereas the least accurate belong to the hip iliac crest (MD $\approx 20$ mm).
We observe that the supervision of the bone masks $M_B$ is stronger in feet and skull than in the hip iliac crest, which is often not visible (see \figref{fig:skin_skel_segmentations}).

\begin{table}[t]\centering
\resizebox{0.8\columnwidth}{!}{
\begin{tabular}{l|cc|cc|}
\cline{2-5}
\multicolumn{1}{c|}{}                                & \multicolumn{1}{c|}{Male}        & Female      & \multicolumn{1}{c|}{Male}                    & Female                  \\ \hline
\multicolumn{1}{|l|}{Method}                         & \multicolumn{2}{c|}{$\cap_R (\%)$  $\uparrow$}  & \multicolumn{2}{c|}{HD (px) $\downarrow$}                                   \\ \hline
\multicolumn{1}{|l|}{$\reg_B$}                       & \multicolumn{1}{c|}{92}          & 94          & \multicolumn{1}{c|}{8.2 $\pm$ 2.6}           & 5.6 $\pm$ 1.7           \\ \hline \hline
\multicolumn{1}{|l|}{OSSO}                           & \multicolumn{1}{c|}{\textbf{88}} & \textbf{89} & \multicolumn{1}{c|}{\textbf{10.6 $\pm$ 3.2}} & \textbf{9.1 $\pm$ 2.3} \\ \hline
\multicolumn{1}{|l|}{AT$\sim$ \cite{ali2013anatomy}} & \multicolumn{1}{c|}{84}          & 88          & \multicolumn{1}{c|}{14.4 $\pm$ 2.9}          & 11.5 $\pm$ 3.1          \\ \hline
\end{tabular}
}
\caption{Quantitative comparison of OSSO and AT \cite{ali2013anatomy}. The $\cap_R$ score standard deviations are all below 2\%.}
\label{tab:eval_predicted_silhouettes}
\end{table}

\subsection{Evaluation on 2D DXA bones masks}
\label{sec:eval_bones_masks}

Next, we quantify how similar the predicted skeletons are to the subject's skeleton.
However, we only have access to 2D DXA bone images ($I_B, M_B$).
In addition, our DXA bone masks $M_B$ are coarse, as some bones, such as the hip bone are not completely segmented. 
To account for this,
we require every bone pixel in $M_B$
to be covered by the skeleton projection, but not the reverse.
Given a skeleton $\reg_B$ and a bone mask $M_B$ we compute their intersection ratio
${\cap}_R (\reg_B, M_B) = 100 |P(\reg_B) \cap M_B | / | M_B|$ as a percentage.
We also compute the directed Hausdorff Distance (HD) from $M_B$ to $P(\reg_B)$
accounting for the maximum pixel to pixel distance.

Table \ref{tab:eval_predicted_silhouettes} presents the results on the test set.
In the first row we evaluate the skeletons $\reg_B$ from \sectref{sec:skel_gloss_fit} to validate that they faithfully match the masks $M_B$. 
We obtain mean intersection percentages of $92\%$ and $94\%$ 
and mean HDs of $8.2$ and $5.6$ pixels for male and females, respectively.
OSSO obtains mean intersection percentages of $88\%$ and $89\%$ 
and mean HDs of $10.6$ and $9.1$ pixels,
while 
AT obtains mean intersection percentages of $84\%$ and $88\%$ 
and mean HDs of $14.4$ and $11.5$ pixels for male and females respectively.
Consistently, the OSSO predictions have higher mean intersection values and
lower HD than those of AT.

\begin{figure}
    \includegraphics[width=\columnwidth]{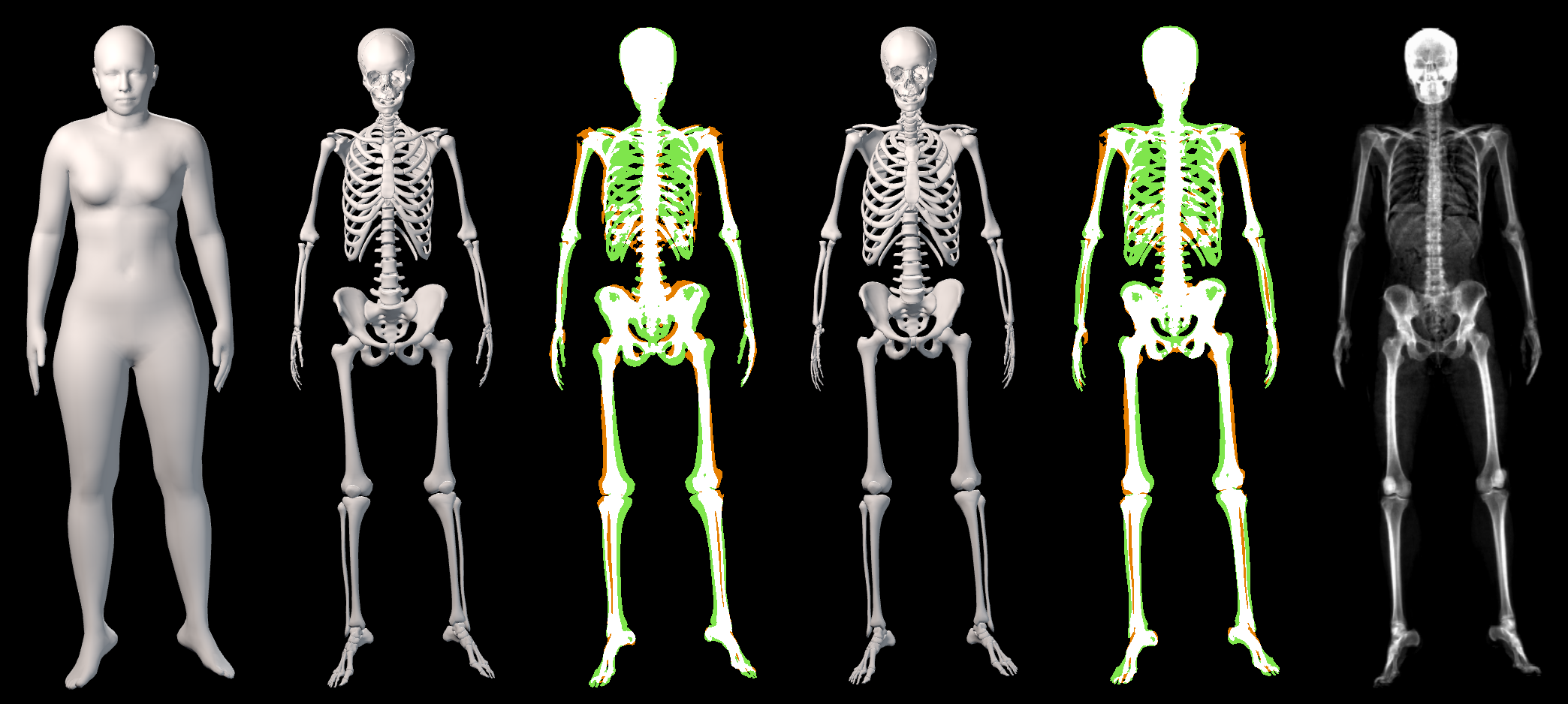}
    \includegraphics[width=\columnwidth]{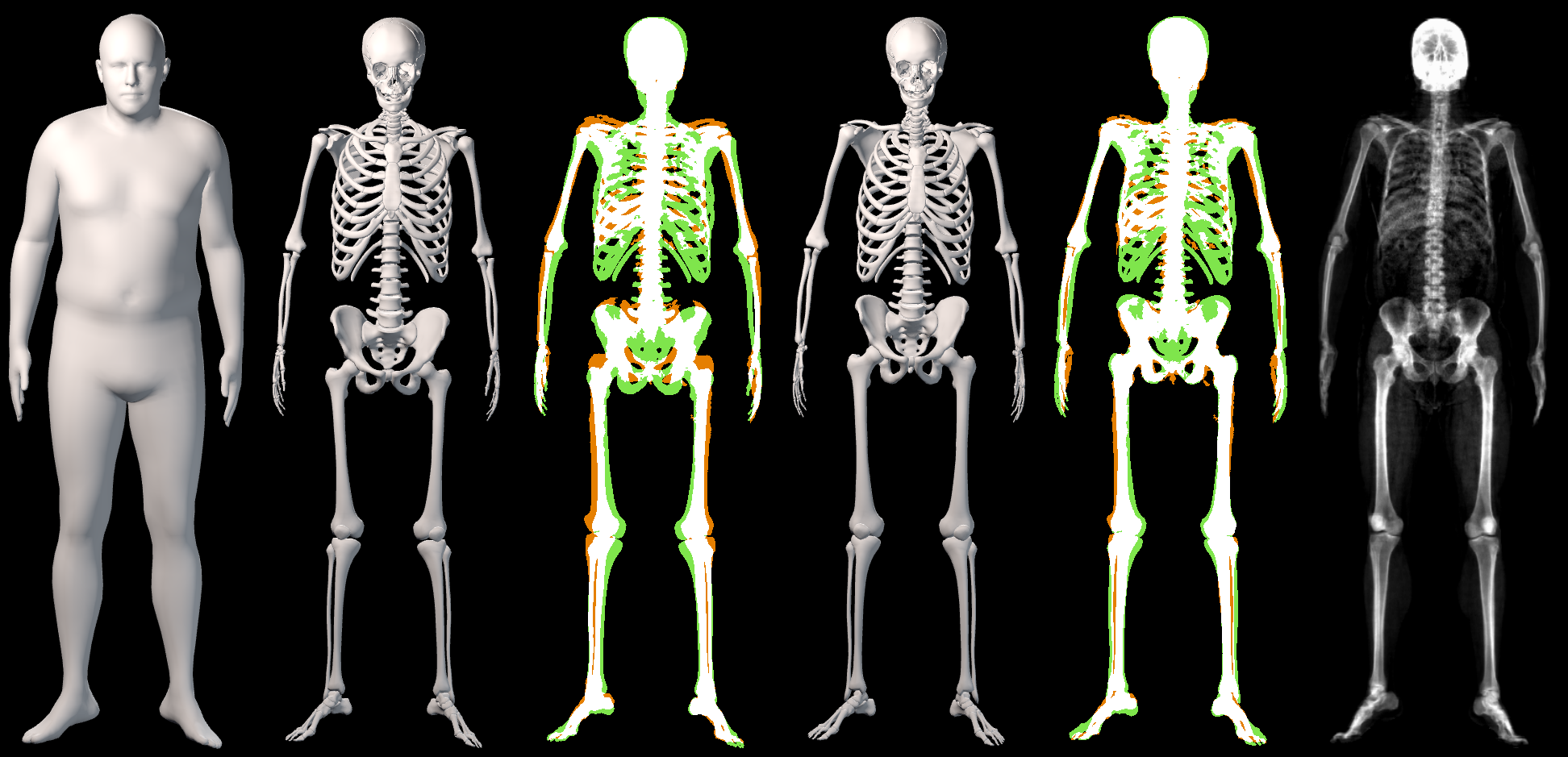}
    \vspace{-0.25in}
    \caption{From left to right: input $\reg_S$, AT prediction, overlap with $M_B$, OSSO prediction, overlap with $M_B$, $I_{B}$ DXA. Overlap image color code: orange is $M_B$ only (false negative), white is the intersection of both (true positive) and green predicted only.}
\label{fig:qualitative_results}
\end{figure}

The presented metric has a limitation: predicting all the skin volume as bone
would obtain a perfect result (${\cap}_R = 1$, $HD = 0$).
In \figref{fig:qualitative_results} and Sup.~Mat.~we show that
visually, OSSO's predictions are coherent and match the DXA bone images better than Anatomy Transfer.
In Sup.~Mat. we provide examples of  subjects with high Body Mass Index, for which Anatomy Transfer predicts a stretched skeleton, while ours are closer to the DXA skeleton mask.

\subsection{Generalization to new poses}
\begin{figure}
    \centering
    \includegraphics[width=0.24\columnwidth]{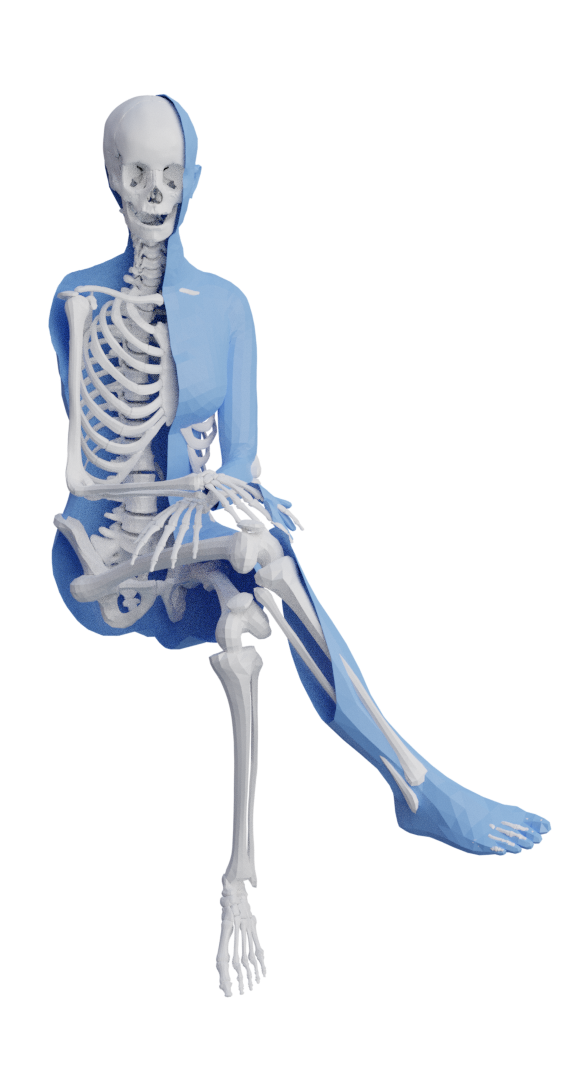}
    \includegraphics[clip, width=0.24\columnwidth]{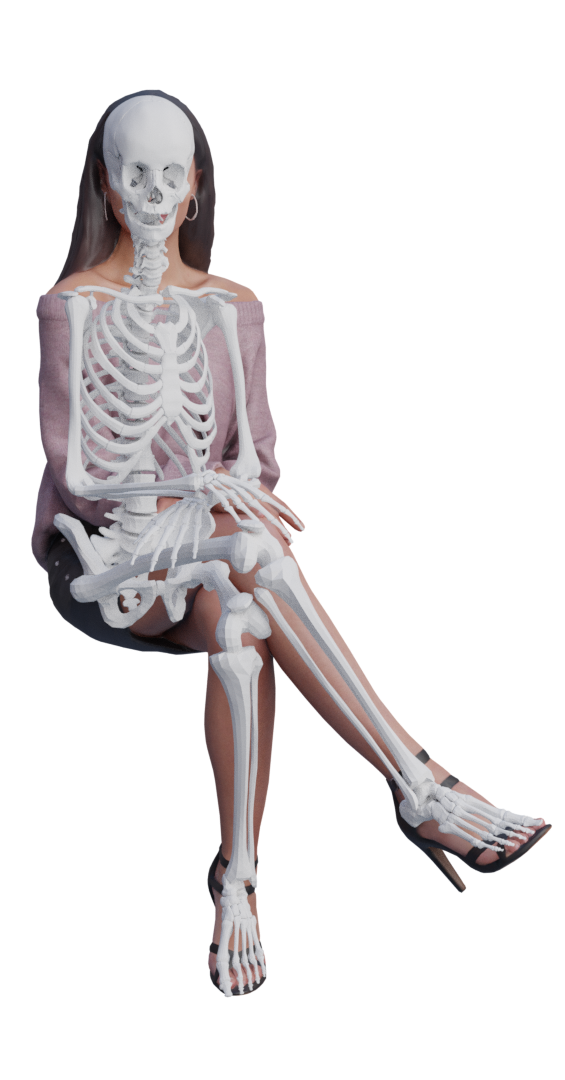}
    \includegraphics[clip, width=0.24\columnwidth]{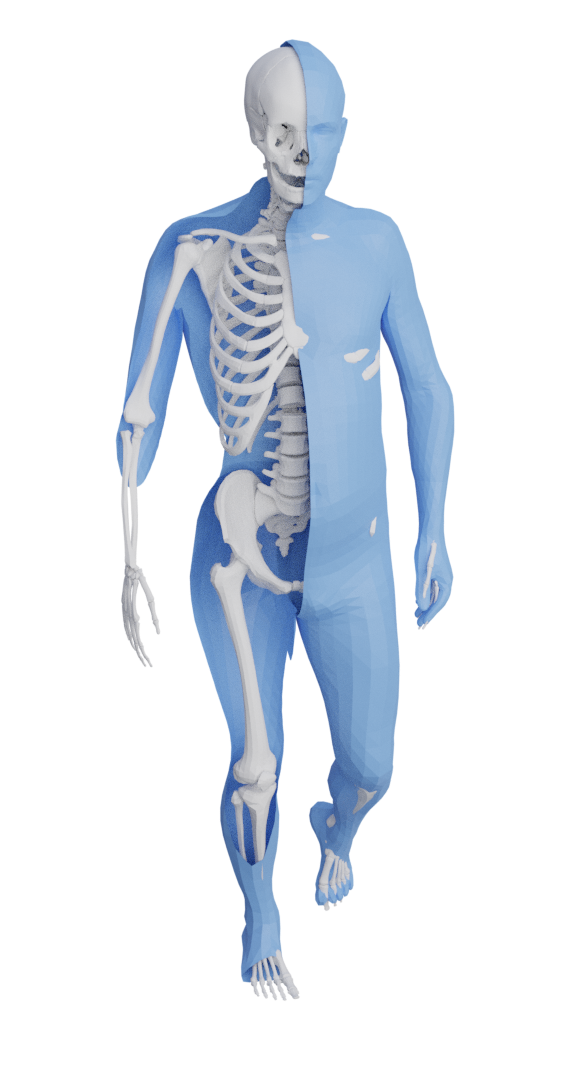}
    \includegraphics[clip, width=0.24\columnwidth]{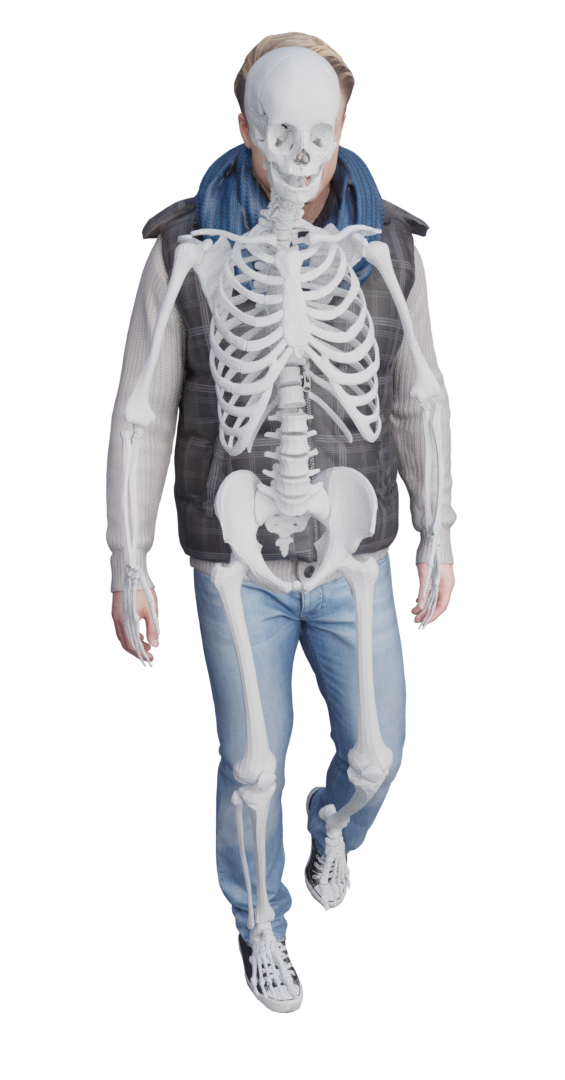}
    \vspace{-0.2in}
    \caption{Qualitative evaluation of the skeleton inference in arbitrary poses. OSSO yields visually plausible results.}
    \label{fig:new_poses_quali}
\end{figure}

Our regressors and model are learned from a limited set of poses, 
yet OSSO can predict skeletons from STAR bodies in arbitrary poses (\sectref{sec:reposing}).
We show several examples in \figref{fig:teaser}, \figref{fig:new_poses_quali}
and Sup.~Mat.
The clothed scans are from RenderPeople \cite{renderpeople} and are part of the AGORA dataset \cite{patel2021agora}, which includes high-quality SMPL fits to the scans taking into account clothing.
We fit STAR to the SMPL bodies (the templates have the same topology), and then apply OSSO to estimate the posed skeleton.
Unfortunately, we cannot quantitatively evaluate the accuracy of the posed skeletons.
Although some minor skin interpenetrations remain,
the obtained results are visually plausible.

%% file: sections/conclusion.tex
\section{Conclusion}

OSSO addresses the problem of predicting the skeleton of a person from their external body shape.
We use STAR~\cite{osman2020star} to represent the skin surface, and use a novel method to learn a parametric shape model of the anatomical skeleton using thousands of DXA scans.
We learn a mapping from the external body shape to the skeleton and can repose the skeleton inside the body subject to various constraints.
We evaluate OSSO using 2D DXA images from the UK Biobank dataset where the skin as well as the structure of the bones are visible.
Our skeletal predictions quantitatively outperform the state-of-the-art on silhouette reprojection error.
Qualitatively they are also better aligned with the DXA images.
To our knowledge, this is the first method to quantitatively validate the accuracy of a skeleton predicted from the body surface. 

\textbf{Limitations, Future Work, and Risks.}
OSSO predicts a person's skeleton from their body shape.
If the parametric body model does not accurately represent someone's shape, then the skeleton prediction is likely to be poor.
We use STAR instead of SMPL because it captures a broader range of body shapes, but it still has limitations.
It does not well represent bodies that are extremely thin, extremely obese, with scoliosis, the elderly or children, amputees, and transgender individuals.
In Sup.~Mat., we show failure cases for out of distribution bodies.

Our current skeleton and its joints, while more realistic than those of STAR, are still an approximation to the true human skeleton. 
For example, we use simplified models of the spine and lower arm.
We plan to add more anatomical detail to the model in future work.
One possible path would be to integrate our method with existing anatomical models like OpenSim \cite{seth2018opensim}.
Since models like STAR are not controlled by an anatomic skeleton, posed models may deviate from valid human shapes.
Future work should retrain models like STAR with our more realistic skeleton.
This could be done by using OSSO to infer the skeleton for 3D training scans in a variety of poses.
The blend weights and pose-corrective shapes of a STAR-like model could then be learned and controlled by the true skeleton.

A limitation of the present study is that our evaluation is only performed for lying down poses.
Ideally, we would like to train and test our method on complex upright poses.
Currently, this is not possible because there are no datasets that capture the inside and the outside of the body in various poses.
Possible technologies include full-body standing MRI scans or bi-plane fluoroscopy.
Both are rare, and the former has a limited range of motion, while the latter can only capture small regions (like the knee) and carries a significant X-ray exposure risk.

Our work is motivated by applications in medicine, biomechanics, sports science, etc.
Possible negative uses of the technology would involve capturing the skeletal data of a person (e.g.~from video) without their permission.
If future work shows that the skeleton is accurate enough to diagnose diseases like arthritis, the technology could be used, without consent, to learn about someone's risk of disease.
 
\textbf{Acknowledgments.}
This research has been conducted using the UK Biobank Resource under the Approved Project ID 51951.
Authors thank the International Max Planck Research School for Intelligent Systems for supporting MK.
MJB has received research gift funds from Adobe, Intel, Nvidia, Meta/Facebook, and Amazon.  MJB has financial interests in Amazon, Datagen Technologies, and Meshcapade GmbH. 
MJB’s research was performed solely at MPI.
SP's work was funded by the ANR SEMBA project. 
We thank Anatoscope for the initial skeleton mesh.

%% file: sections/appendix.tex
\begin{appendices}

\label{appendices}

In this supplementary material, we provide further details of our method and elaborate on the results presented in the main paper. Specifically:

In \sectref{sec:2d_ldm_prediction} we detail how we train a 2D landmark predictor from DXA silhouettes and quantitatively evaluate the accuracy of the 2D predicted landmarks on the synthetic data.
This section extends Sec. 3.2 of the main document.

In \sectref{sec:alignments}, we provide further details about the skin and skeleton registrations to the DXA images.
This section provides further details of Sec.s 3 and 4 of the main paper.

In \sectref{sec:pca} we present an evaluation of the skeleton shape space obtained in Sec. 4.1 of the main paper.

In \sectref{sec:qualitative_results} we provide quantitative and qualitative results to complement the Sec. 5 from the main document.

In \tabref{tab:notation_table} we summarize the notation used in the paper for an easy reference.

\input{sections/notation_table}

\section{Predicting 2D landmarks on DXA scans}
\label{sec:2d_ldm_prediction}

\begin{figure}[H]
    \centering
    \includegraphics[width=0.85\columnwidth]{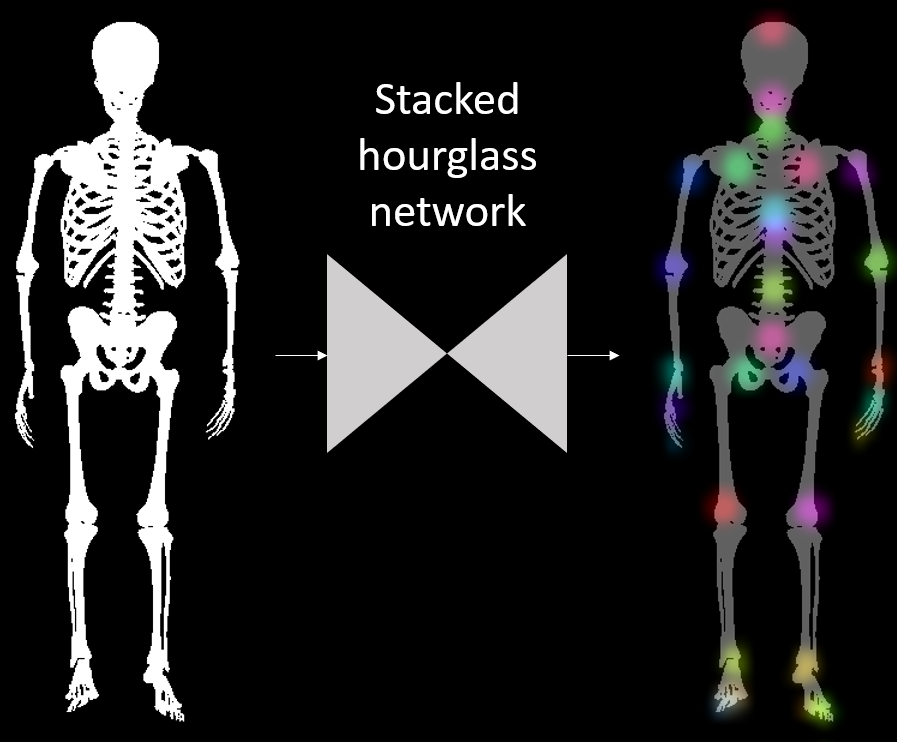}
    \caption{From a skeleton mask, a stacked hourglass network predicts the 2D locations of the landmarks $\tilde{\landmarks}_I$.}
    \label{fig:2d_landmarks_prediction_recap}
\end{figure}

In order to register the skin and skeleton models to the DXA scans, we need 2D landmarks on the scans. In this section we explain how we generate the synthetic dataset (\sectref{sec:init_model}, \sectref{sec:synth_mask}) to train a 2D landmark predictor from DXA skeleton silhouettes (\sectref{sec:2d_ldm_training}) and evaluate the prediction (\sectref{sec:2d_ldm_evaluation}).
The 2D landmark prediction from DXA silhouette is illustrated in  \figref{fig:2d_landmarks_prediction_recap}.

\subsection{Initial model creation}
\label{sec:init_model}

To generate synthetic skeleton silhouettes that look similar to real DXA bone masks $M_B$, we create an articulated skeleton model $\skelinit$, rigged with the STAR body model  \cite{osman2020star} parameters.

We first generate 21 STAR bodies by sampling the STAR shape space $\shapespace_S$.
We consider the mean body, and then, for the $n_\beta=10$ first components of the STAR shape space, we sample two new body shapes with the shape parameters $\shape=\{-2, 2\}$. 
Using Anatomy Transfer (AT)~\cite{ali2013anatomy}, we register a template skeleton mesh to each of these body shapes.
Effectively we enforce the skin of the AT mesh to match the STAR mesh.

With the obtained registrations, we define the mean skeleton shape $\skelinit(\beta=0, \theta=0)$, as the obtained AT skeleton on  STAR's mean shape. 
Then, for each shape space component, we compute the skeleton offsets to the mean skeleton and use these offsets to define an initial skeleton shape space.
From these, we compute the shape vectors of $\bones$ as $\shapespace_i = \vt_{\shape_i=2} -\vt_{\shape_i=-2} $  for i in $[0, n_\beta]$, else $\shapespace_i  = \mathbf{0}$ .

To pose the skeleton, we rig it with the same kinematic tree as STAR. For each skeleton bone we manually define to which body part it belongs.
This is straightforward as the initial template skeleton has the individual bones identified.
It is important to note that the created skeleton model $\skelinit(\beta, \theta)$ can change its shape and pose
using the same shape and pose parameters as STAR.

This initial model has an obvious drawback:
the kinematic joint locations are not consistent with the anatomic skeleton articulations. 
Still, it is sufficient to easily generate plausible synthetic bone masks and the corresponding landmark annotations.

\begin{figure}[H]
    \centering
    \includegraphics[width=1\columnwidth]{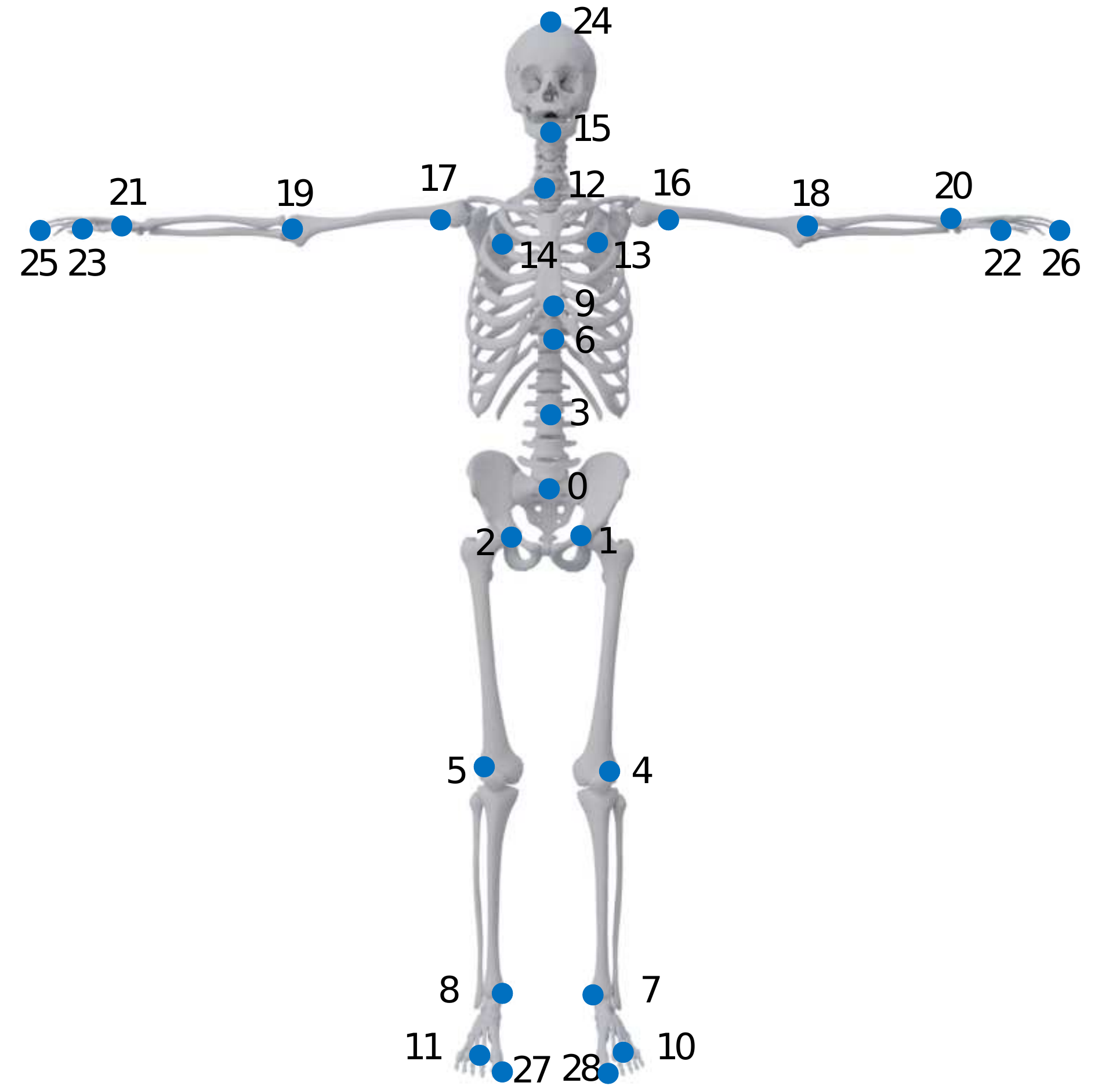}
    \caption{Position of the 3D landmarks $\landmarks_I$ on the Stitched Puppet skeleton model $P_B$. These markers correspond to the location of the STAR 3D joints plus 5 additional landmarks.}
    \label{fig:L_i_landmarks}
\end{figure}

We define 29 landmarks on the skeleton mesh (\figref{fig:L_i_landmarks}). 
The first 24 correspond to the closest vertex to the STAR joint locations.
Additionally we select the tip of the head, fingers and feet.
We denote these initial landmarks $\landmarks_I$
or $\landmarks_I(\mat{M})$
if we make explicit the mesh $\mat{M}$ on which the landmarks are defined.

\subsection{Generating synthetic DXA masks}
\label{sec:synth_mask}

We use the skeleton model $\skelinit $ to generate synthetic skeleton binary masks $\hat{M}_B$ with their corresponding 2D landmarks, that we denote $\tilde{\landmarks}_I$ to explicitly distinguish them from the 3D landmarks $\landmarks_I$.

We generate synthetic skeleton shapes by uniformly sampling the STAR shape space $\shape$ in the range $[-2.5, 2.5]^{10}$. 
As the poses in DXA scans are relatively constrained, i.e.~lying down with arms at the side, we manually define a {\it lying pose} $\pose_L$ 
and sample new angles from a uniform distribution centered at $\pose_L$ within a small range.

\begin{figure}[H]
     \begin{mdframed}[backgroundcolor=black]
     \includegraphics[width=0.159\columnwidth]{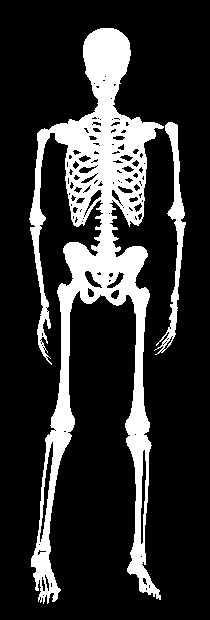}
     \includegraphics[width=0.159\columnwidth]{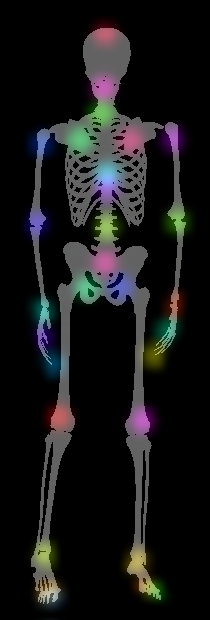}
     \includegraphics[width=0.159\columnwidth]{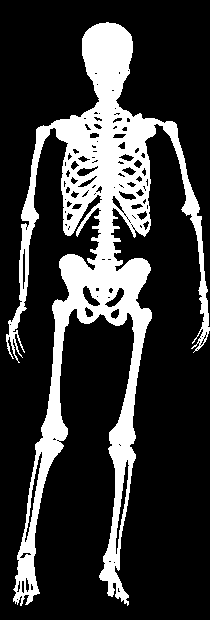}
     \includegraphics[width=0.159\columnwidth]{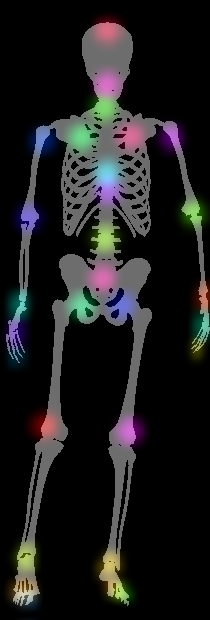}
     \includegraphics[width=0.159\columnwidth]{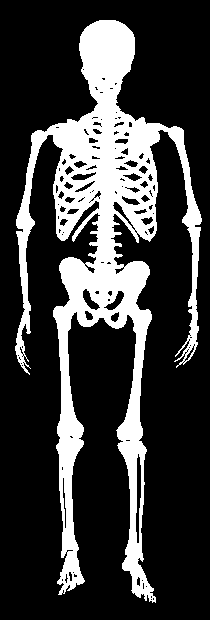}
     \includegraphics[width=0.159\columnwidth]{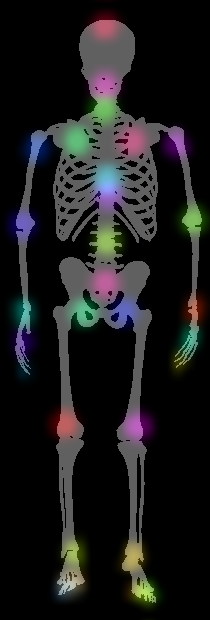} \\
     \end{mdframed}
     \vspace{-0.15in}
    \caption{Pairs of synthetic skeleton masks (in white) and 2D landmarks $\tilde{\landmarks}_I$ (color-coded) overlayed on the mask (in gray).
}
\label{fig:2d_ldm_training_data}
\end{figure}

With the sampled shape and pose parameters, we render the silhouette of the skeleton and the corresponding landmark image.
The virtual camera is orthographic to match the DXA scanner camera, and the field of view is set depending on the sample body height to leave a specific margin on the top and bottom of the image. This margin is sampled to match the margin distribution observed on the DXA dataset.  A sample of the generated paired data is presented in \figref{fig:2d_ldm_training_data}. 

To bridge the domain gap between the synthetic silhouettes and the DXA ones, 
we augment the data by eroding, and partially masking the rendered skeleton silhouettes, while keeping the landmarks fixed.

\subsection{Training a 2D landmarks predictor}
\label{sec:2d_ldm_training}

From the synthetic silhouettes of the skeleton $\hat{M}_B$, we train the landmark detector using a stacked hourglass network~\cite{newell2016stacked} with 8 stacks. The network takes a 256x256 binary silhouette as input and outputs a 29x64x64 tensor, where each channel contains the position for one of the 29 landmarks $\tilde{\landmarks}_I$. 

\begin{figure}[H]
    \begin{mdframed}[backgroundcolor=black]
        \includegraphics[width=0.159\columnwidth]{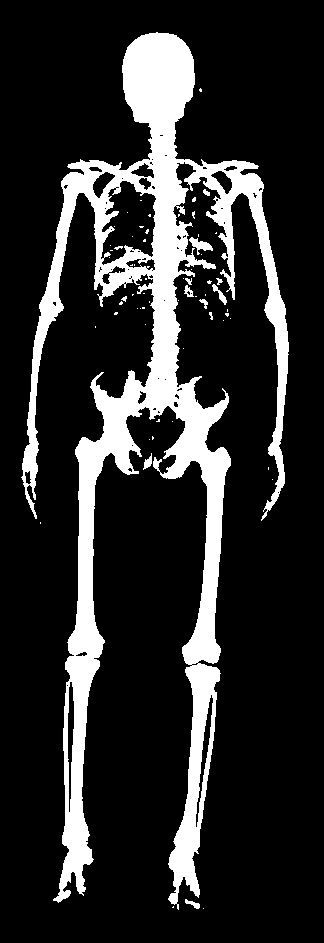}
        \includegraphics[width=0.159\columnwidth]{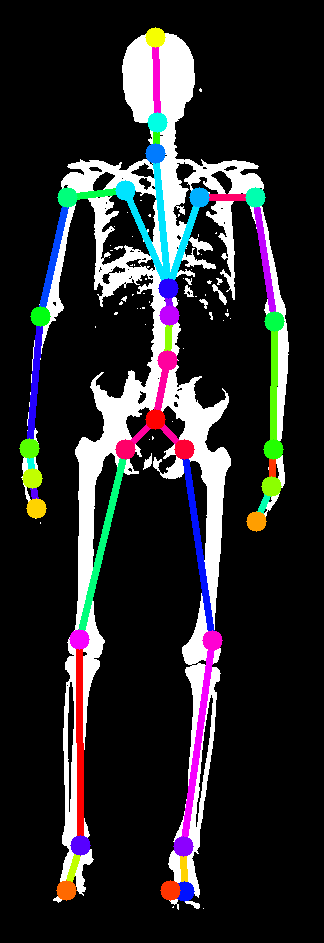}
        \includegraphics[width=0.159\columnwidth]{figures/subjects/6_dxa_skel_mask.png}
        \includegraphics[width=0.159\columnwidth]{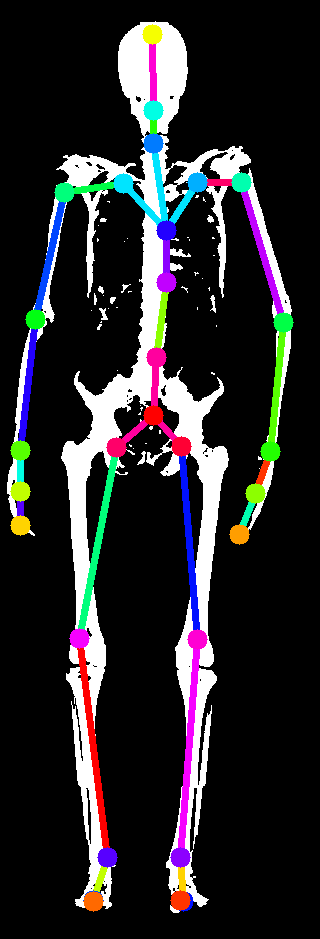}
        \includegraphics[width=0.159\columnwidth]{figures/subjects/9_dxa_skel_mask.png}
        \includegraphics[width=0.159\columnwidth]{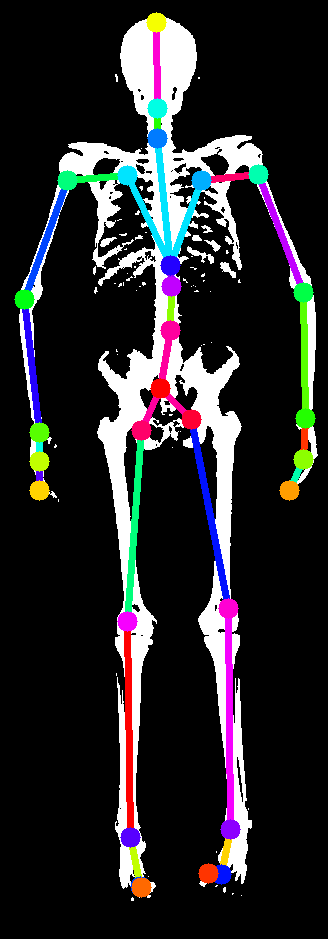}
    \end{mdframed}
    \vspace{-0.15in}
    \caption{Pairs of input and predicted 2D landmarks $\tilde{\landmarks}_I$ on real DXAs. The network learned on synthetic data generalizes well to real data.}
\label{fig:2d_ldm_inference}
\vspace{-0.1in}
\end{figure}

In \figref{fig:2d_ldm_inference}, we show qualitative results
of the predicted landmarks on binary masks from real DXA images.
We visually inspected the predicted 2D landmarks and observe that the silhouette simplification strategy combined with our data augmentation technique allows to obtain very good qualitative results on real DXA images.

\subsection{2D landmarks prediction evaluation}
\label{sec:2d_ldm_evaluation}

\begin{table}
\centering
          \begin{tabular}[t]{|c|c|}
              \hline
              &err. (mean $\pm$ std)\\ 
              \hline
                L0 & \textcolor{green}{0.73} $\pm$ 0.35 \\  L1 & 0.95 $\pm$ 0.40 \\  L2 & 0.81 $\pm$ 0.38 \\  L3 & 0.90 $\pm$ 0.46 \\  L4 & 1.14 $\pm$ 0.54 \\  L5 & 1.12 $\pm$ 0.60 \\  L6 & 0.78 $\pm$ 0.46 \\  L7 & 1.16 $\pm$ 0.63 \\  L8 & 1.24 $\pm$ 0.68 \\  L9 & 1.07 $\pm$ 0.37 \\  L10 & 1.18 $\pm$ 0.52 \\  L11 & 1.18 $\pm$ 0.62 \\  L12 & 0.87 $\pm$ 0.41 \\  L13 & 0.87 $\pm$ 0.41 \\  L14 & 1.01 $\pm$ 0.43 \\
              \hline
          \end{tabular}
          \begin{tabular}[t]{|c|c|}
              \hline
              &err. (mean $\pm$ std)\\ 
              \hline
                L15 & 0.78 $\pm$ 0.37 \\  L16 & 1.01 $\pm$ 0.47 \\  L17 & 0.87 $\pm$ 0.50 \\  L18 & 1.22 $\pm$ 0.62 \\  L19 & 1.01 $\pm$ 0.54 \\  L20 & 1.22 $\pm$ 0.69 \\  L21 & 1.21 $\pm$ 0.56 \\  L22 & 1.08 $\pm$ 0.75 \\  L23 & 1.04 $\pm$ 0.69 \\  L24 & 0.75 $\pm$ 0.43 \\  L25 & \textcolor{red}{1.87} $\pm$ 1.39 \\  L26 & 1.53 $\pm$ 1.02 \\  L27 & 1.23 $\pm$ 0.61 \\  L28 & 1.23 $\pm$ 0.67 \\
              \hline
          \end{tabular}
          \vspace{-0.1in}
\caption{Prediction error in pixels of the predicted 2D landmark $\tilde{\landmarks}_I$ on synthetic skeleton silhouettes.
Landmark numbers are visually shown on the mesh in ~\figref{fig:L_i_landmarks}.}
\label{tab:eval_2D_landmark_error}
\end{table}

As the original DXA images do not have annotations, we only evaluate quantitatively on the synthetic dataset. 
We evaluate the landmarks predicted by the stacked hourglass network on 100 unseen synthetic skeleton silhouettes. 
The prediction error is measured in pixels on an image of size 256x256 pixels. The per landmark errors are reported in  Table \ref{tab:eval_2D_landmark_error}.

Most errors are on the order of one pixel.
The highest prediction errors are for the tip of the middle fingers (L25 and L26) and the toes (L27 and L28). We observe that due to the resizing of the skeleton mask from the original image size (approx 800x800) to the size of the network (256x256), fine structures such as fingers and toes are degraded or lost. This is numerically visible with the standard deviations of the finger markers which are over 1 pixel. 

\section{Skin and skeleton registrations to DXA}
\label{sec:alignments}

This section provides further details to complement the sections 3.3, 3.4 and 3.5 of the main paper.

\subsection{Skeleton model based on Stitched Puppet}

We create a parametric skeleton model to align to the DXA skeleton silhouettes based on the {\it stitched puppet}~\cite{zuffi2015stitched}.

The {\it stitched puppet} model, as the name implies, represents an articulated deformable structure, the human body, as a collection of parts that are stitched together at the part interfaces. The model has per-part shape spaces and a pose parametrization in terms of location of each part center and its global rotation. The {\it stitched puppet} can be seen as a graphical model, where part parameters are defined at each node, and edge potentials represent stitching costs, that favor the parts to be connected and have smooth skin connections. The original model ~\cite{zuffi2015stitched} is fit to 3D scans of people with non-parametric particle belief propagation. In order to define a stitched puppet model given an existing mesh, one needs to define a segmentation of the faces into parts, duplicate the vertices that belong to different adjacent parts, and define stitching potentials that act as springs between the corresponding duplicated vertices.

\begin{figure*}
    \centering
    \includegraphics[width=0.32\textwidth]{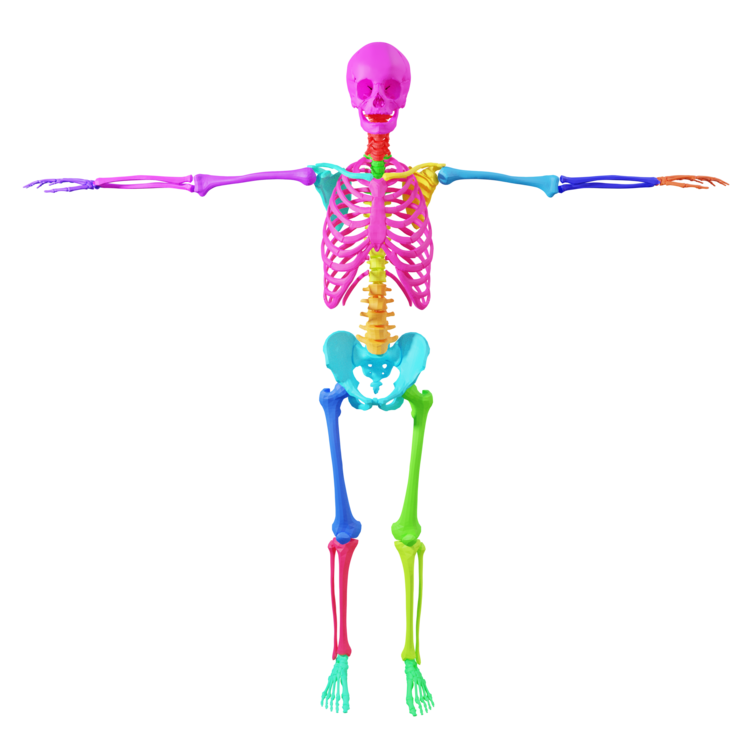}
    \includegraphics[width=0.3\textwidth]{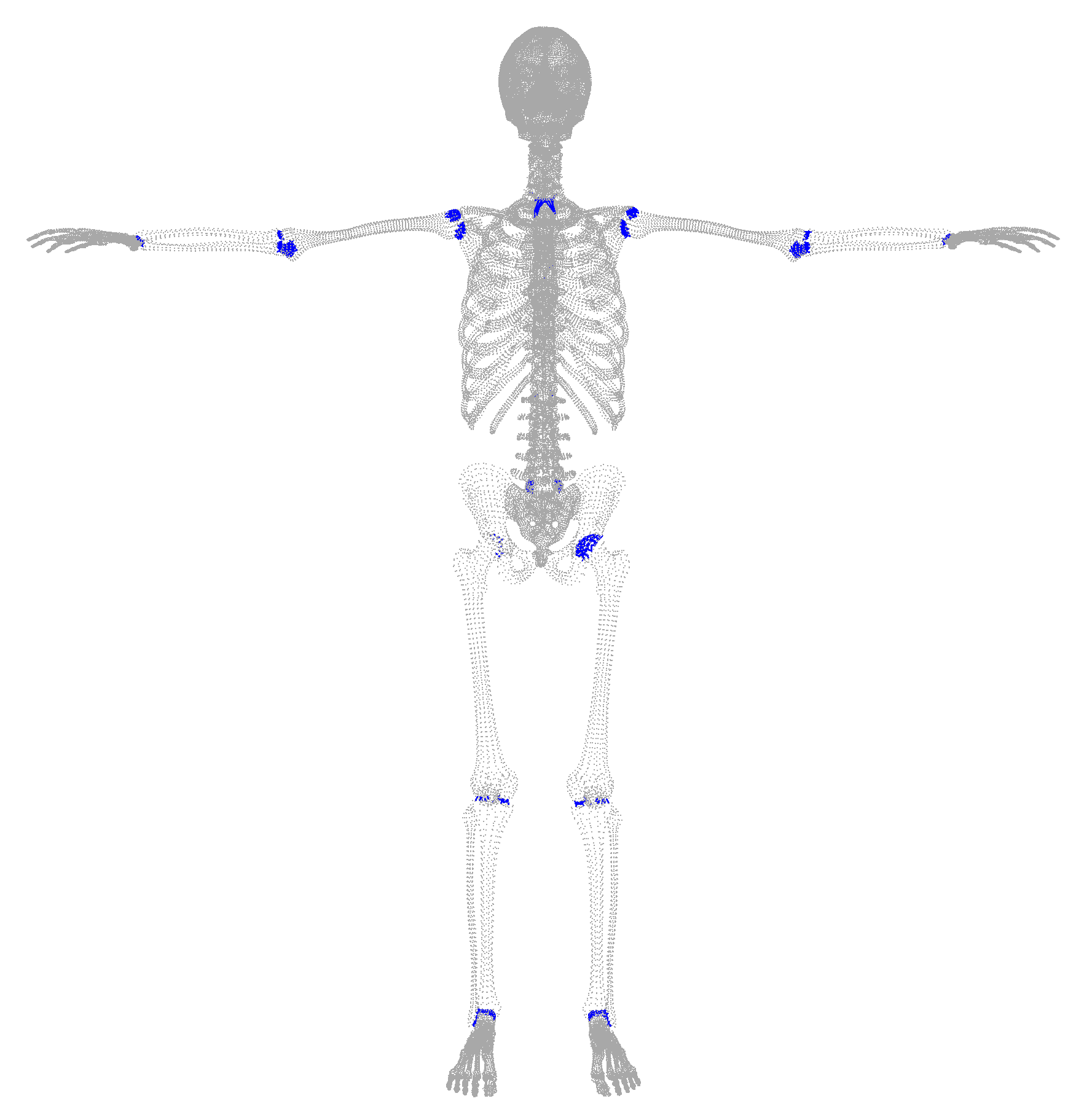}
    \includegraphics[width=0.32\textwidth]{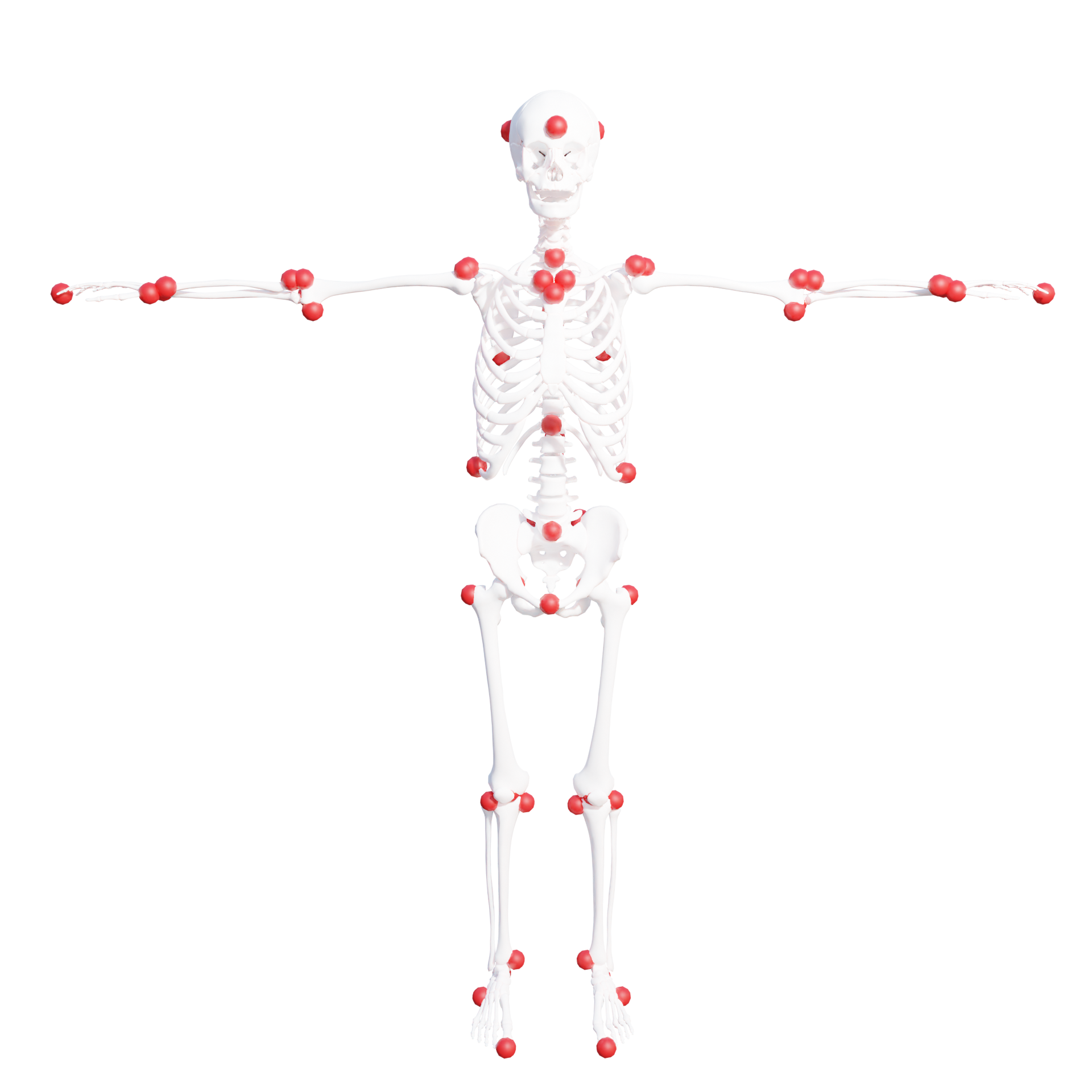}
    \caption{Our \textit{stitched puppet} skeleton model, with the different bone groups (left), the interface point between the groups (center) and the 3D landmarks $\landmarks_B$ (right).}
    \label{fig:stitched_puppet}
\end{figure*}

In our skeleton model, we manually define 21 groups of bones that belong to the same anatomic part, and define the interfaces between these parts.
In \figref{fig:stitched_puppet} we show the different parts with color codes, their interfaces, as well as the 3D landmarks $\landmarks_B$ defined on the bones.

\subsection{Registration costs}

In this section, we detail the costs used for the skeleton registration (Sec 3.5 of the main paper) and the final reposing (Sec 4.3 of the main paper). In this section, we denote the vertices of $\stitched$ as $v_{sp}$, the vertices of $\STAR$ as $v_{st}$ and $z$ the anterior-posterior axis.  $v^z$ denotes the $z$ component of vertex $v$ and $v^n$ the mesh normal at this vertex.

\paragraph{Skeleton to DXA registration.}

\begin{figure}[H]
    \centering
    \includegraphics[width=1\columnwidth]{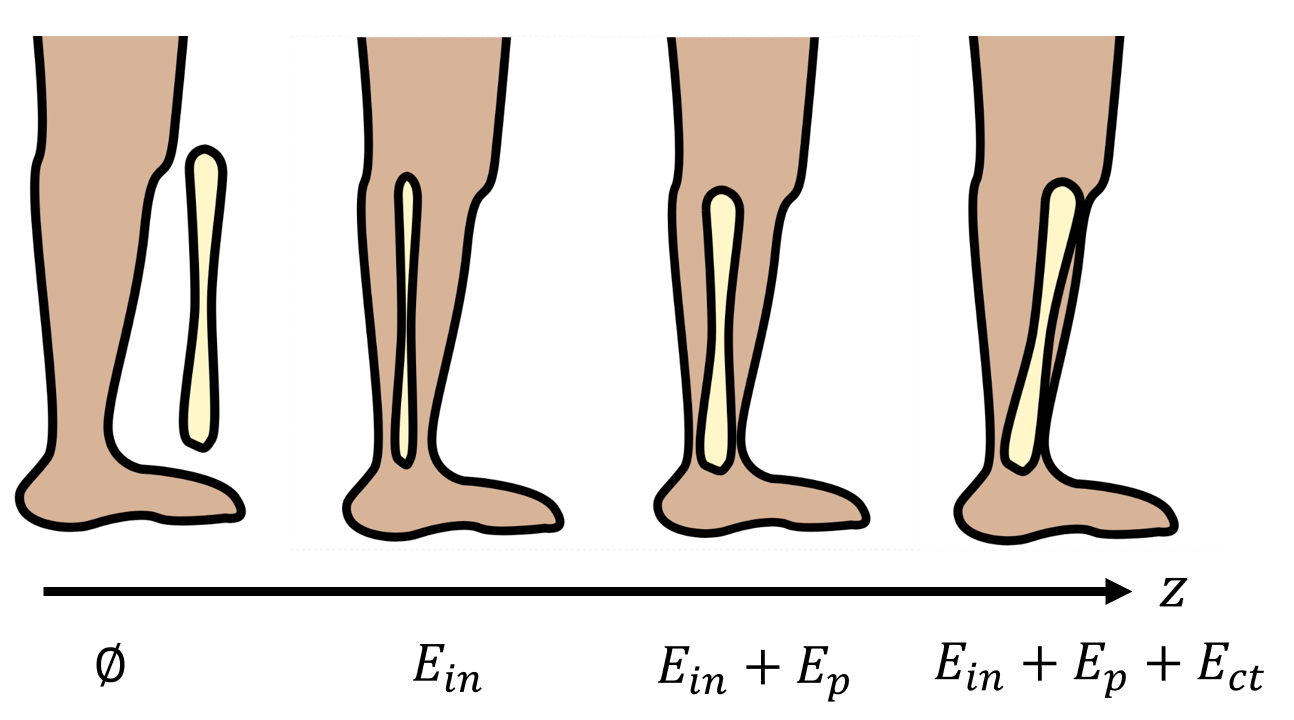}
    \caption{We illustrate the intuition behind the costs on a profile view of the tibia in the leg. From the frontal projected silhouette, there is no constraint for the bone to be inside the body along the z axis. We use $E_{in}$ to force it to be inside. Forcing it inside is not enough as it could squeeze and collapse; thus, we enforce the bone to be close to the skin surface with $E_{p}$. In addition, there are regions where the bones are not covered by muscle and fat and should, therefore, lie close to the skin surface.  We use $E_{ct}$ to enforce these manually defined areas of contact.}
    \label{fig:leg_diagram}
\end{figure}

In Sec. 3.5 of the main paper, we introduce the cost $E_{i}$ to constrain the skeleton inside the body. 
We decompose $E_{i}$ as $E_{i} = E_{in} + E_{p} + E_{ct}$ and illustrate the intuition of each cost in \figref{fig:leg_diagram}.

The energy term $E_{in}$ forces the skeleton to be inside the body along the front-back axis.

\begin{equation}
E_{in} = \max(0, D_z(\stitched(\shape_{B}, \vect{t}, \vect{r}), \reg_S)) 
\end{equation}

where $D_z$ is the distance along $z$ between a $\stitched$ vertex and the closest skin vertex.

The term $E_{p}$ forces vertices of the skeleton to be close to specific areas of the skin along the front-back axis.
For several manually defined pairs of skeleton vertices and skin area $A$, we define

\begin{equation}
E_{p} = v_{sp}^z - \sum_{v_{st} \in A}(v_{st}^z).
\end{equation}

The energy $E_{ct}$ forces the {\it contact} between some specific vertices of the skeleton and the skin, like the elbow or the finger tips. 

We define pairs of skin and skeleton vertices $(v_{sp}, v_{st})$ and want them to be at a fixed small distance $e=5mm$.
Effectively, $E_{ct}$ is the per vertex distance: 
\begin{equation}
E_{ct} = v_{sp} - (v_{st} - e \cdot v_{st}^n)
\end{equation}

\paragraph{Skeleton unposing.}

\begin{figure}[H]
    \centering
    \includegraphics[width=1\columnwidth]{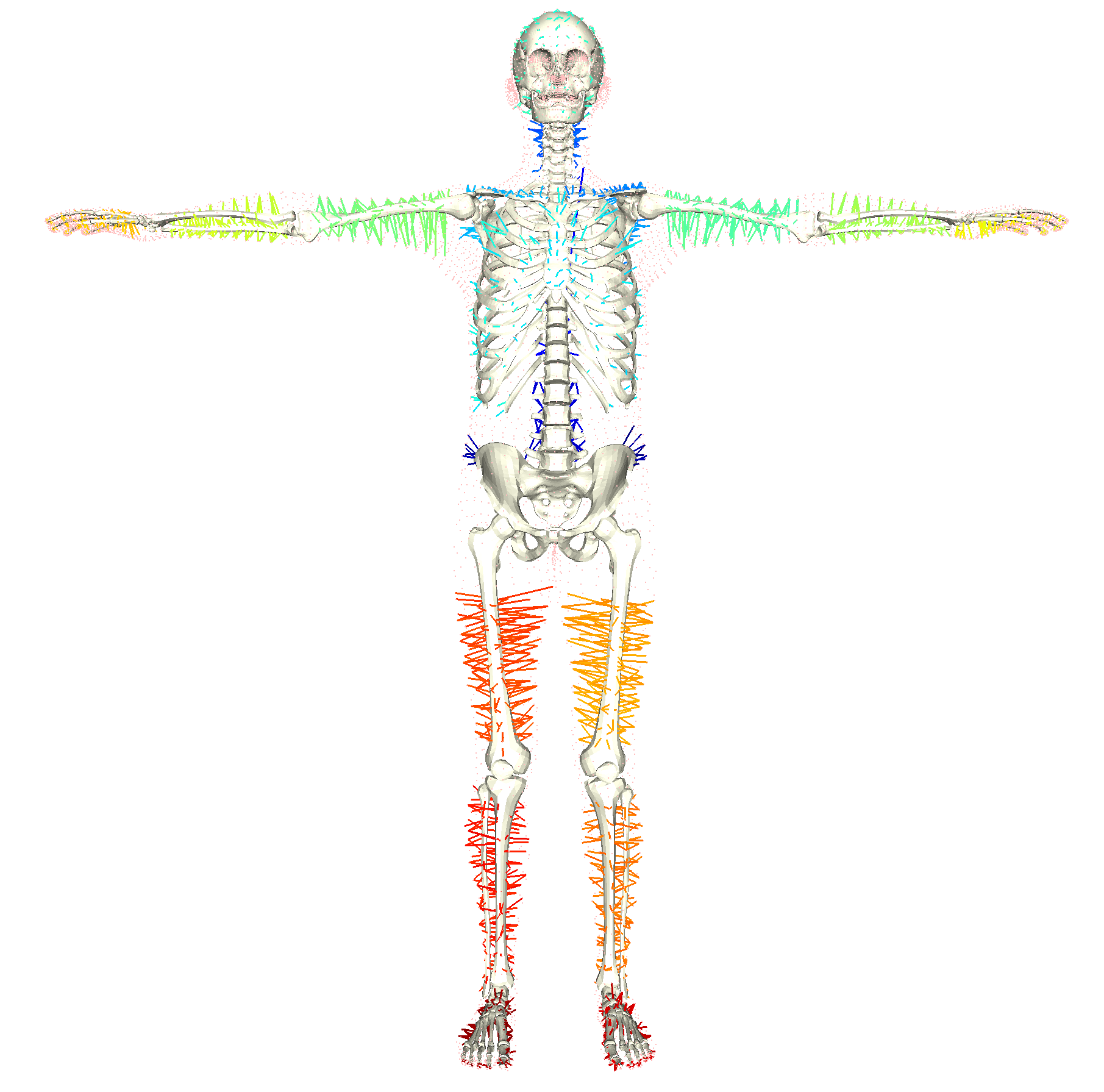}
    \vspace{-0.1in}
    \caption{Skin to skeleton pairs used in the cost $E_d$. We color the links in each part with a different color for visualization purposes.}
    \label{fig:skin_springs}
\end{figure}

In Sec. 3.5 of the main paper, we introduce $E_d$, a cost that enforces the conservation of the skeleton to skin distance when changing the pose. In \figref{fig:skin_springs} we illustrate the pairs of skin and skeleton vertices that are used for this cost. Our heuristic is that each of these pairs has a fixed distance $d_0$ that should be constant independent of the 3D pose.

\paragraph{Skeleton reposing.}

In Sec. 4.3 of the main paper, we use the costs $E_j$ and $E_l$ in the skeleton inference optimization.

The term $E_j$ models ball joints in the shoulders, elbows and hips. It forces the bone heads to stay in their sockets.
For each articulation, we define vertices ${s_i}, {s_j}$ on the skeleton template that define a joint socket of a bone head. At each optimization step, we fit spheres with centers $S_i, S_j$ to each groups of vertex and force each of spheres to stay at a similar distance during the optimization:  
\begin{equation}
E_j(\vect{t}, \vect{r}; \stitched_0) = ||S_i(\vect{t}, \vect{r}) - S_j(\vect{t}, \vect{r})|| - d_{s0}
\end{equation}

This cost is not sufficient to model the knee movement, so we define stitching costs approximating the human knee ligaments. 
We create pairs of vertices $(l_i, l_j)$ at the bone locations where the ligaments are attached, and define the per-vertex cost $E_l = ||l_i - l_j|| - d_{l_0}$.

The distances $d_{l0}$ and  $d_{s0}$ are defined such that $E_j(\vect{t_0}, \vect{r_0}; \stitched_0)=0$ and $E_l(\vect{t_0}, \vect{r_0}; \stitched_0)=0$.

\section{Skeleton shape space evaluation}
\label{sec:pca}
In section 3.6 of the main  paper, we detail how we learn a skeleton shape space from the unposed skeleton meshes. In this section, we present an evaluation of the compactness of the shape space as well as its generalization ability.

\subsection{Variance}
\begin{figure}[H]
    \centering
    \includegraphics[width=1\columnwidth]{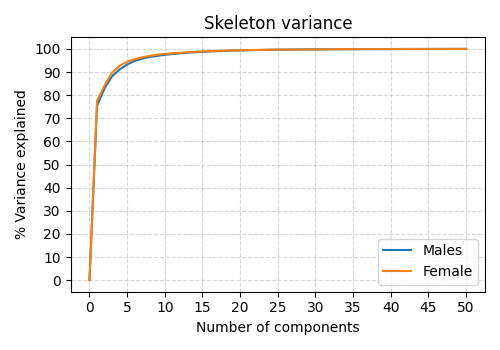}
    \caption{Cumulative variance of the skeleton shape space.}
    \label{fig:pca_variance}
\end{figure}

To evaluate the compactness of our skeleton shape space, we compute the variance explained by each component of the PCA space. 
The cumulative variance plot is shown \figref{fig:pca_variance}. With 3, 5 and 10 components, the male PCA model respectively captures 91.1\%, 94.8\% and 97.8\% of the skeleton's variance. The female model respectively 92.7\%, 95.6\% and 98.1\%.

\subsection{Shape space generalisation}

We next evaluate how the skeleton shape space generalises to unseen subjects.
We compute the skeleton shape space from the training dataset and we evaluate how accurately it can reconstruct 200 left out unposed skeletons.
We project each of the test set meshes onto the first $N$ basis vectors of the shape space
and we reconstruct the bones using only these coefficients.

We then measure how much information is lost in this projection by computing the per-vertex distance between the original mesh and the projected and reconstructed mesh. We aggregate this per-vertex error for each mesh and obtain the errors reported in Table \ref{tab:pca_reconstruction_error}.

As we can see, with a small number of components, such as 5, mean errors are below 6 mm. When  using 10 components, the reconstruction mean errors are below 4 mm. The created bones shape space can capture the  shape of left out subjects with errors below 4 millimeters.

\begin{table}
\begin{tabular}{c|cc|}
\cline{2-3}
\multicolumn{1}{l|}{}               & \multicolumn{2}{c|}{error (mm) (mean $\pm$ std)}                        \\ \hline
\multicolumn{1}{|c|}{Nb components} & \multicolumn{1}{c|}{Male}            & \multicolumn{1}{c|}{Female} \\ \hline
\multicolumn{1}{|c|}{3}             & \multicolumn{1}{c|}{7.59 $\pm$ 4.79} & 7.79 $\pm$ 4.86             \\
\multicolumn{1}{|c|}{5}             & \multicolumn{1}{c|}{5.55 $\pm$ 3.49} & 5.14 $\pm$ 3.27             \\
\multicolumn{1}{|c|}{10}            & \multicolumn{1}{c|}{3.14 $\pm$ 2.19} & 3.02 $\pm$ 2.14             \\ \hline
\end{tabular}
\vspace{-0.1in}
\caption{Skeleton reconstruction error given the number of principal components used. The errors are in millimeters.}
\label{tab:pca_reconstruction_error}
\end{table}

\section{Extended results}
\label{sec:qualitative_results}

This section complements the presented results in Sec. 5 of the main document.

\subsection{Skin alignment qualitative evaluation}
\begin{figure}[!]
    \centering
    \begin{mdframed}[backgroundcolor=black, align=center]
    \includegraphics[width=0.495\columnwidth]{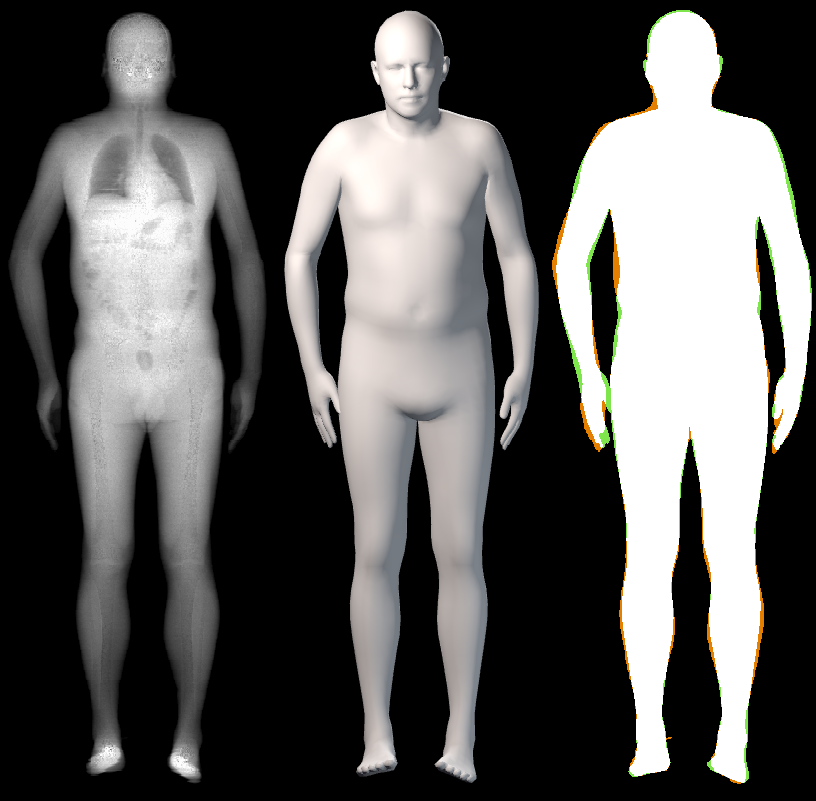}
    \includegraphics[width=0.495\columnwidth]{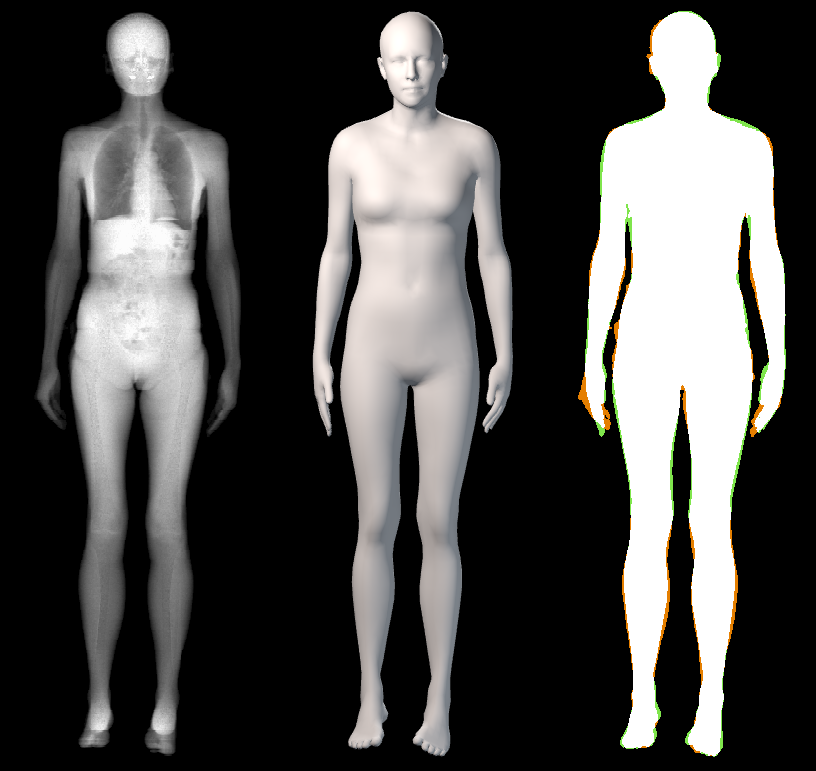}
    \includegraphics[width=0.495\columnwidth]{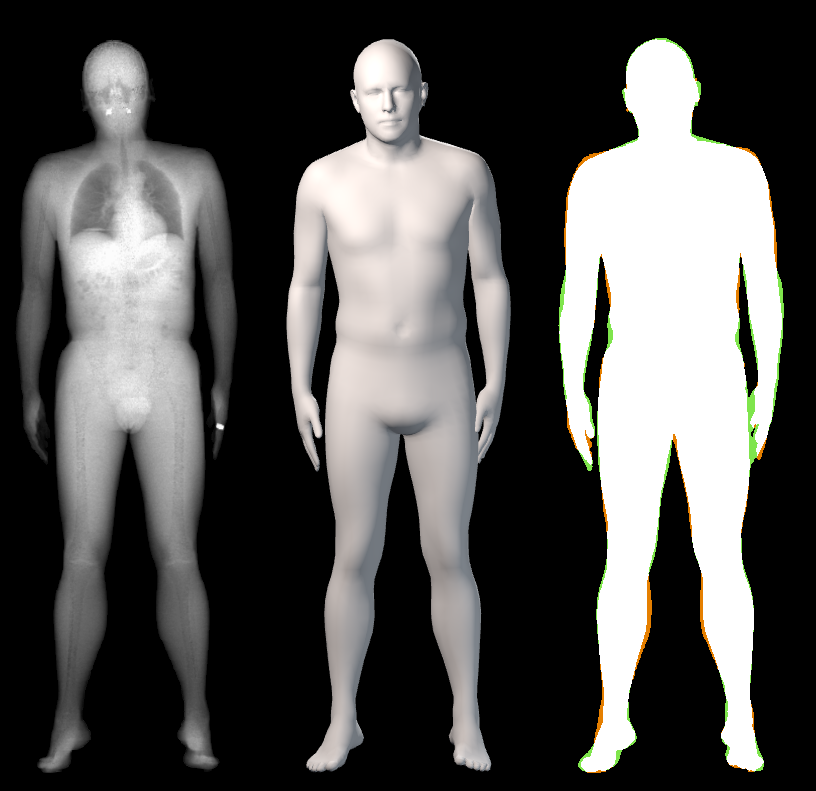}
    \includegraphics[width=0.495\columnwidth]{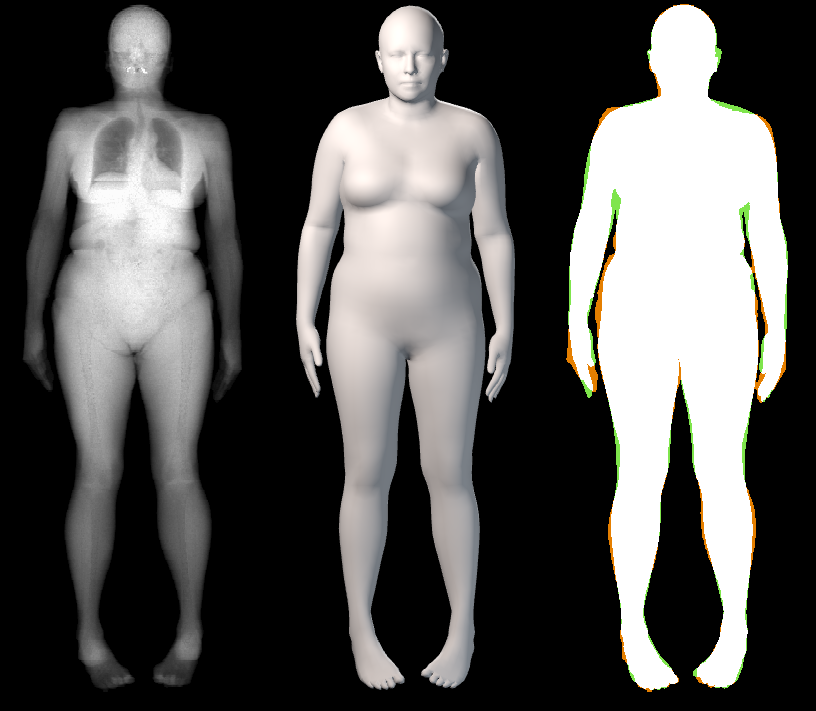}
    \includegraphics[width=0.495\columnwidth]{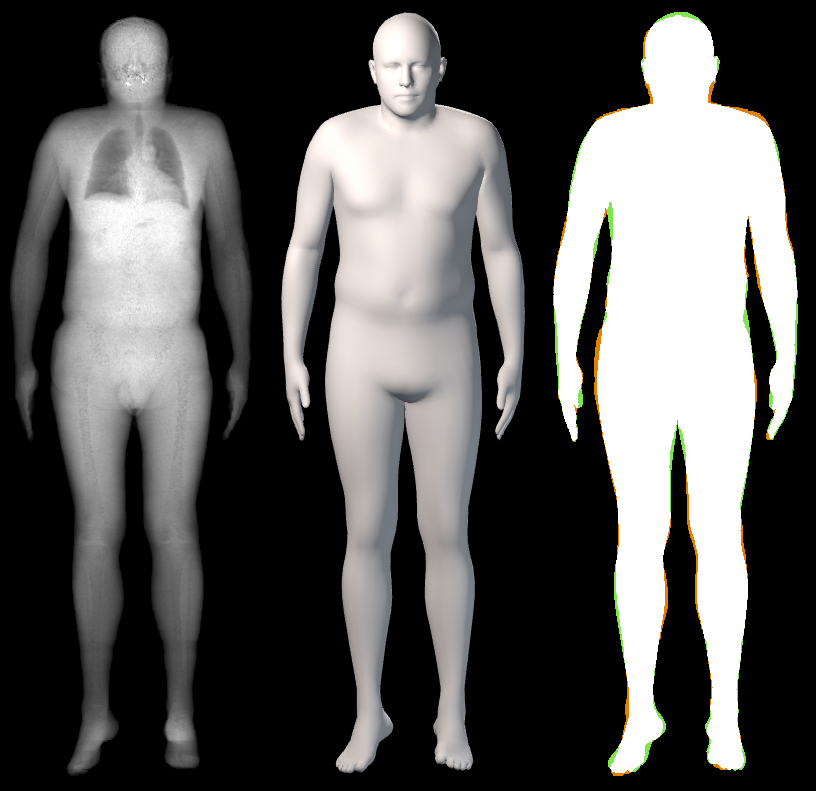}
    \includegraphics[width=0.495\columnwidth]{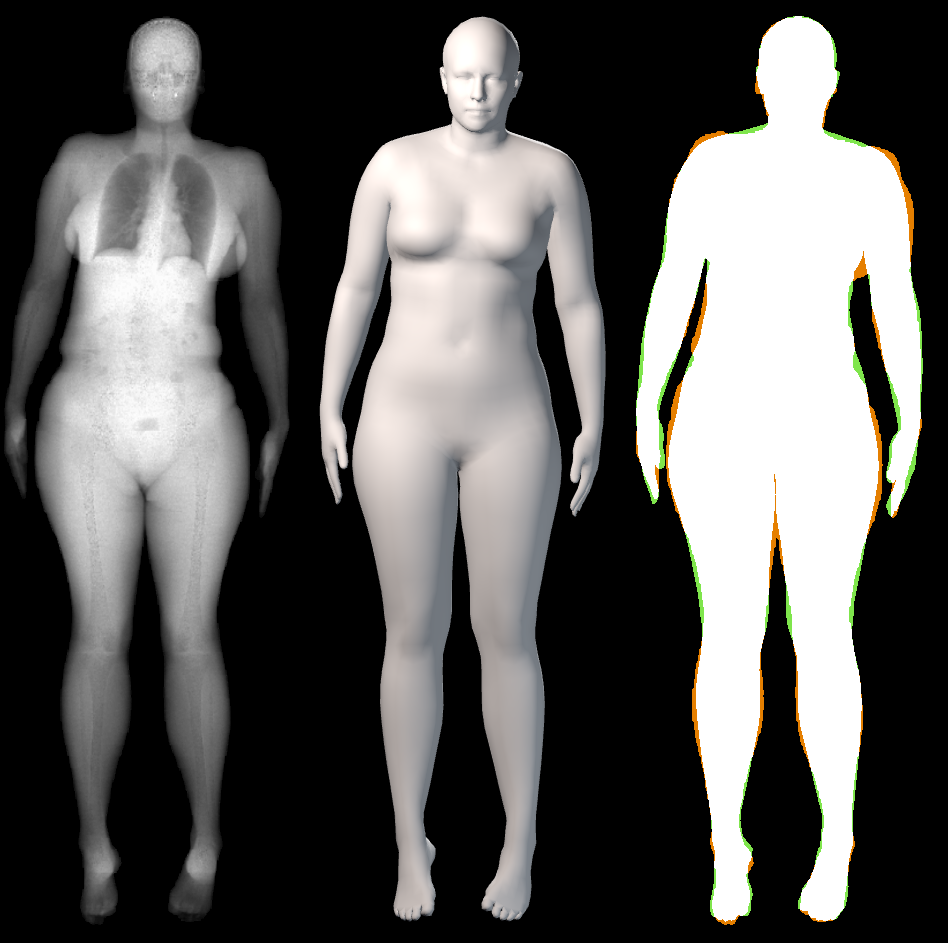}
    \includegraphics[width=0.495\columnwidth]{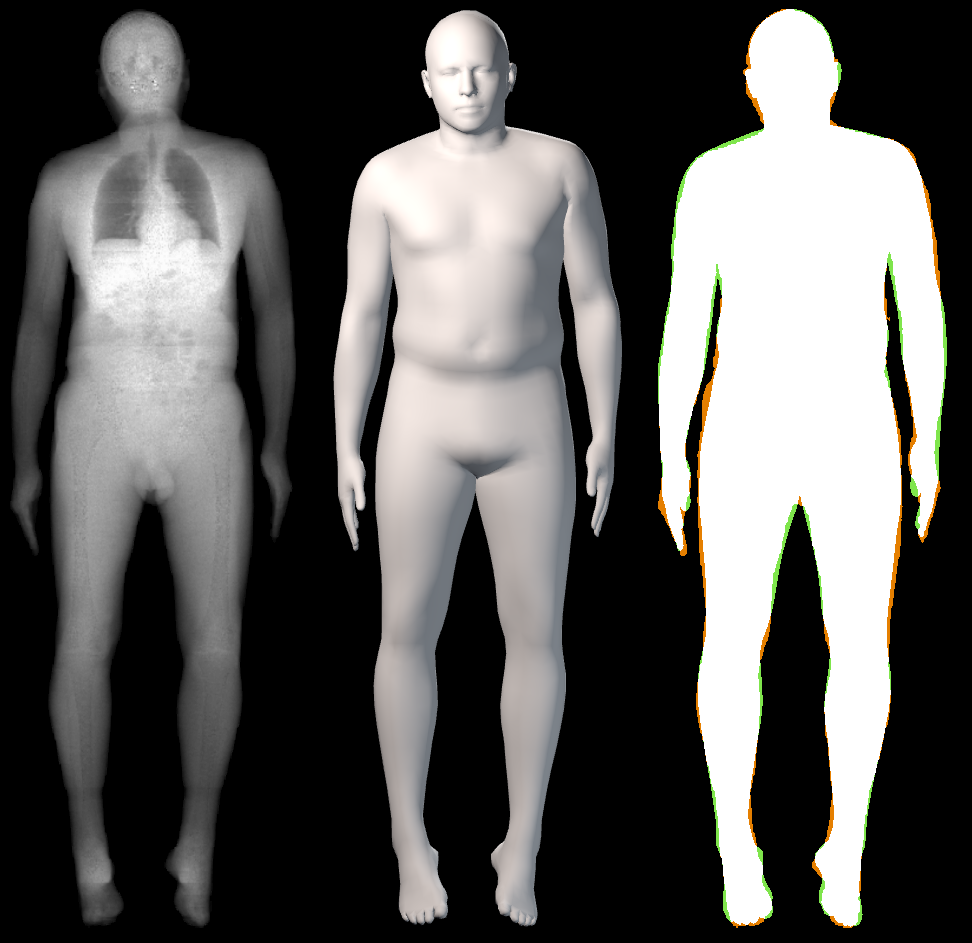}
    \includegraphics[width=0.495\columnwidth]{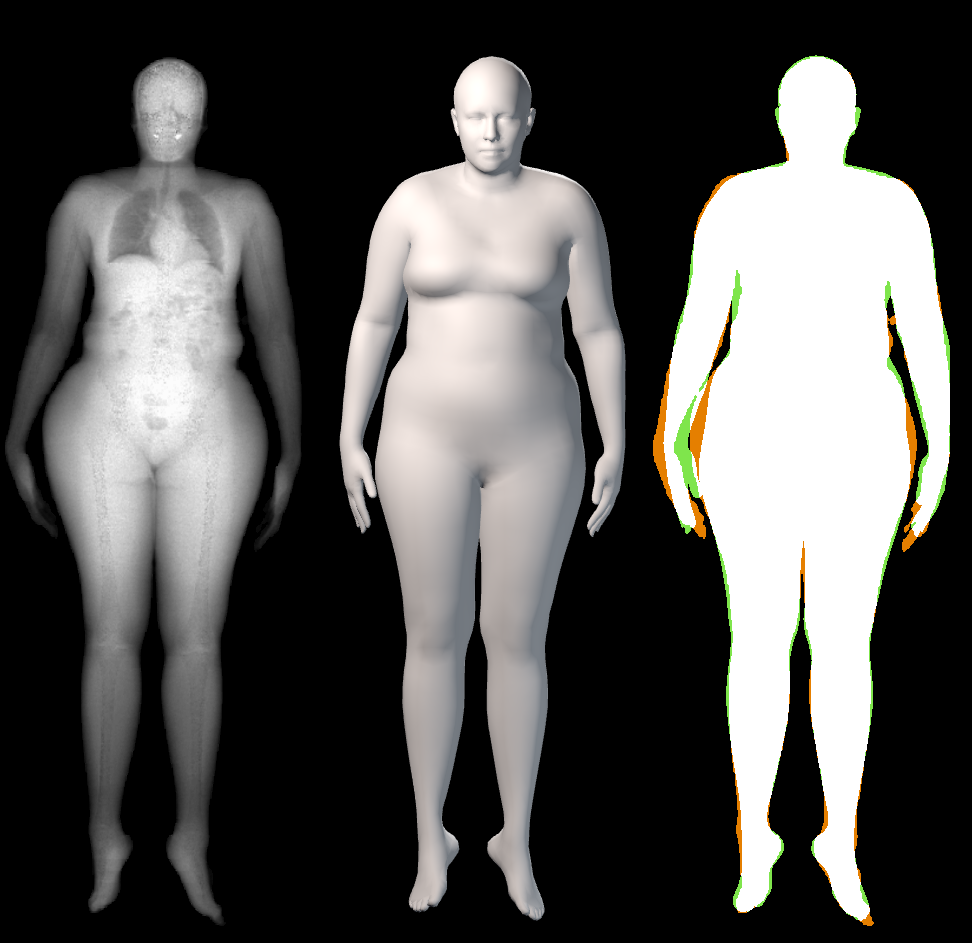}
    \includegraphics[width=0.495\columnwidth]{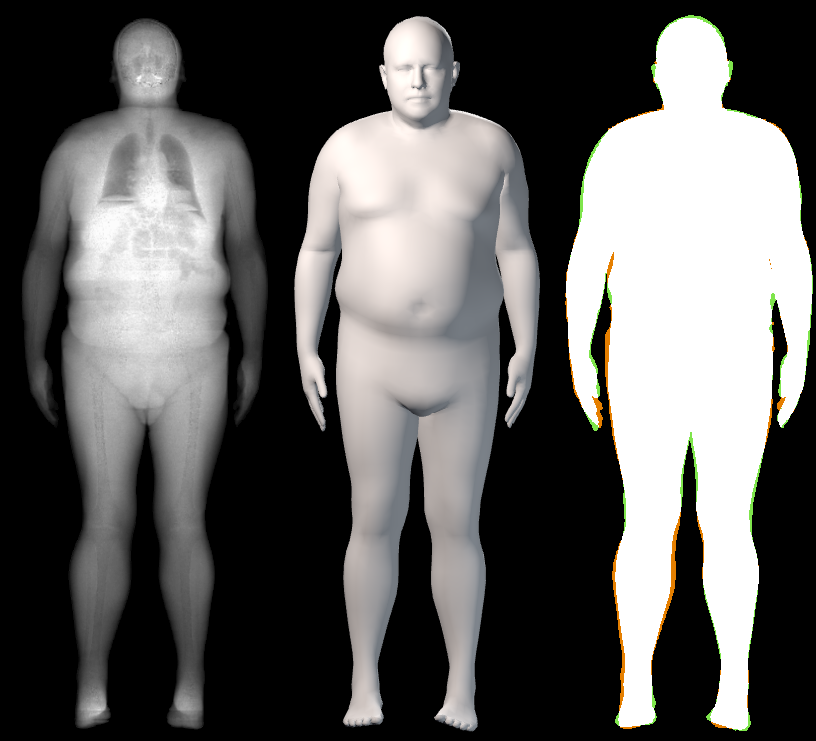}
    \includegraphics[width=0.495\columnwidth]{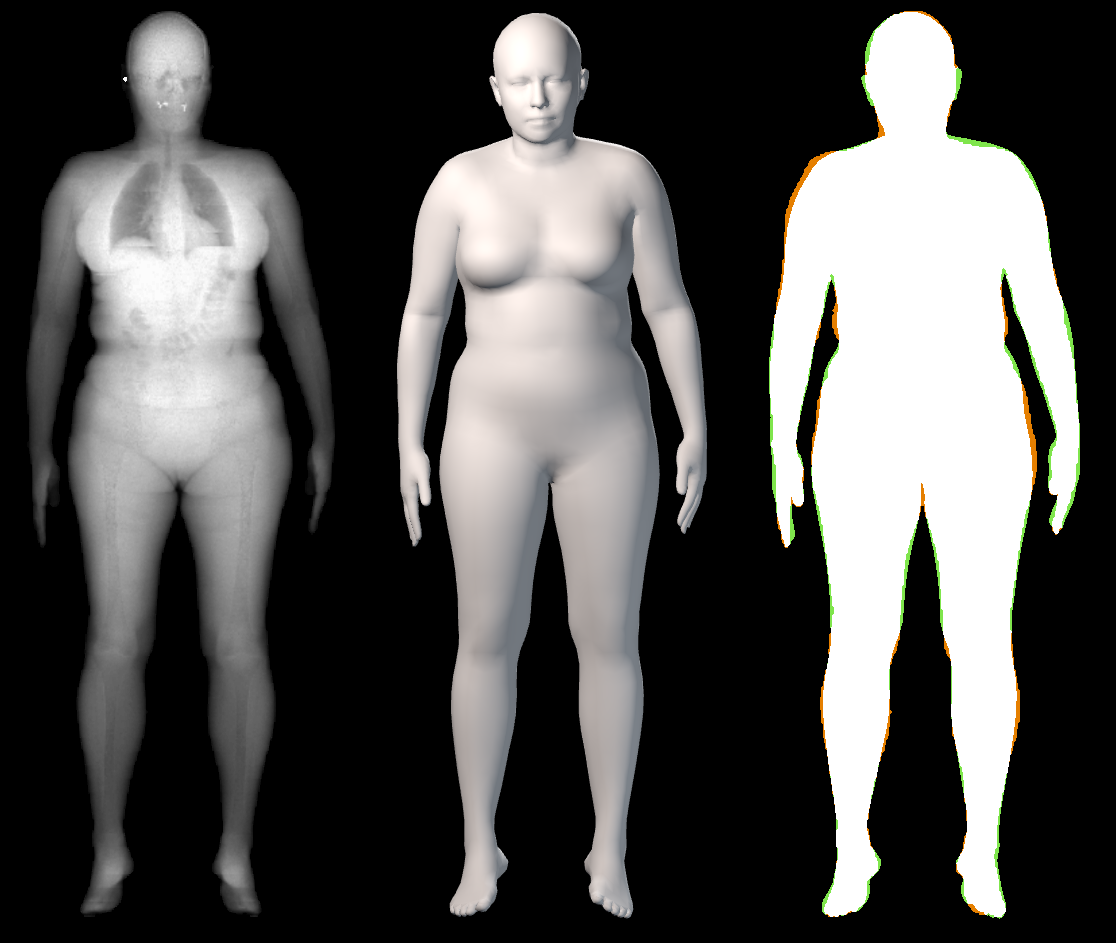}
         \end{mdframed}
    \vspace{-0.1in}
    \includegraphics[width=0.495\columnwidth]{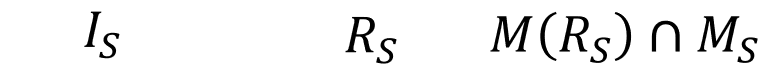}
    \includegraphics[width=0.495\columnwidth]{figures/supmat/skin_alignment/label_skin_alignment.PNG}
    \caption{Comparison of the aligned STAR models $\reg_S$ with the target DXA masks $M_S$ for subjects sampled from the curated dataset. On the left we show males and on the right females. The masks intersection is color-coded as follow: green: $\reg_S$ only, orange: $M_S$ only, white: both.}
    \label{fig:skin_alignment}
\end{figure}

In this section we illustrate the alignment results of the STAR model on the DXA images. Those alignments were obtained with the optimization presented in Sec. 3.3 of the main paper. 
These results complement the quantitative evaluation reported in Sec. 5.1 of the main manuscript, where the intersection over union coefficient between the DXA mask $M_S$ and the computed skin silhouette is $94 \%$ for females and $95 \%$ for males.
In Figure \ref{fig:skin_alignment}, we show the qualitative results. The color-coded images show that the skin registrations faithfully capture the DXA skin silhouettes.

\begin{figure}
    \centering
    \includegraphics[width=.85\columnwidth]{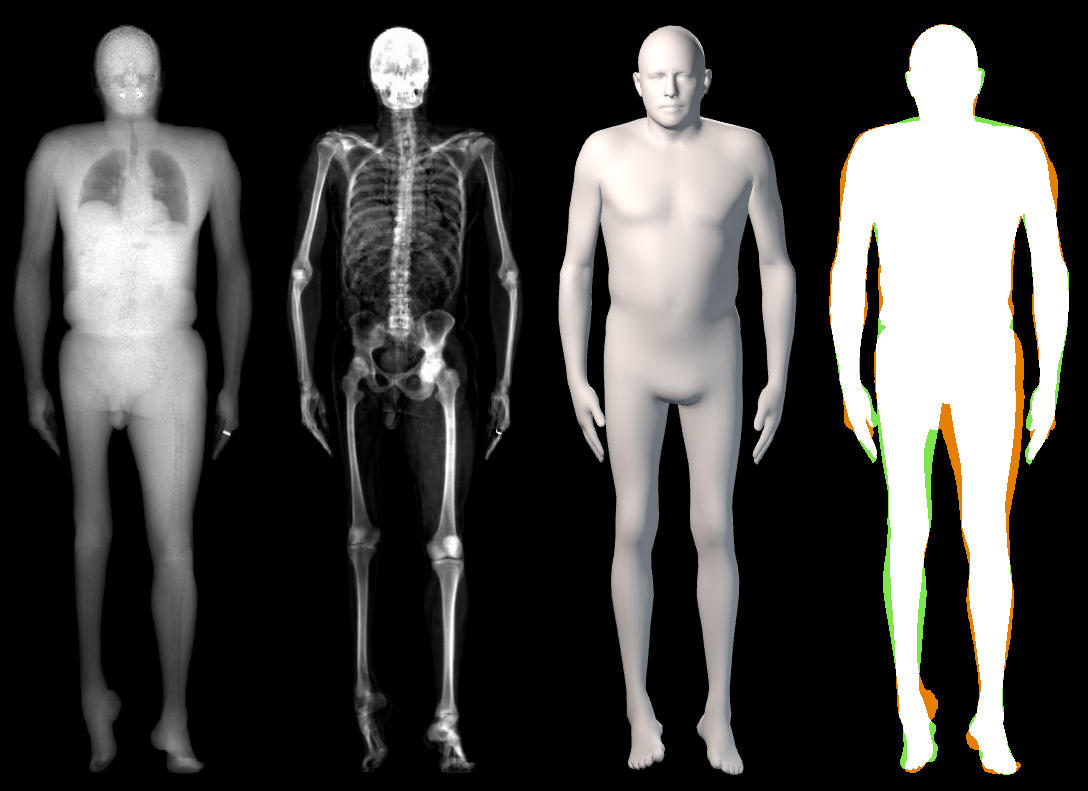}
    \includegraphics[width=.85\columnwidth]{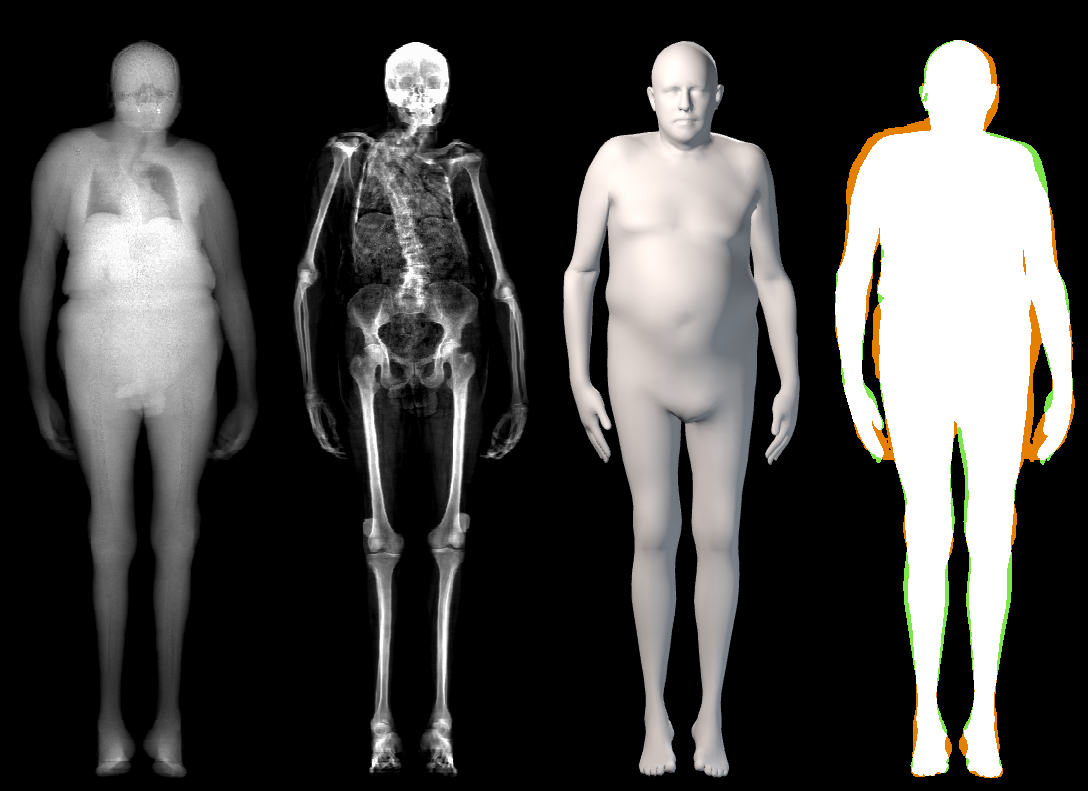}
    \includegraphics[width=.85\columnwidth]{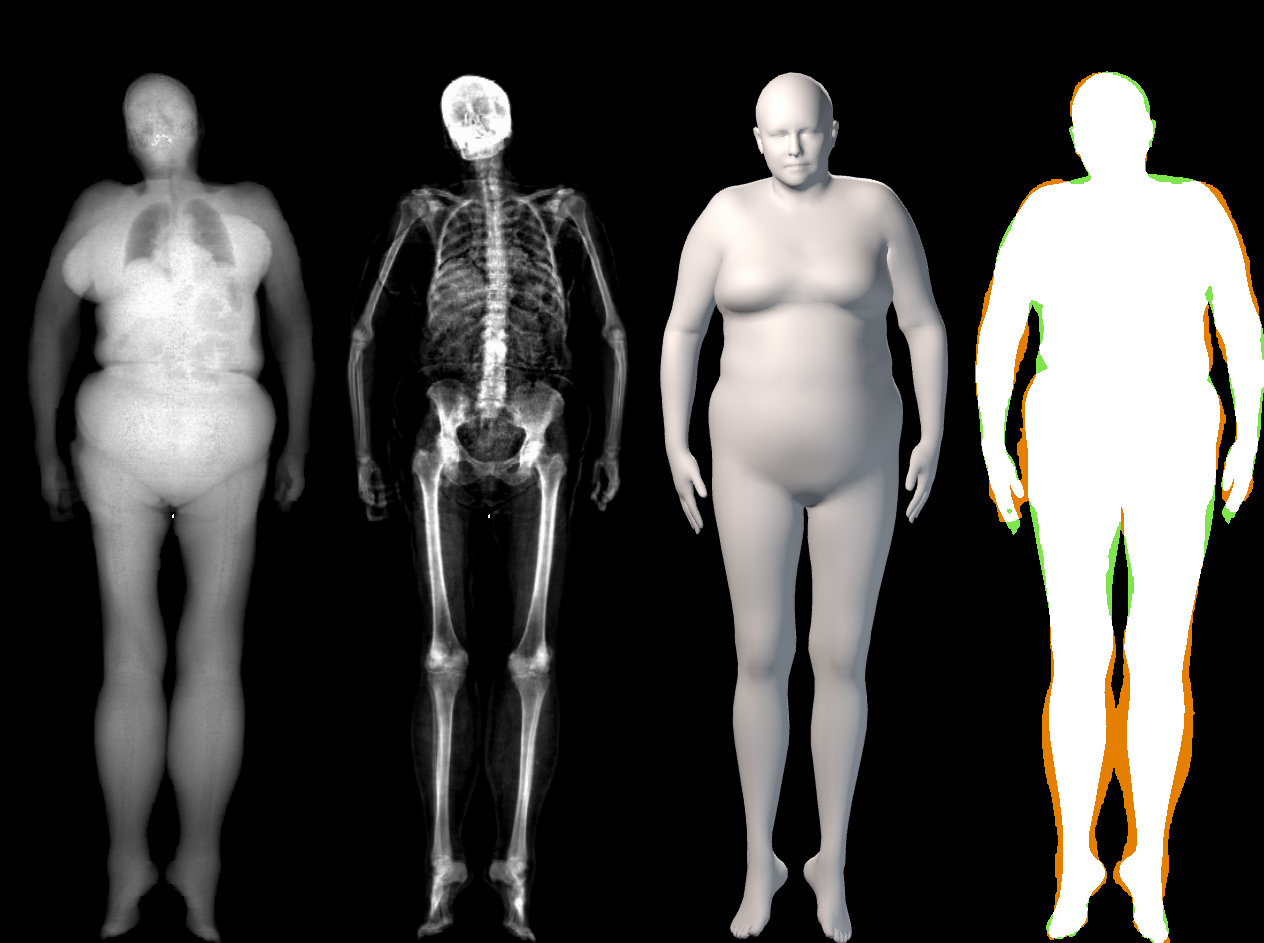}
    \includegraphics[width=.85\columnwidth]{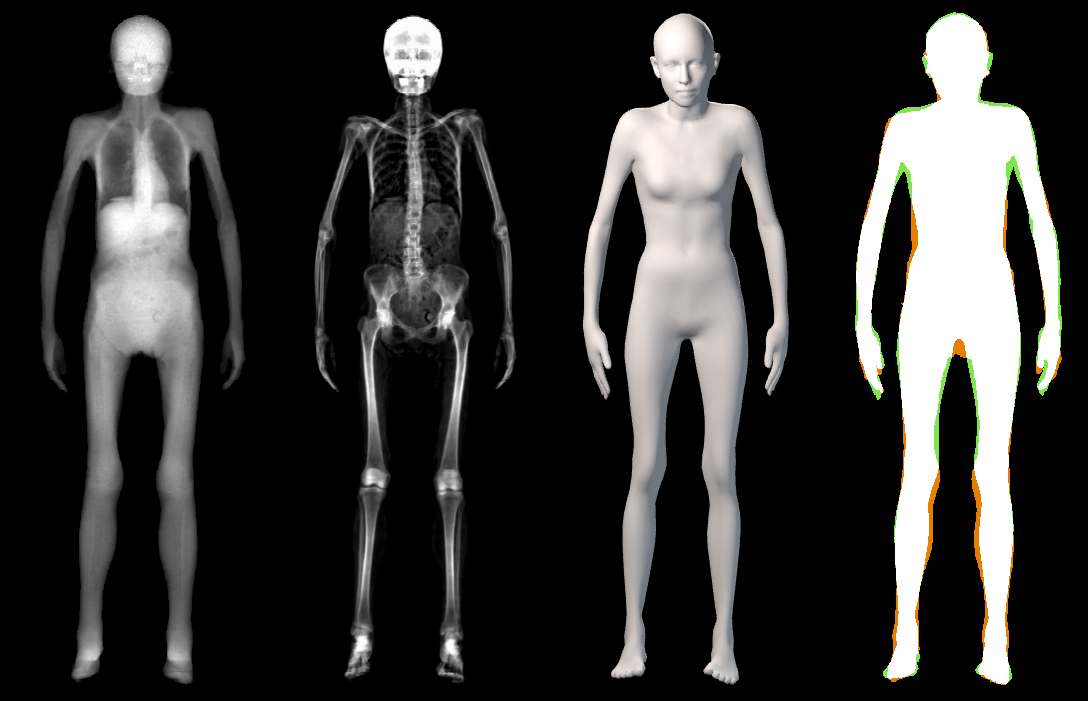}
    \vspace{-0.1in}
     \includegraphics[width=.85\columnwidth]{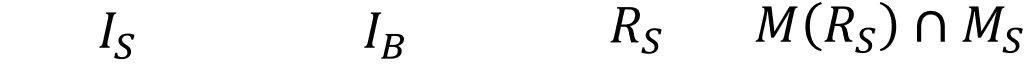}
    \caption{Failure cases. For each subject, we show $I_{S}$, $I_{B}$, the fitted skin mesh $\reg_S$ and the intersection of both masks. The masks intersection is color-coded as follow: green: $\reg_S$ only, orange: $M_S$ only, white: both. The STAR model can not faithfully capture the shape of these subjects.}
    \label{fig:skin_alignment_fail}
\end{figure}

As mentioned in the last paragraph of Sec. 3.3, we use the
quality of the fit
to detect and remove failure cases from our datasets, i.e. subjects whose body shape can not be explained with STAR. 
In \figref{fig:skin_alignment_fail}, we show some failure cases with low intersection over union values. 
These examples include subjects with atrophied or swollen limbs, severe scoliosis or very low BMI. 
In practice, we used the alignment score to remove outliers of the available DXA scans (about 1\%) to constitute a curated dataset containing a training set of 1000 subjects and a test set of 200 subjects for each gender.

\begin{figure}[!]
    \centering
    \begin{mdframed}[backgroundcolor=black, align=center]
        \includegraphics[width=0.495\columnwidth]{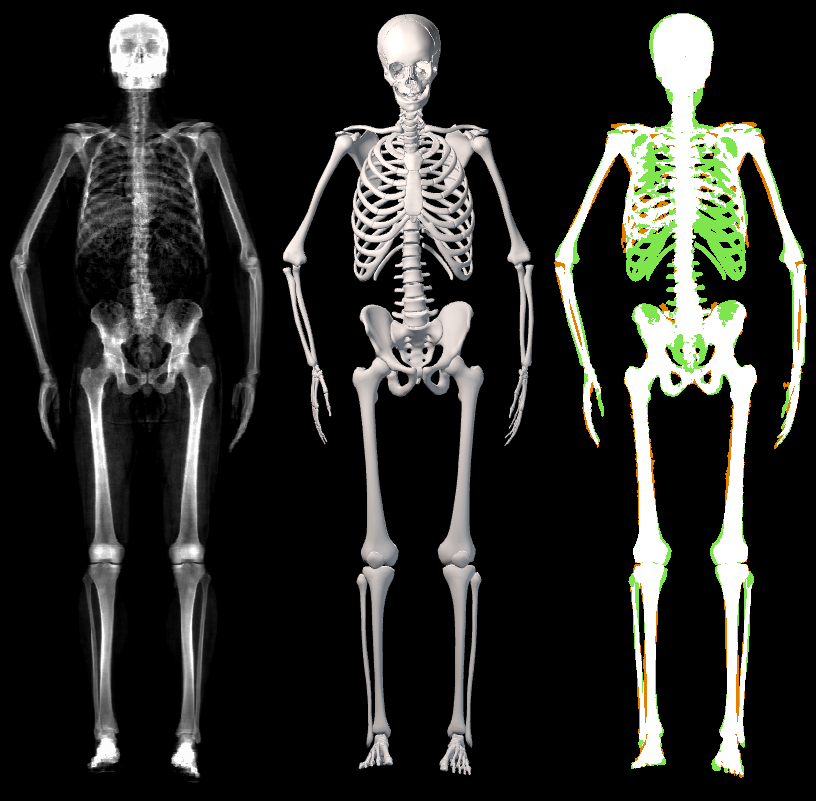}
        \includegraphics[width=0.495\columnwidth]{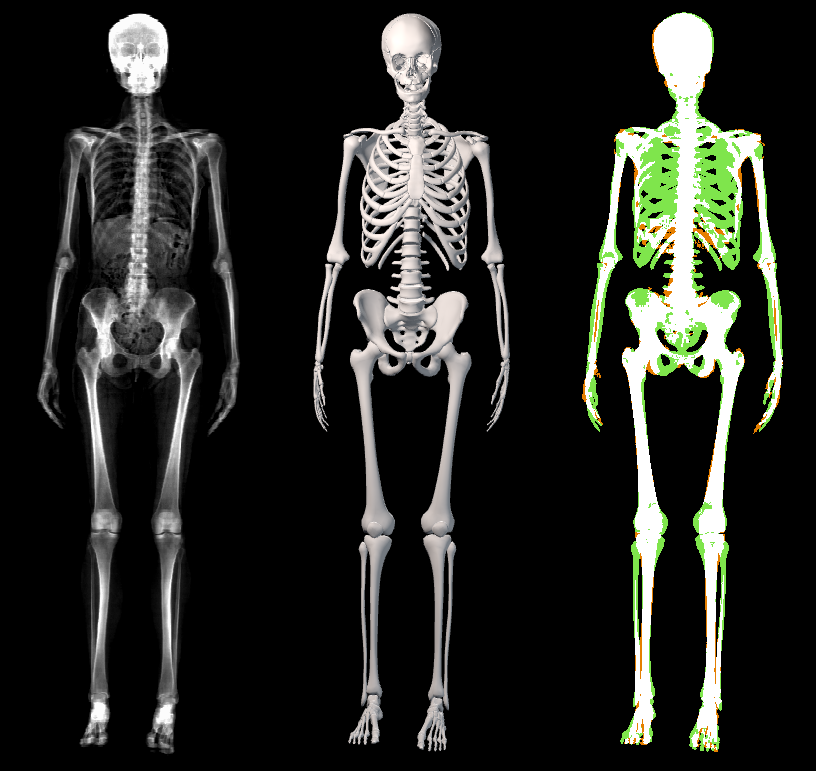}
        \includegraphics[width=0.495\columnwidth]{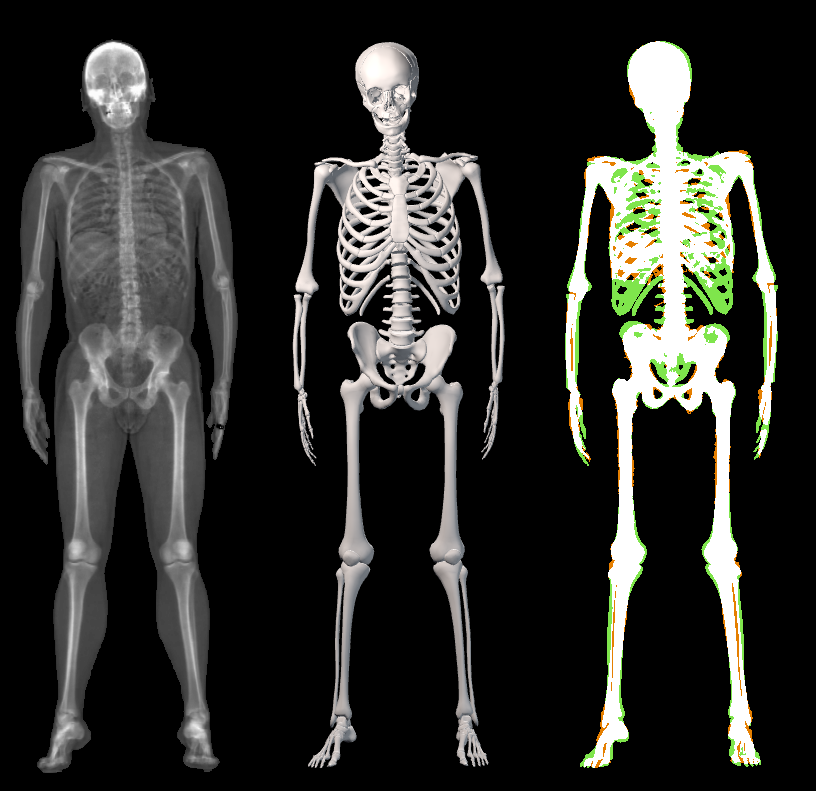}
        \includegraphics[width=0.495\columnwidth]{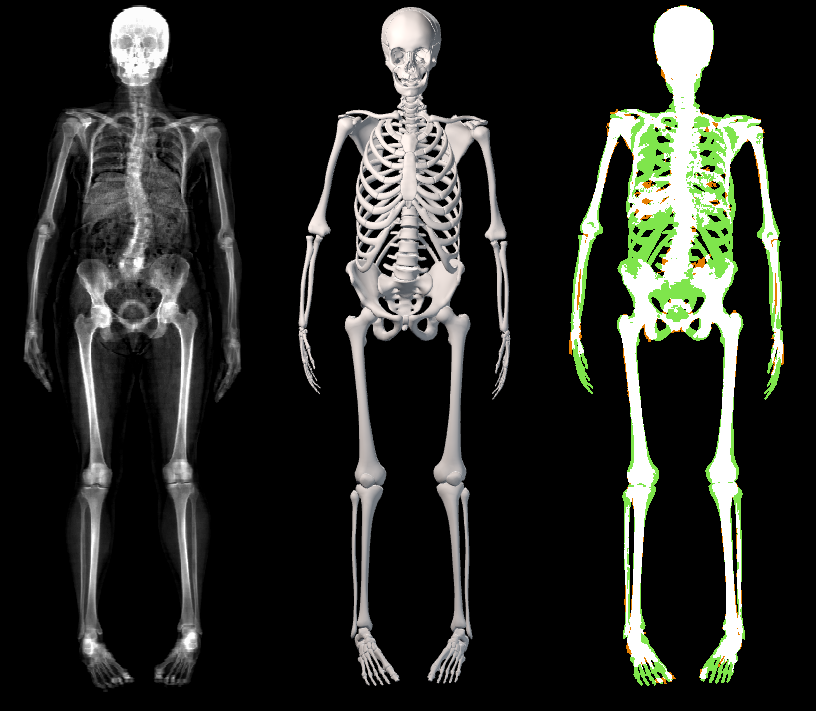}
        \includegraphics[width=0.495\columnwidth]{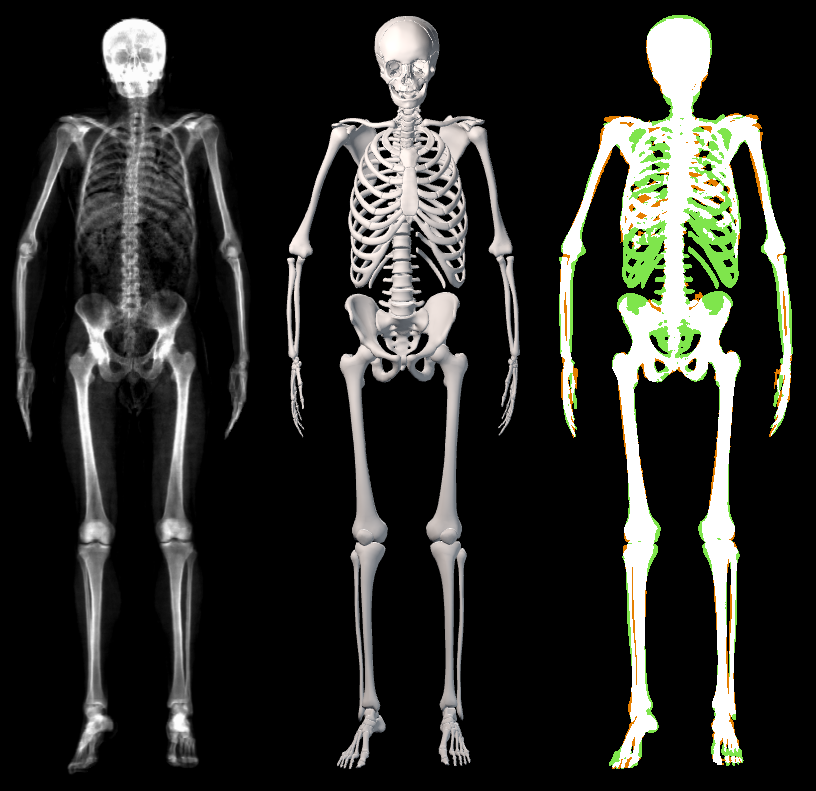}
        \includegraphics[width=0.495\columnwidth]{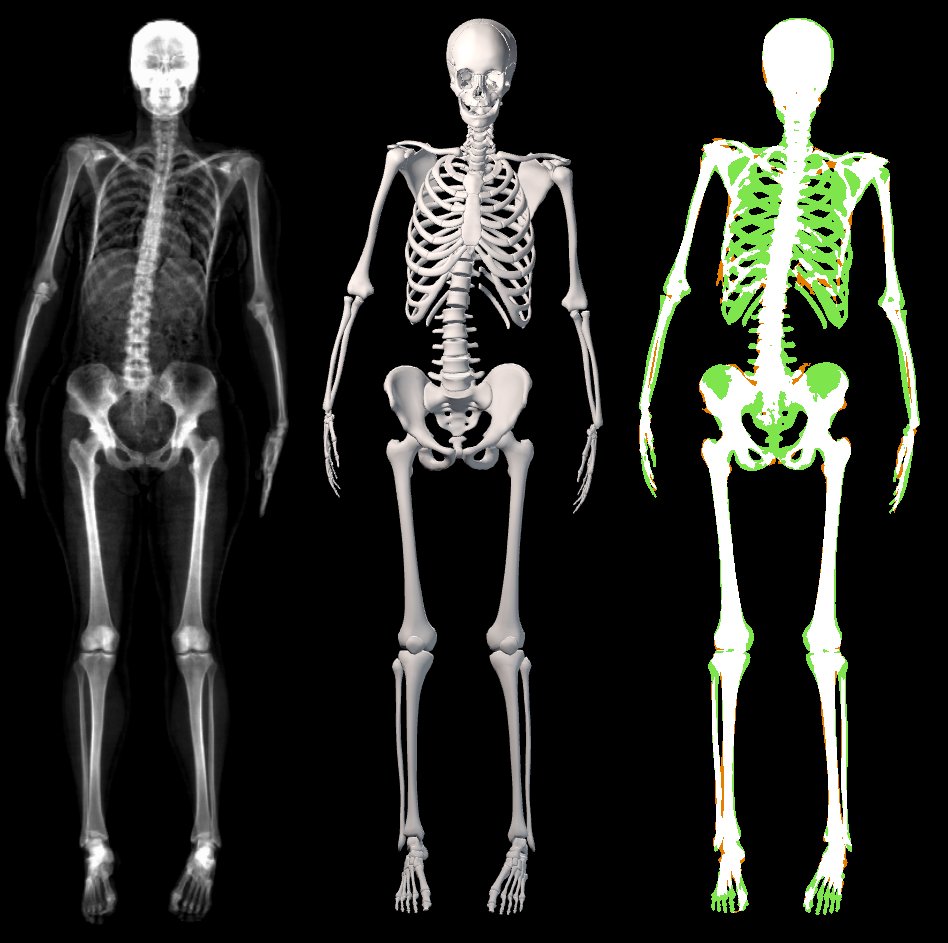}
        \includegraphics[width=0.495\columnwidth]{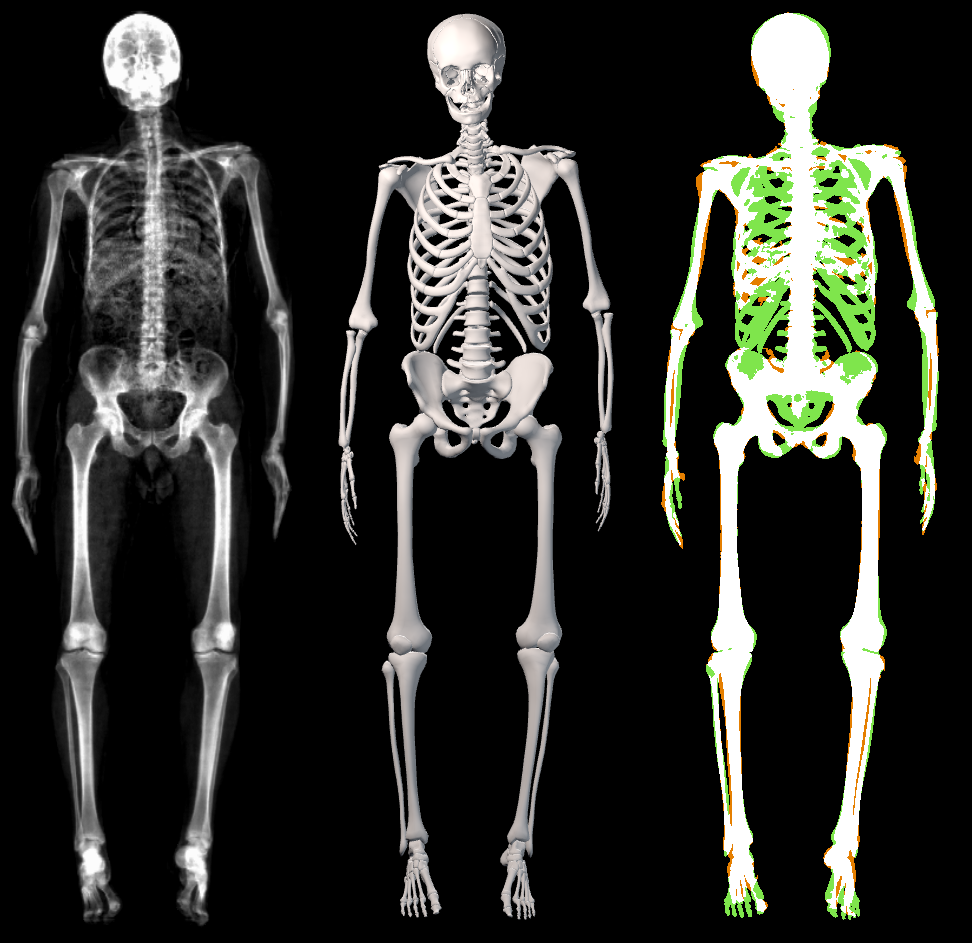}
        \includegraphics[width=0.495\columnwidth]{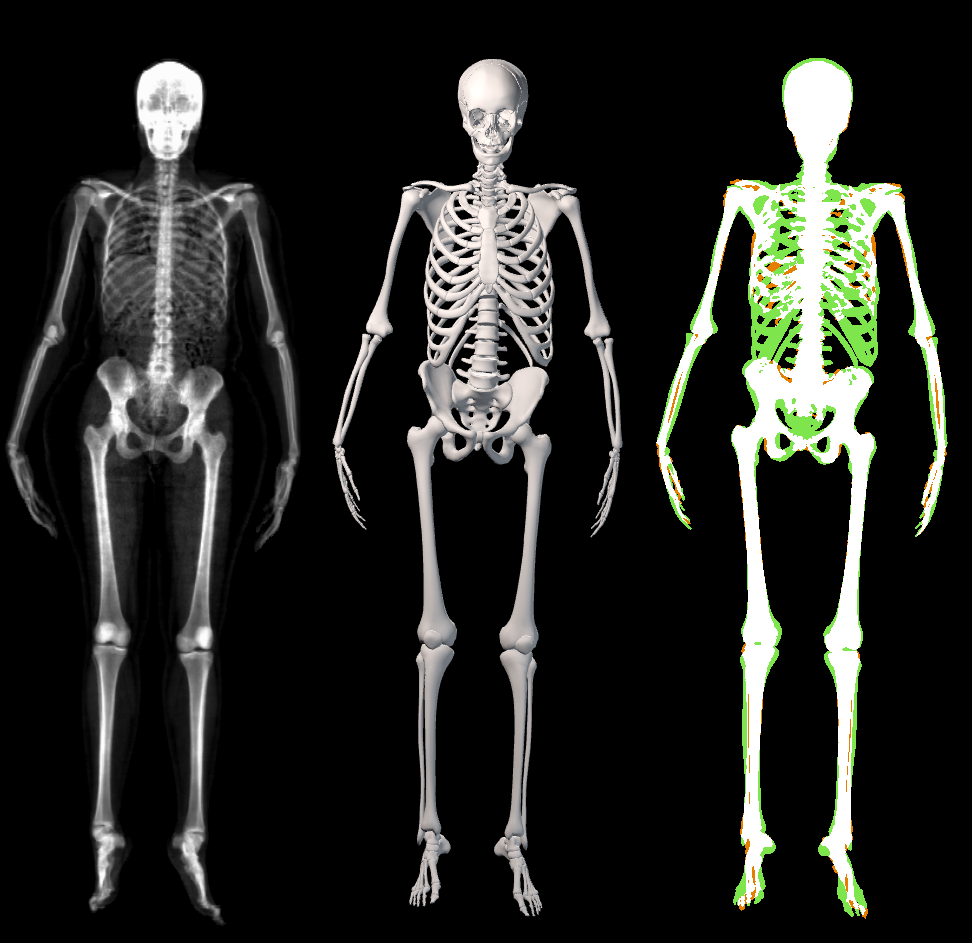}
        \includegraphics[width=0.495\columnwidth]{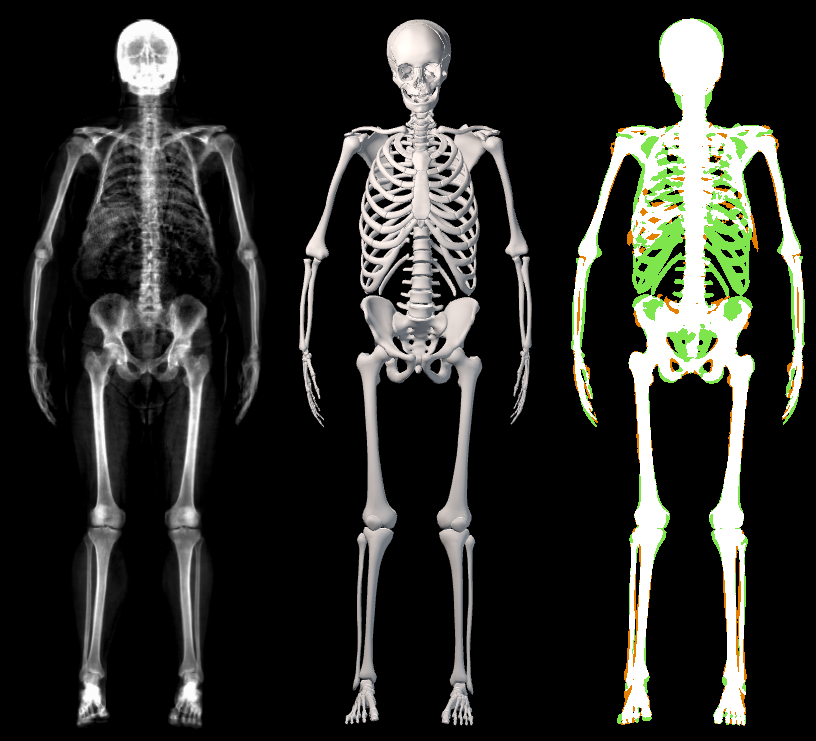}
        \includegraphics[width=0.495\columnwidth]{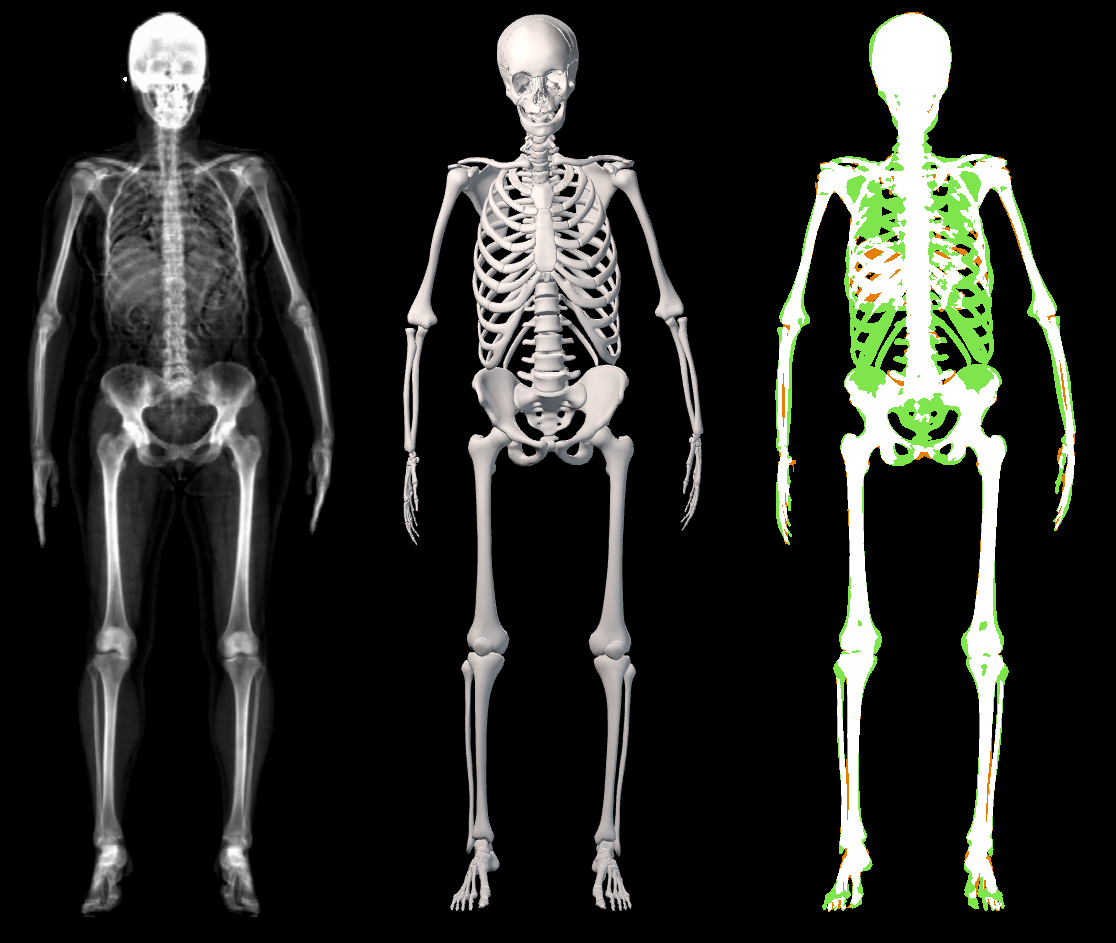}
    \end{mdframed}
    \includegraphics[width=0.495\columnwidth]{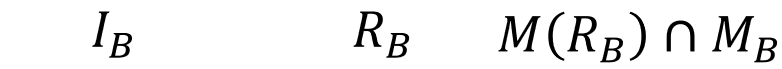}
    \includegraphics[width=0.495\columnwidth]{figures/supmat/skel_alignment/label_skel_alignment.PNG}

    \caption{Comparison of the registered skeleton $\reg_B$ with the target DXA masks $M_B$ for subjects sampled from the training dataset. On the left we show males and on the right females. The masks difference is color-coded as follow: green: $\reg_B$ only, orange: $M_B$ only, white: both.}
    \label{fig:skel_alignment}
\end{figure}

\subsection{Skeleton 3D landmarks regression evaluation}
\begin{figure}[H]
    \centering
    \includegraphics[width=1\columnwidth]{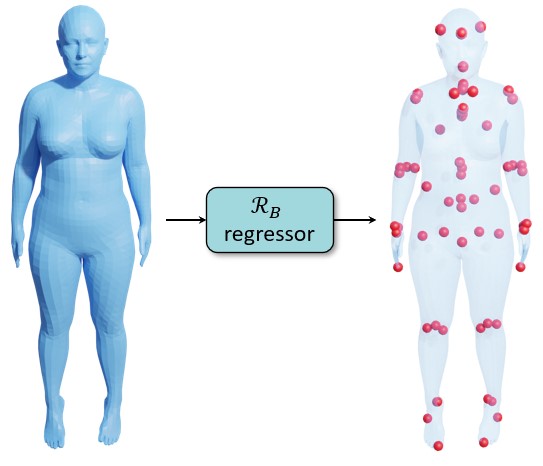}
    \vspace{-0.1in}
    \caption{Given a skin mesh, the landmark regressor lets us compute the landmark 3D locations as a linear combination of the skin mesh vertices locations.}
    \label{fig:3d_landmarks_regression_recap}
\end{figure}

In Sec. 4.1 of the main paper, we explain how we train a regressor that, taking as input the vertices of the skin, predicts the 3D location of the landmarks $\landmarks_B$ (presented in \figref{fig:stitched_puppet} right).
This regression is learned in a normalized lying down pose as illustrated in~\figref{fig:3d_landmarks_regression_recap}.

To evaluate the $\landmarks_B$ regressor accuracy, we learn the regressor from the 1000 train subjects and evaluate on the 200 left out subjects.
We compute the 3D distance between the regressed landmarks position and its ground truth position. 
In Sec. 5.2 of the main paper we provide a general evaluation on the accuracy of the regressor as well as a discussion of the results.
The detailed per landmark errors are listed in Table \ref{tab:3d_regressed_ldm_evaluation}.

\begin{figure*}[p]
    \centering
    \includegraphics[width=1\textwidth]{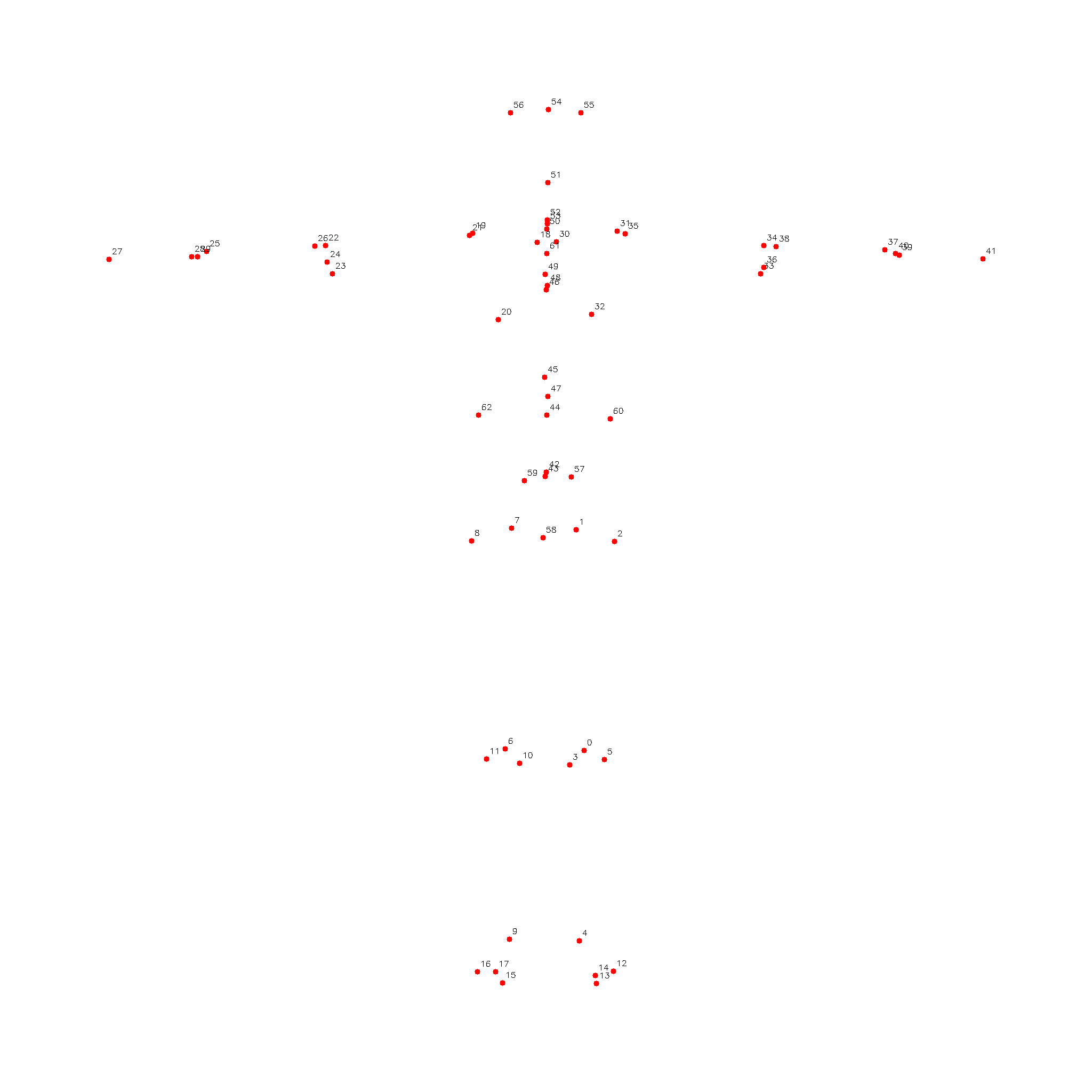}
    \caption{Landmarks $\landmarks_B$ on the skeleton mesh with landmark number.}
    \label{fig:landmark_numbers}
\end{figure*}

\input{sections/supmat_ldm_array}

\subsection{Skeleton registration qualitative evaluation}

Next we show qualitative results of the skeleton registrations $\reg_B$ in \figref{fig:skel_alignment}. The subjects are the same as in \figref{fig:skin_alignment}.
These results complement the Sec. 5.3 of the main document,
and precisely, the numeric value reported in the first row of Table 1 in the main document.

\subsection{OSSO VS Anatomy Transfer comparison}

\begin{figure*}[p]
    \centering
    \begin{mdframed}[backgroundcolor=black]
        \includegraphics[width=0.490\columnwidth]{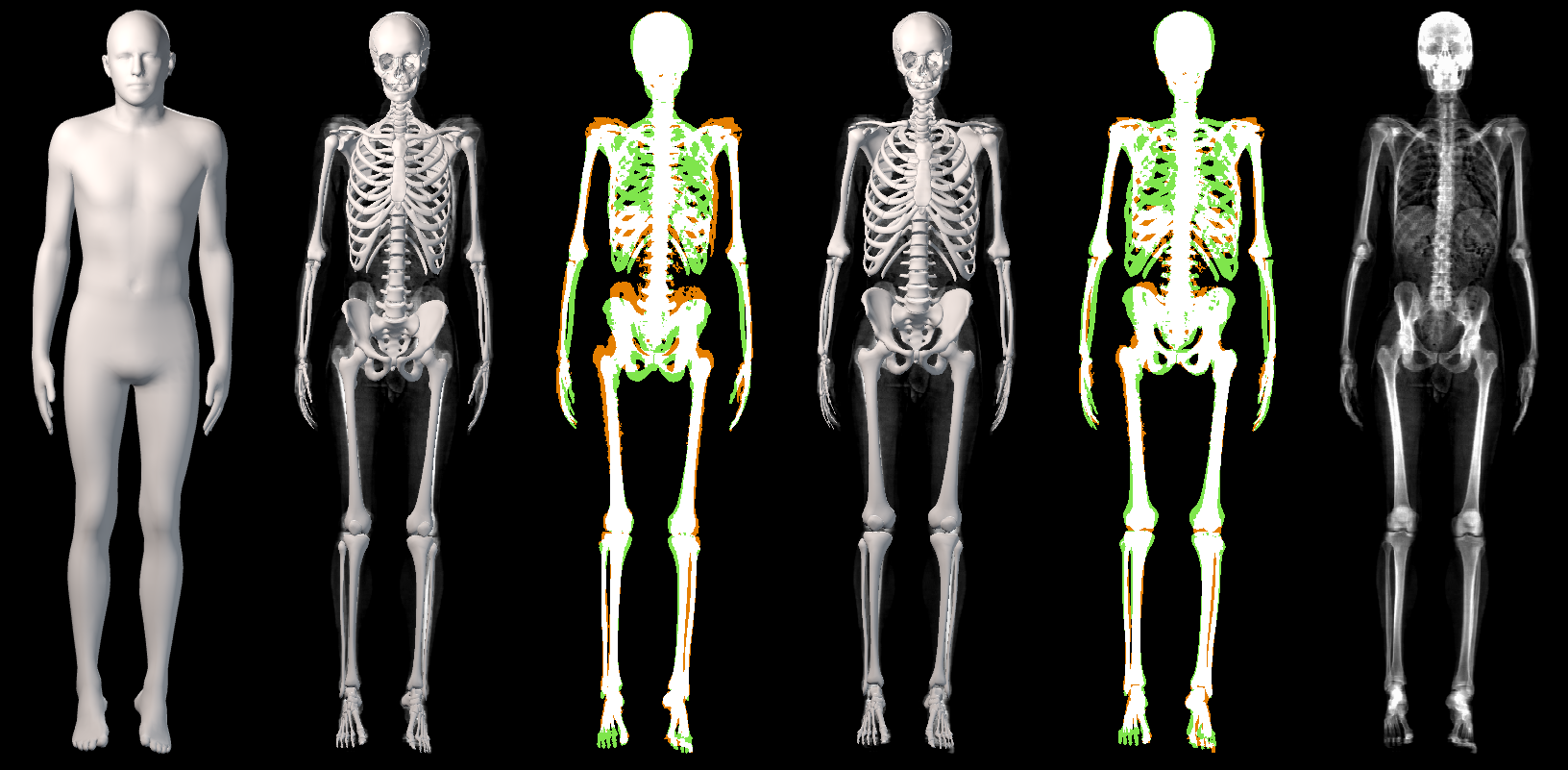}
        \includegraphics[width=0.490\columnwidth]{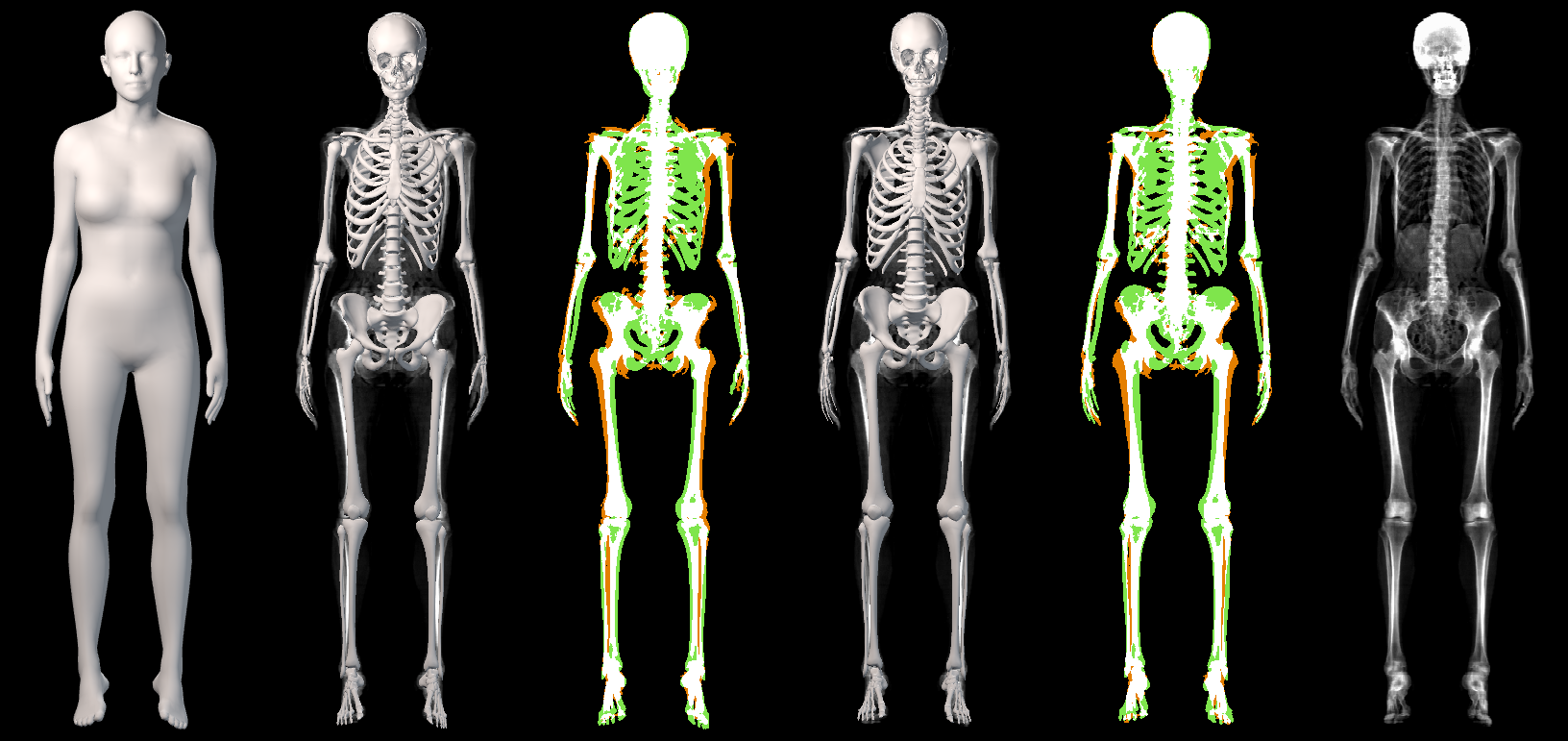}
        
        \includegraphics[width=0.490\columnwidth]{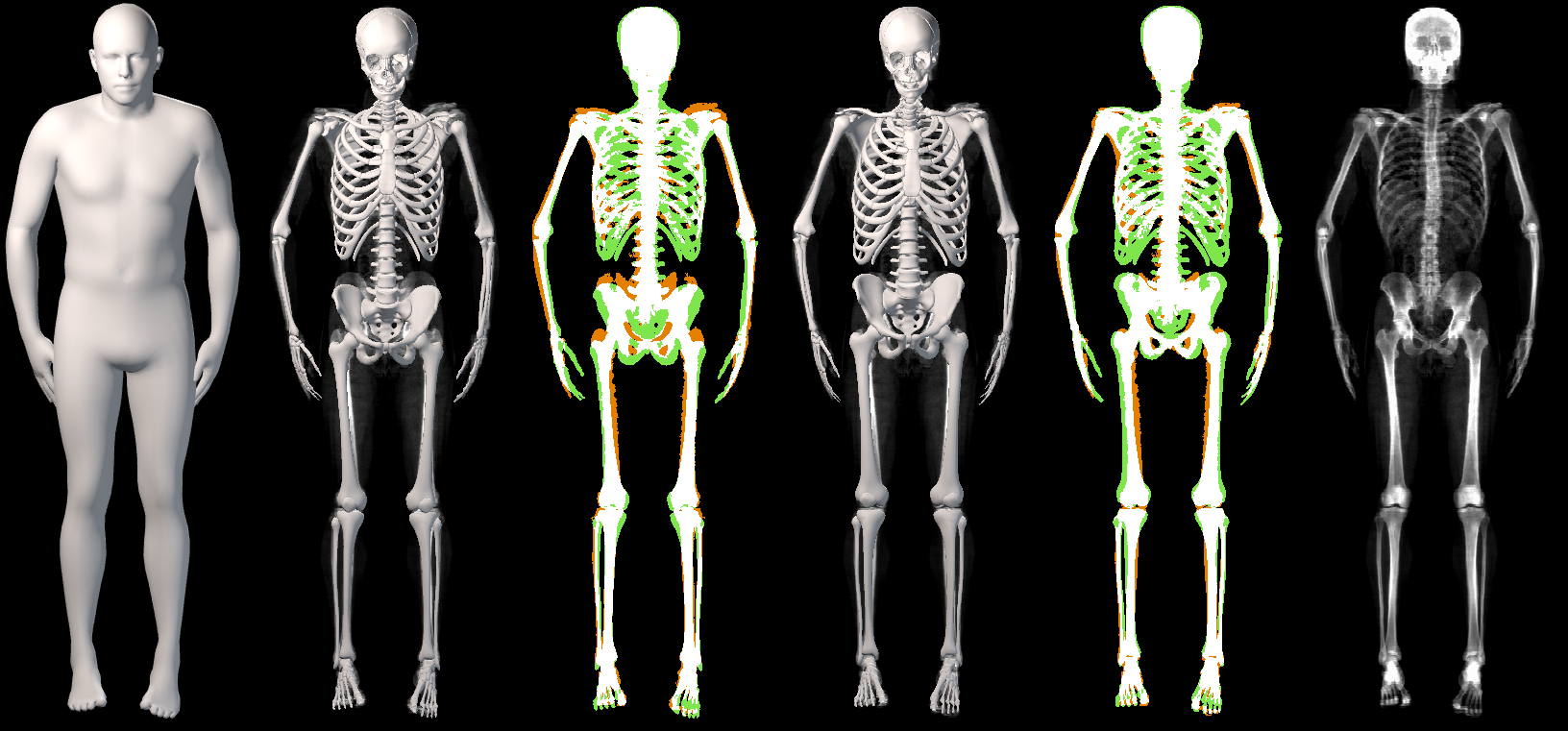}
        \includegraphics[width=0.490\columnwidth]{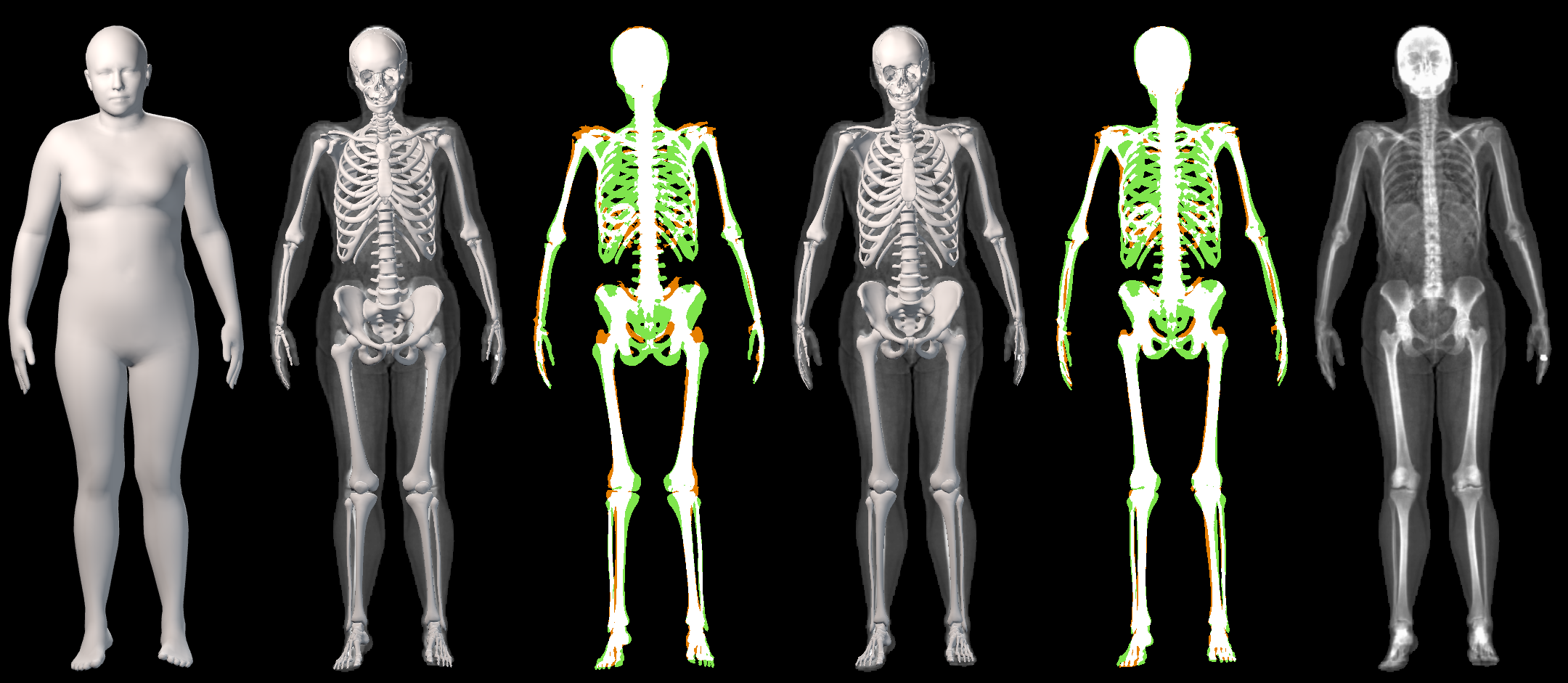}
        
        \includegraphics[width=0.490\columnwidth]{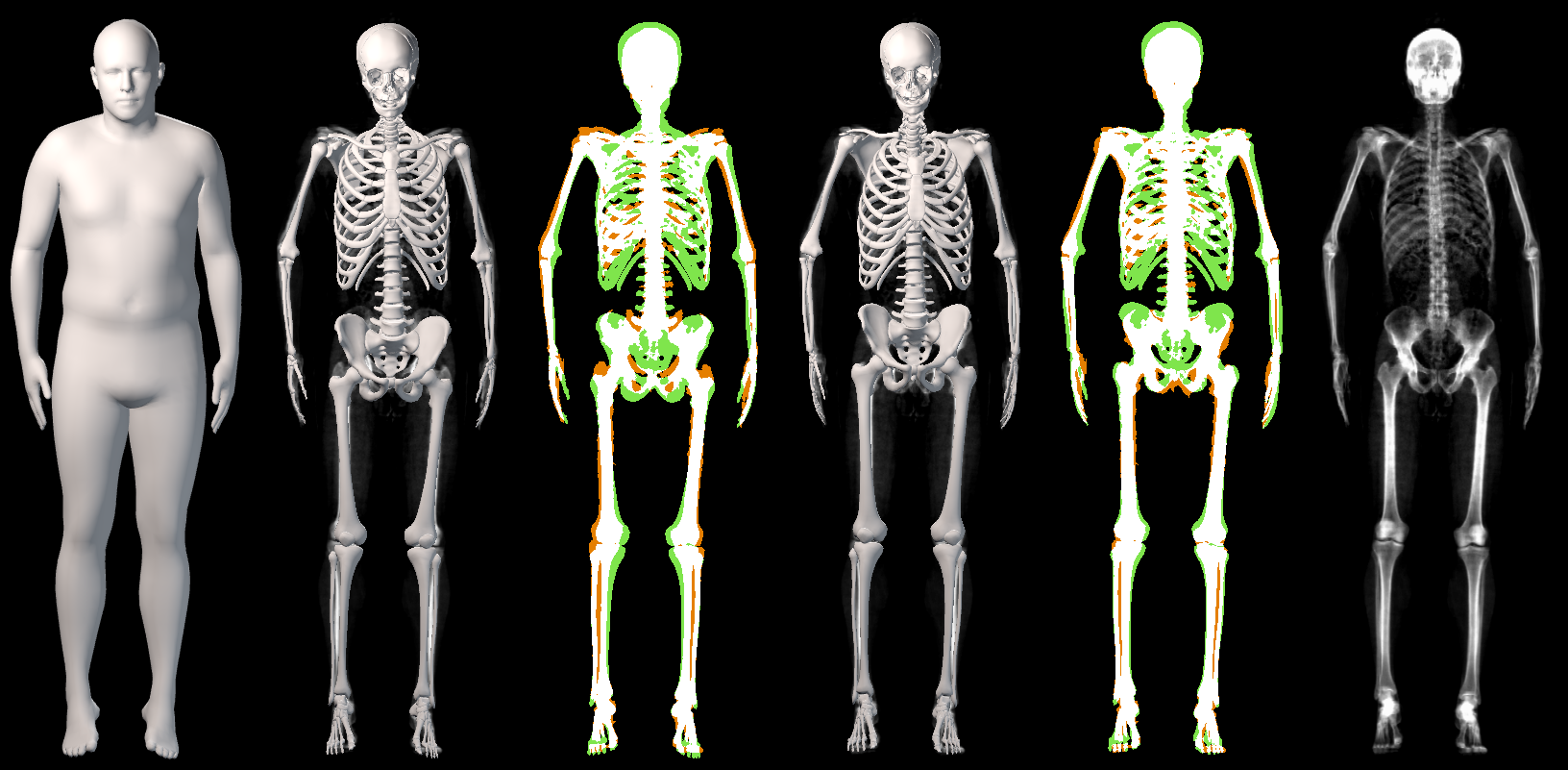}
        \includegraphics[width=0.490\columnwidth]{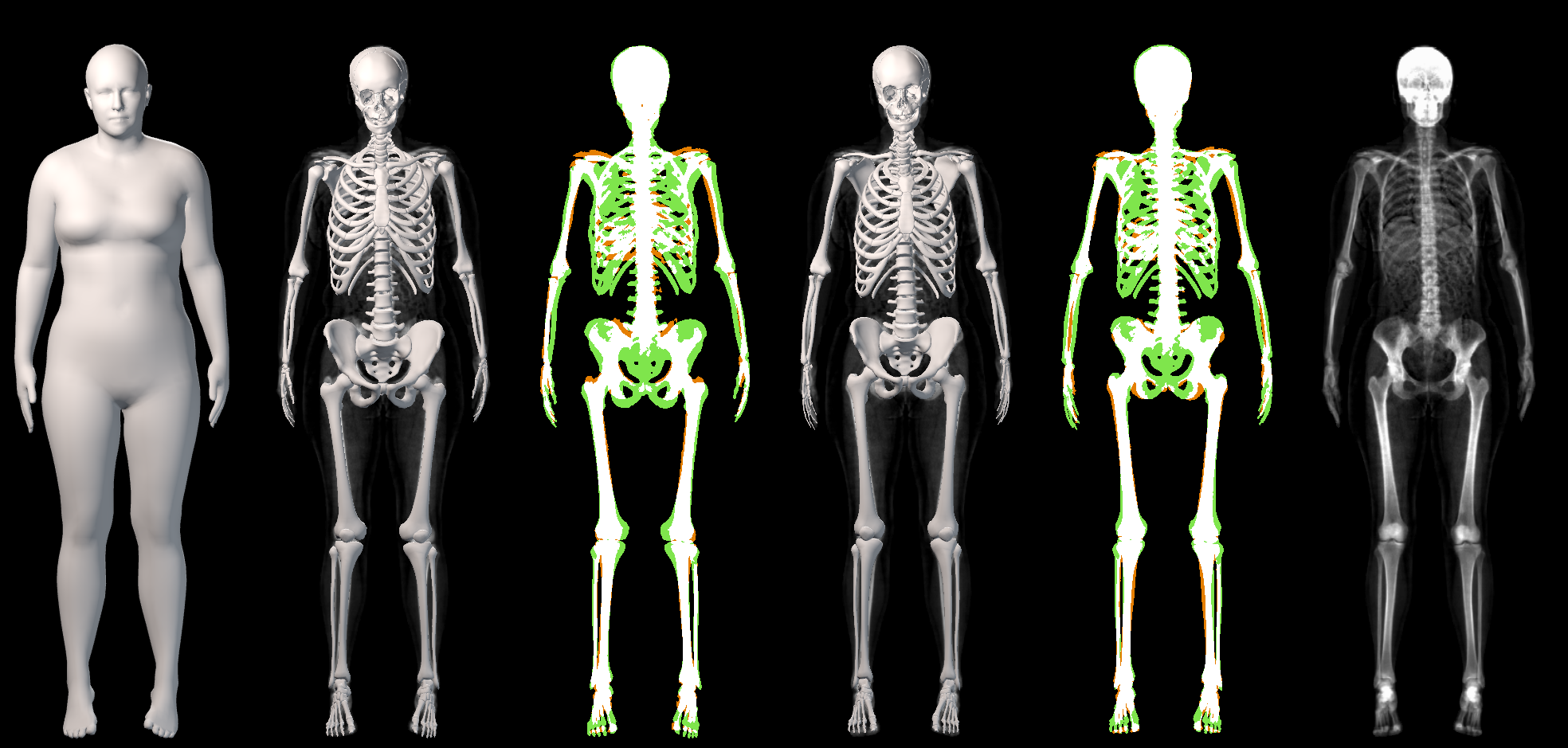}
        
        \includegraphics[width=0.490\columnwidth]{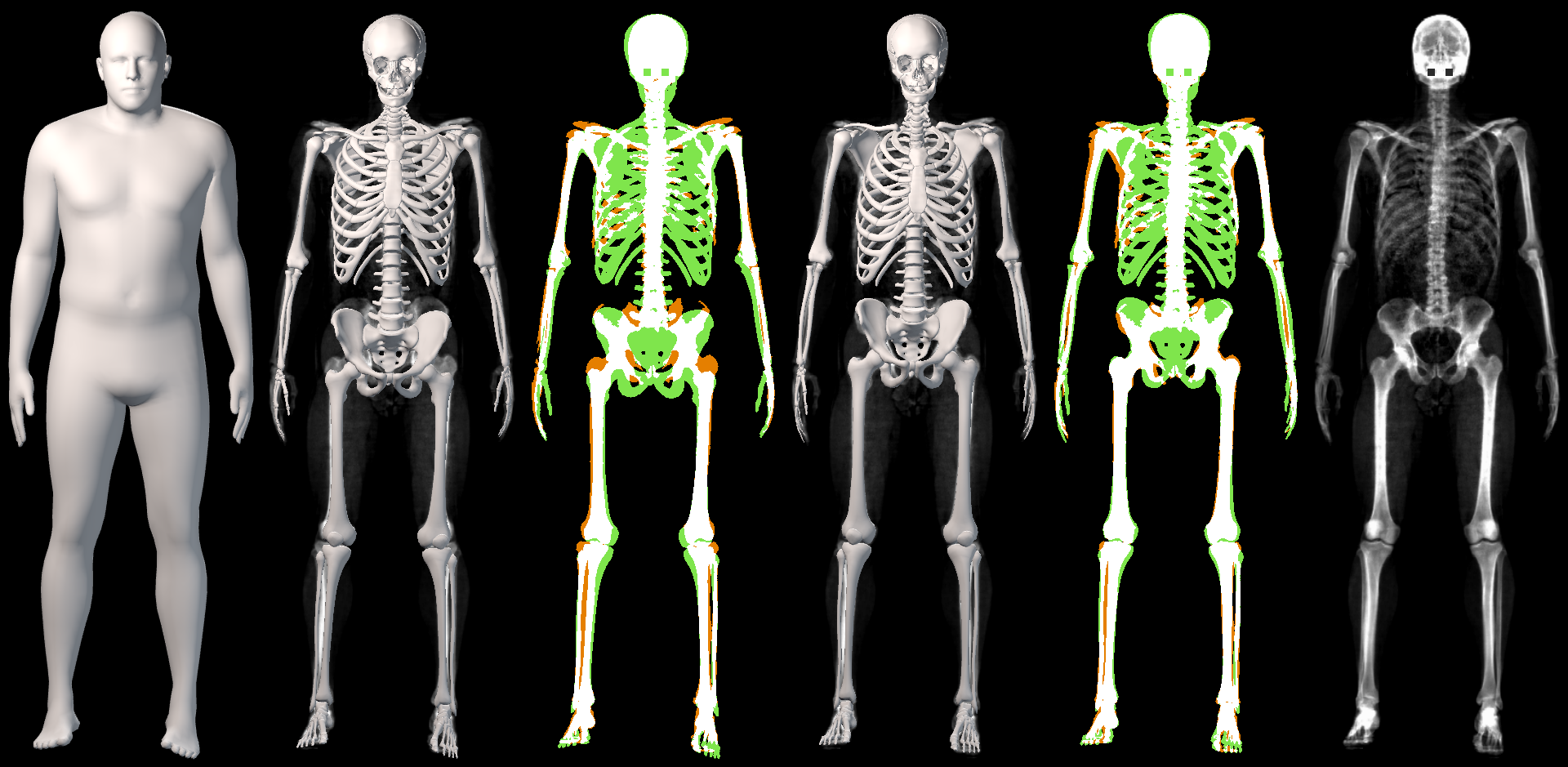}
        \includegraphics[width=0.490\columnwidth]{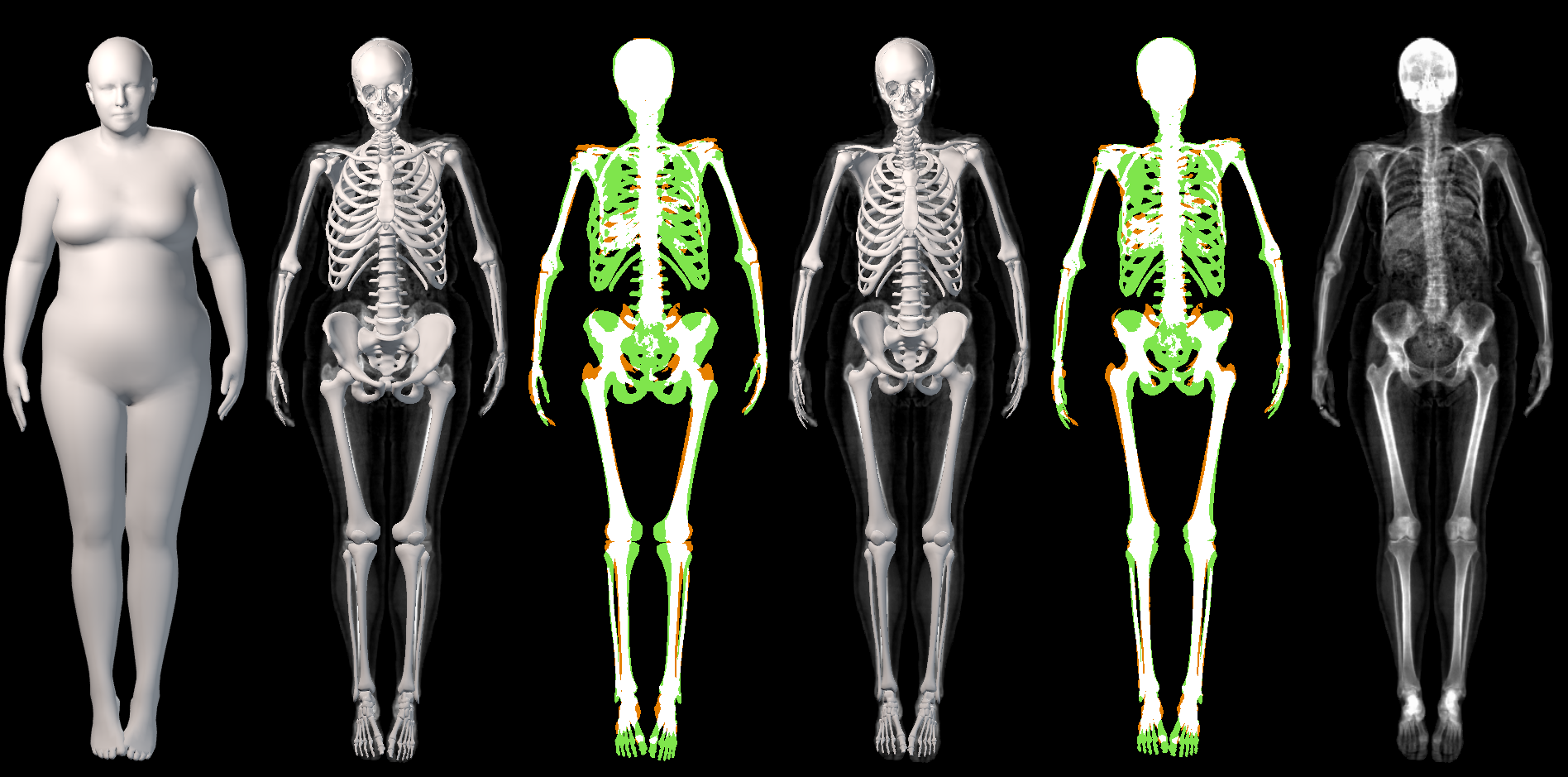}
        
        \includegraphics[width=0.490\columnwidth]{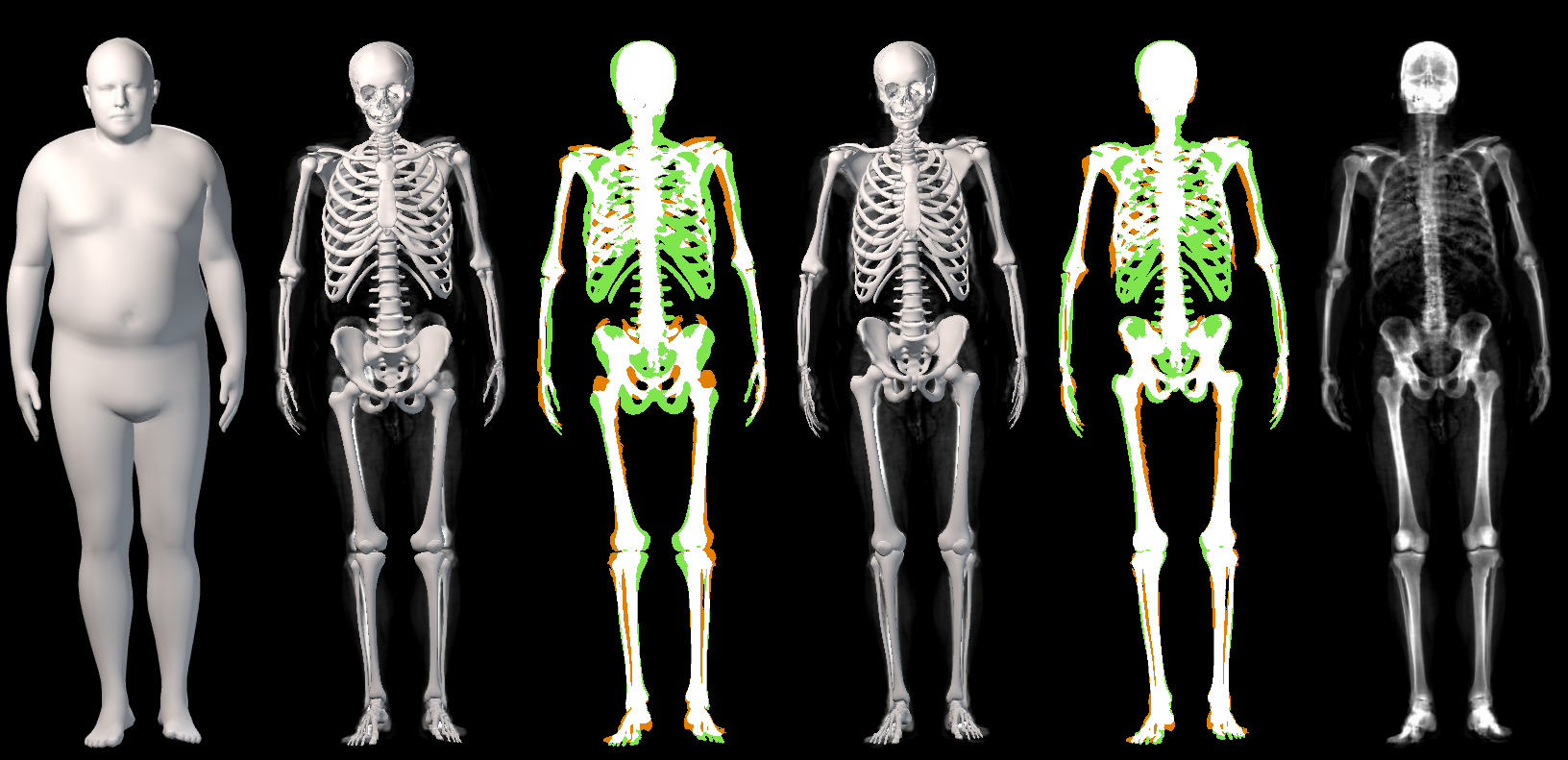}
        \includegraphics[width=0.490\columnwidth]{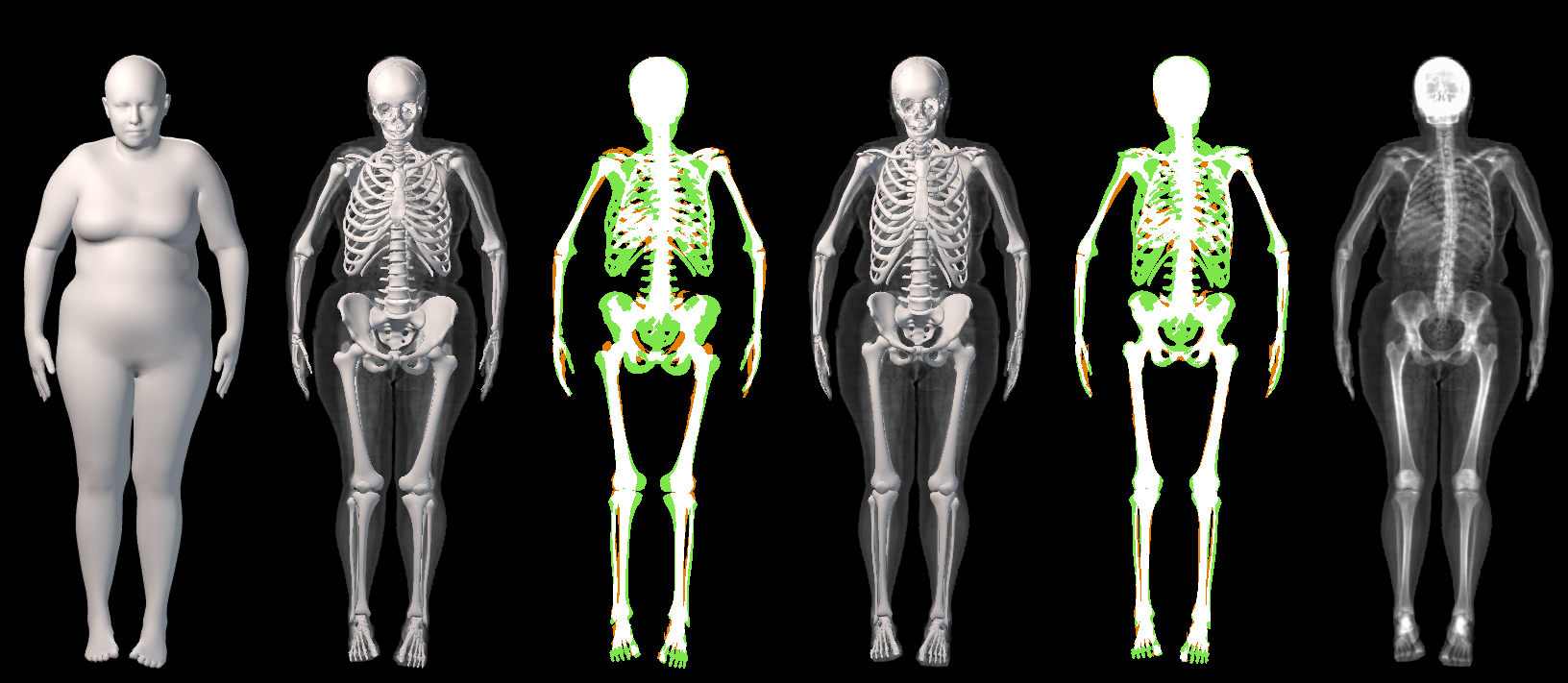}
    \end{mdframed}
    \includegraphics[width=0.490\textwidth]{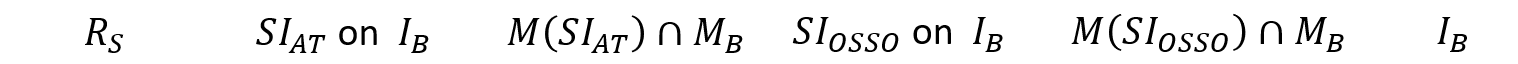}
    \includegraphics[width=0.490\textwidth]{figures/supmat/comparison_anatoscope/label_comparison.PNG}

    \caption{For each subject, we show in the order (1) $\reg_S$, (2) $SI_{AT}$ superimposed with the ground truth DXA $I_B$, (3) the overlap of $M(SI_{AT})$ and $I_B$, (4) $SI_{OSSO}$ superimposed with the ground truth DXA $I_B$, (5) the difference between $M(SI_{OSSO})$ and $M_B$, (6) the ground truth DXA $I_B$}
    \label{fig:reg_vs_anatoscope}
\end{figure*}

In Figure \ref{fig:reg_vs_anatoscope}, we present a qualitative comparison between our OSSO predictions and the ones from Anatomy Transfer. This results complement Sec. 5.3 of the main document.

From the DXA test set, we select 5 subjects spanning the dataset BMI distribution. From the skin alignment $\reg_S$, we infer the skeleton and compare it to the subject's skeleton DXA image. 
We denote $SI_{AT}$ the skeleton inferred with AT and $SI_{OSSO}$ the skeleton inferred with OSSO. $M(SI)$ is the mask rendered from the mesh $SI$.

As can be seen from the images, our predictions do better capture the global shape of the skeletons. 
Particularly, Anatomy Transfer often estimates the location of the hips to be too low with respect to  the actual hips location. 
Our method predicts a skeleton which is visually closer to the one observed in the DXA images.

\subsection{Skeleton inference qualitative evaluation}

\paragraph{Lateral view}
Fig. \ref{fig:tpose_side} shows side views of the inference result in T-pose. 
While there is no ground truth to evaluate this pose with, the results are plausible.

\begin{figure*}
\centering
\includegraphics[height=7cm]{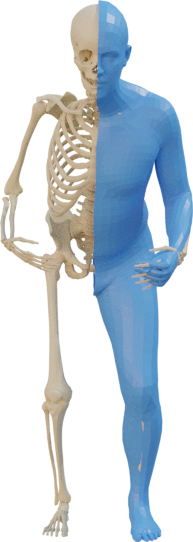}
\includegraphics[height=7cm]{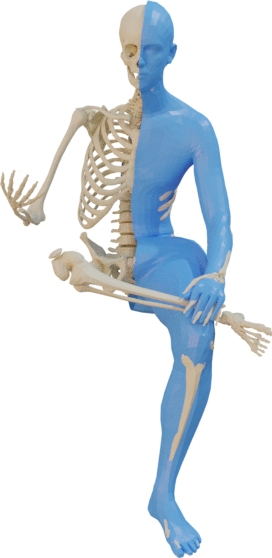}
\includegraphics[height=7cm]{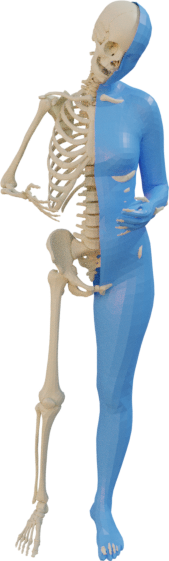}
\includegraphics[height=7cm]{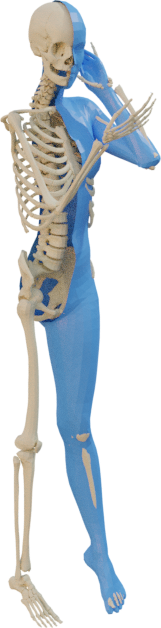}
\includegraphics[height=7cm]{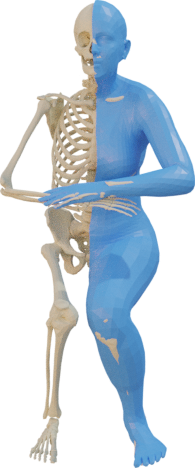}
\includegraphics[height=7cm]{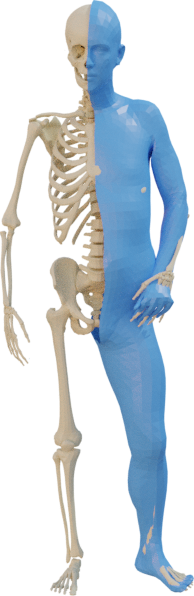}
\includegraphics[height=7cm]{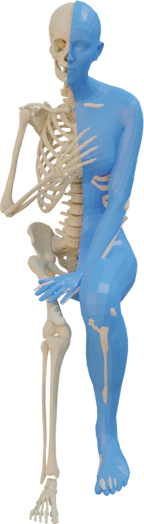}
\includegraphics[height=7cm]{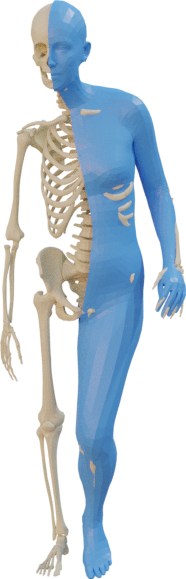}
\includegraphics[height=7cm]{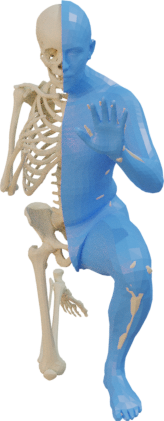}
\includegraphics[height=7cm]{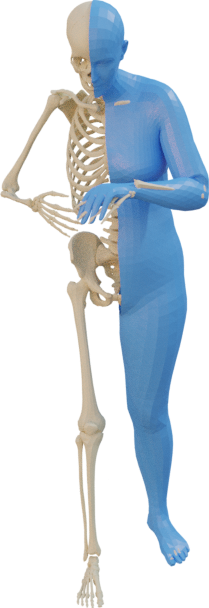}
\includegraphics[height=7cm]{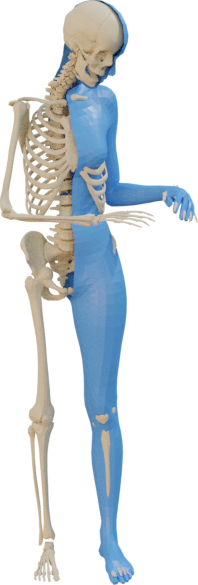}
\includegraphics[height=7cm]{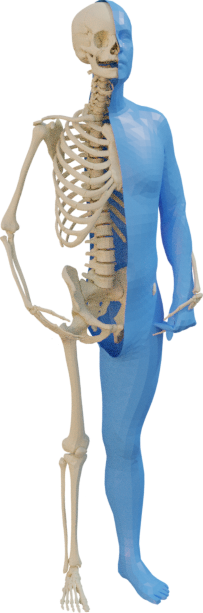}
\includegraphics[height=7cm]{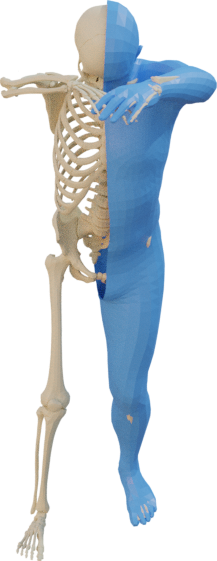}
\includegraphics[height=7cm]{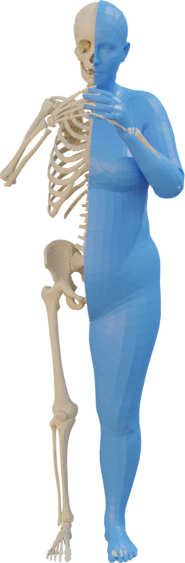}
\includegraphics[height=7cm]{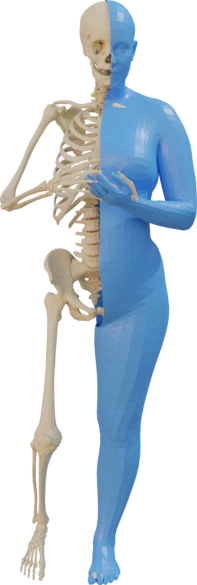}
\includegraphics[height=7cm]{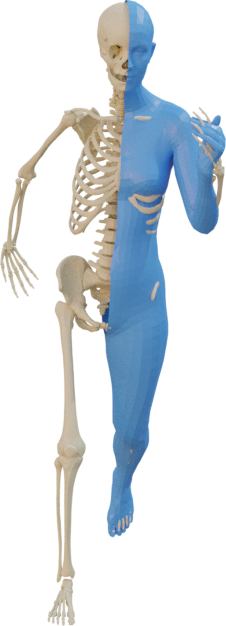}
\includegraphics[height=7cm]{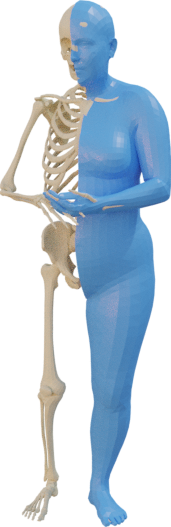}
\includegraphics[height=7cm]{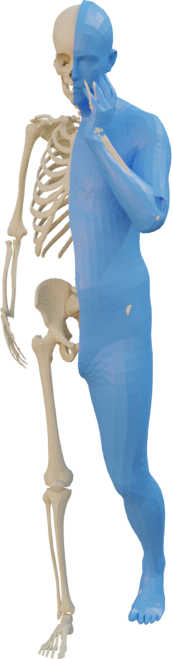}
\includegraphics[height=7cm]{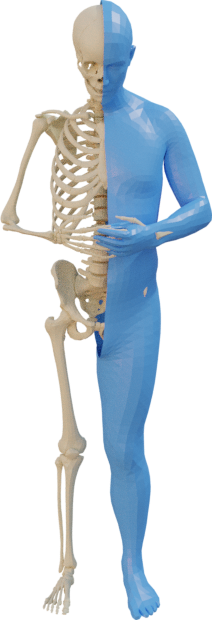}

\caption{Given SMPL bodies aligned to RenderPeople subjects \cite{renderpeople, patel2021agora}, we use OSSO to infer the underlying skeleton.}
\label{fig:rp_inference}
\end{figure*}

\vspace{-0.1in}
\paragraph{Inference on subjects from AGORA \cite{patel2021agora}}
Fig. \ref{fig:rp_inference} shows the inferred skeletons for subjects with different shapes and poses.

\newcommand{\bps}{1cm}
\newcommand{\tps}{0.5cm}

\begin{figure}[H]
  \centering
  \includegraphics[trim={0 \bps{} 0 \tps{}}, clip, width=0.2\columnwidth]{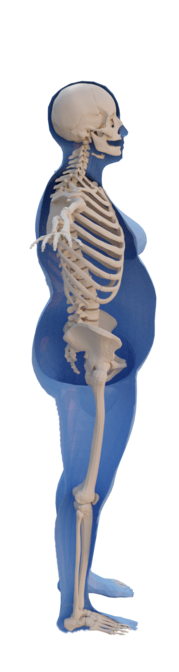}
  \includegraphics[trim={0 \bps{} 0 \tps{}}, clip, width=0.2\columnwidth]{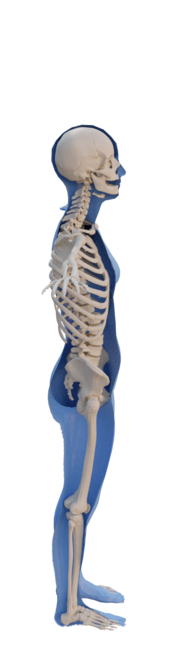}
  \includegraphics[trim={0 \bps{} 0 \tps{}}, clip, width=0.2\columnwidth]{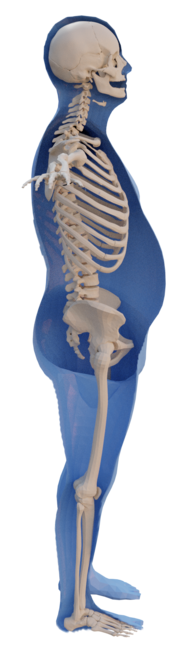}
  \includegraphics[trim={0 \bps{} 0 \tps{}}, clip, width=0.2\columnwidth]{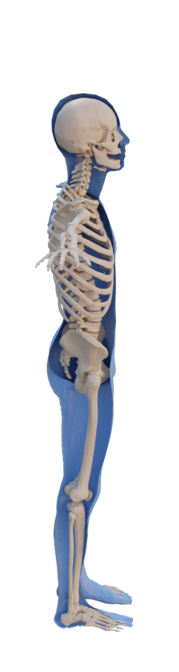}
  \caption{Lateral views of skeletons inferred with OSSO.}
  \label{fig:tpose_side}
\end{figure}

\end{appendices}

%% file: sections/notation_table.tex
\begin{table}[htbp]
\centering
\begin{tabular}{r c p{0.60\columnwidth}}
\toprule

\multicolumn{3}{c}{}\\
\multicolumn{3}{c}{\underline{Table of notation in OSSO}}\\
\multicolumn{3}{c}{}\\
$I_{S}$ & $\triangleq$ & Dxa soft tissue image (skin) \\
$I_{B}$ & $\triangleq$ & Dxa bone image (skeleton) \\

$M_S$ & $\triangleq$ & Skin mask segmented from $I_{S}$ \\
$M_B$ & $\triangleq$ & Skeleton mask segmented from $I_{B}$ \\

$\STAR(\shape_S, \pose_S)$ & $\triangleq$ & STAR body model \cite{osman2020star}\\
$\shapespace_S$ & $\triangleq$ & STAR shape space \\
$\skelinit(\shape_S, \pose_S)$ & $\triangleq$ & The initial skeleton model rigged to the STAR shape and pose parameters\\
$\stitched(\vect{t}, \vect{r}, \shape_B)$  & $\triangleq$ & Our {\it skeleton stitched puppet} model \\

$\hat{M}_B$  & $\triangleq$ & Synthetic skeleton mask generated with $\skelinit$\\ 

$\landmarks_I$ & $\triangleq$ & 29 3D landmarks whose 24 firsts correspond to STAR joints location and the closest skeleton vertices in $\skelinit$\\ 
$\tilde{\landmarks}_I$ & $\triangleq$ & 2D landmarks predicted from $M_B$. \\

$\reg_S$ & $\triangleq$ & STAR body model registered to $M_S$ \\
$\reg_B$ & $\triangleq$ & Our {\it skeleton stitched puppet} model registered to $M_S$ \\
$\unposed_B$ & $\triangleq$ & $\reg_B$ unposed in T pose \\

$\landmarks_B$  & $\triangleq$ &  63 3D landmarks defined as vertices on the skeleton mesh template\\

$\regressor_B$  & $\triangleq$ &  Regressor to predict skeleton landmarks $\landmarks_B$ from a STAR body model registration $\reg_S$\\

$\tilde{\landmarks}_B$  & $\triangleq$ &  $\landmarks_B$ landmarks location inferred with the regressor $\regressor_B$\\

$\shapespace_B$  & $\triangleq$ & PCA model of the skeleton learned from $\unposed_B$ \\

$\shapespace_S$  & $\triangleq$ & STAR PCA shape space\\

$\regressor_{\beta}$  & $\triangleq$ &  Regressor to predict skeleton shape components $\beta_B \in \shapespace_B$ from STAR shape components $\beta_S \in \shapespace_S$\\

$SI_{AT}$ & $\triangleq$ & Skeleton mesh inferred with AT \\
$SI_{OSSO}$ & $\triangleq$ & Skeleton mesh inferred with OSSO \\

\bottomrule
\end{tabular}
\caption{Table of Notation}
\label{tab:notation_table}
\end{table}

%% file: sections/supmat_ldm_array.tex
\begin{table*}
\centering
        \begin{tabular}[t]{|c|c|c|}
            \hline
            & female & male  \\ 
            \hline
            &   \multicolumn{2}{c|}{err. (mm) (mean $\pm$ std)} \\ 
            \hline
             L0 & 9.03 $\pm$ 5.52 & 10.28 $\pm$ 10.28  \\  L1 & 14.41 $\pm$ 8.79 & 12.60 $\pm$ 12.60  \\  L2 & \textcolor{red}{15.74 $\pm$ 8.49} & 13.90 $\pm$ 13.90  \\  L3 & 9.99 $\pm$ 4.81 & 10.69 $\pm$ 10.69  \\  L4 & \textcolor{green}{4.23 $\pm$ 2.00} & \textcolor{green}{4.42 $\pm$ 4.42}  \\  L5 & 8.38 $\pm$ 5.39 & 9.37 $\pm$ 9.37  \\  L6 & 9.72 $\pm$ 5.80 & 10.81 $\pm$ 10.81  \\  L7 & 14.76 $\pm$ 8.36 & 13.95 $\pm$ 13.95  \\  L8 & \textcolor{red}{15.93 $\pm$ 8.47} & 14.59 $\pm$ 14.59  \\  L9 & \textcolor{green}{4.06 $\pm$ 1.97} & \textcolor{green}{4.57 $\pm$ 4.57}  \\  L10 & 10.76 $\pm$ 5.14 & 11.12 $\pm$ 11.12  \\  L11 & 9.46 $\pm$ 5.57 & 9.86 $\pm$ 9.86  \\  L12 & \textcolor{green}{2.03 $\pm$ 1.04} & \textcolor{green}{1.96 $\pm$ 1.96}  \\  L13 & \textcolor{green}{2.89 $\pm$ 1.73} & \textcolor{green}{2.58 $\pm$ 2.58}  \\  L14 & \textcolor{green}{3.34 $\pm$ 2.00} & \textcolor{green}{3.26 $\pm$ 3.26}  \\  L15 & \textcolor{green}{3.67 $\pm$ 2.05} & \textcolor{green}{3.49 $\pm$ 3.49}  \\  L16 & \textcolor{green}{2.42 $\pm$ 1.35} & \textcolor{green}{2.28 $\pm$ 2.28}  \\  L17 & \textcolor{green}{3.33 $\pm$ 1.81} & \textcolor{green}{3.15 $\pm$ 3.15}  \\  L18 & 11.20 $\pm$ 5.47 & 10.90 $\pm$ 10.90  \\  L19 & 9.91 $\pm$ 5.01 & 8.44 $\pm$ 8.44  \\  L20 & 11.50 $\pm$ 5.83 & 13.34 $\pm$ 13.34  \\  L21 & 9.96 $\pm$ 4.94 & 8.53 $\pm$ 8.53  \\  L22 & 6.76 $\pm$ 3.16 & 6.93 $\pm$ 6.93  \\  L23 & 7.17 $\pm$ 3.56 & 7.24 $\pm$ 7.24  \\  L24 & 5.29 $\pm$ 2.65 & 5.87 $\pm$ 5.87  \\  L25 & 5.31 $\pm$ 2.69 & \textcolor{green}{4.99 $\pm$ 4.99}  \\  L26 & 7.74 $\pm$ 3.92 & 7.47 $\pm$ 7.47  \\  L27 & 5.72 $\pm$ 3.46 & \textcolor{green}{4.57 $\pm$ 4.57}  \\  L28 & 5.44 $\pm$ 2.68 & 5.22 $\pm$ 5.22  \\  L29 & 6.66 $\pm$ 3.22 & 6.40 $\pm$ 6.40  \\  L30 & 10.83 $\pm$ 5.08 & 10.85 $\pm$ 10.85  \\  L31 & 8.94 $\pm$ 4.84 & 8.10 $\pm$ 8.10  \\ 
            \hline
            
            \hline
        \end{tabular}
        \begin{tabular}[t]{|c|c|c|}
            \hline
            & female & male  \\ 
            \hline
            &   \multicolumn{2}{c|}{err. (mm) (mean $\pm$ std)} \\ 
            \hline
             L32 & 10.75 $\pm$ 5.10 & 11.65 $\pm$ 11.65  \\  L33 & 6.88 $\pm$ 3.37 & 6.40 $\pm$ 6.40  \\  L34 & 6.23 $\pm$ 2.58 & 6.42 $\pm$ 6.42  \\  L35 & 8.47 $\pm$ 4.79 & 7.96 $\pm$ 7.96  \\  L36 & 5.28 $\pm$ 2.53 & 5.21 $\pm$ 5.21  \\  L37 & \textcolor{green}{4.91 $\pm$ 2.63} & \textcolor{green}{4.24 $\pm$ 4.24}  \\  L38 & 7.19 $\pm$ 3.00 & 6.95 $\pm$ 6.95  \\  L39 & \textcolor{green}{4.92 $\pm$ 2.52} & \textcolor{green}{4.28 $\pm$ 4.28}  \\  L40 & 5.27 $\pm$ 2.66 & \textcolor{green}{4.47 $\pm$ 4.47}  \\  L41 & 6.39 $\pm$ 3.76 & \textcolor{green}{4.65 $\pm$ 4.65}  \\  L42 & 12.68 $\pm$ 7.17 & 10.93 $\pm$ 10.93  \\  L43 & 12.40 $\pm$ 7.77 & 11.08 $\pm$ 11.08  \\  L44 & 11.26 $\pm$ 6.14 & 10.44 $\pm$ 10.44  \\  L45 & 11.96 $\pm$ 5.93 & 9.85 $\pm$ 9.85  \\  L46 & 9.22 $\pm$ 4.40 & 9.37 $\pm$ 9.37  \\  L47 & 10.33 $\pm$ 5.51 & 10.13 $\pm$ 10.13  \\  L48 & 9.37 $\pm$ 4.21 & 9.78 $\pm$ 9.78  \\  L49 & 6.84 $\pm$ 3.29 & 7.69 $\pm$ 7.69  \\  L50 & 8.16 $\pm$ 3.93 & 7.62 $\pm$ 7.62  \\  L51 & \textcolor{green}{4.57 $\pm$ 2.21} & \textcolor{green}{4.53 $\pm$ 4.53}  \\  L52 & 7.85 $\pm$ 3.95 & 6.68 $\pm$ 6.68  \\  L53 & 5.82 $\pm$ 2.89 & 5.13 $\pm$ 5.13  \\  L54 & \textcolor{green}{0.95 $\pm$ 0.52} & \textcolor{green}{0.98 $\pm$ 0.98}  \\  L55 & \textcolor{green}{1.69 $\pm$ 0.89} & \textcolor{green}{1.90 $\pm$ 1.90}  \\  L56 & \textcolor{green}{1.40 $\pm$ 0.74} & \textcolor{green}{1.47 $\pm$ 1.47}  \\  L57 & 12.81 $\pm$ 7.43 & 11.38 $\pm$ 11.38  \\  L58 & \textcolor{red}{15.95 $\pm$ 9.94} & 13.96 $\pm$ 13.96  \\  L59 & 12.62 $\pm$ 6.91 & 11.32 $\pm$ 11.32  \\  L60 & \textcolor{red}{20.13 $\pm$ 10.65} & \textcolor{red}{17.36 $\pm$ 17.36}  \\  L61 & 10.62 $\pm$ 4.44 & 8.80 $\pm$ 8.80  \\  L62 & \textcolor{red}{20.51 $\pm$ 11.31} & \textcolor{red}{16.47 $\pm$ 16.47}  \\ 
            \hline
            
            \hline
        \end{tabular}
     \caption{Errors on the $\landmarks_B$ landmarks regression in millimeters. In green the errors below 5 mm, in red the errors over 15 mm. The landmark numbers are visually shown in \figref{fig:landmark_numbers}. }
     \label{tab:3d_regressed_ldm_evaluation}
\end{table*}